\pdfoutput=1
\documentclass[11pt,oneside]{book}
\usepackage[margin=1.2in]{geometry}
\usepackage{setspace}
\usepackage[toc,page]{appendix}
\usepackage[none]{hyphenat} % turn hyphenation off by default
\usepackage{graphicx}
\usepackage{amsmath}
\usepackage{amssymb}
\usepackage{float}
\usepackage{algorithm,algpseudocode}
\usepackage{algorithmicx}
\usepackage{subfig}
\usepackage{caption}
\usepackage{tabularx}

\begin{document}

\frontmatter

\begin{titlepage}

% You need to edit the details here

\begin{center}
{\LARGE University of Sheffield}\\[1.5cm]
\linespread{1.2}\huge {\bfseries Investigating the effects Diversity Mechanisms have on Evolutionary Algorithms in Dynamic Environments}\\[1.5cm]
\linespread{1}
\includegraphics[width=5cm]{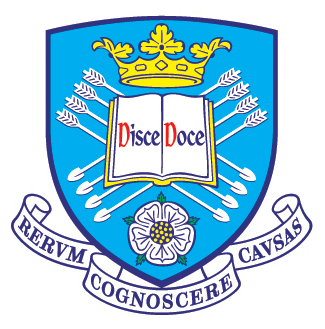}\\[1cm]
{\Large Matthew Hughes}\\[1cm]
{\large \emph{Supervisor:} Dirk Sudholt}\\[1cm]
\large A report submitted in fulfilment of the requirements\\ for the degree of MComp in Computer Science\\[0.3cm] 
\textit{in the}\\[0.3cm]
Department of Computer Science\\[2cm]
\end{center}

\end{titlepage}

% -------------------------------------------------------------------
% Declaration
% -------------------------------------------------------------------

\newpage

\chapter*{\Large Declaration}

\setstretch{1.1} % set the line spacing differently if you wish, but this looks good to me. 

All sentences or passages quoted in this report from other people's work have been specifically acknowledged by clear cross-referencing to author, work and page(s). Any illustrations that are not the work of the author of this report have been used with the explicit permission of the originator and are specifically acknowledged. I understand that failure to do this amounts to plagiarism and will be considered grounds for failure in this project and the degree examination as a whole.\\[1cm]

\noindent Name:\\[1mm]
\rule[1em]{25em}{0.5pt}

\noindent Signature:\\[1mm]
\rule[1em]{25em}{0.5pt}

\noindent Date:\\[1mm]
\rule[1em]{25em}{0.5pt}

% -------------------------------------------------------------------
% Abstract
% -------------------------------------------------------------------

\chapter*{\Large Abstract}

Evolutionary algorithms have been successfully applied to a variety of optimisation problems in stationary environments. However, many real world optimisation problems are set in dynamic environments where the success criteria shifts regularly. Population diversity affects algorithmic performance, particularly on multiobjective and dynamic problems. Diversity mechanisms are methods of altering evolutionary algorithms in a way that promotes the maintenance of population diversity. This project intends to measure and compare the performance effect a variety of diversity mechanisms have on an evolutionary algorithm when facing an assortment of dynamic problems.

\chapter*{\Large Acknowledgements}

I would like to thank my supervisor Dirk Sudholt for the guidance and encouragement he has provided me with throughout this project.

% -------------------------------------------------------------------
% TOC etc
% -------------------------------------------------------------------

\tableofcontents
\listoffigures
\listoftables

\setstretch{1.1} 

\mainmatter

\chapter{Introduction}

\section{Background}
Evolutionary algorithms are a class of algorithms that draw inspiration from the biological process of evolution. The basic principal is that the algorithm generates a population of solutions and then subsequently allows it to evolve (survival of the fittest) with the aim of finding individuals within that population that are highly suited to the environment (which represents the problem)

Evolutionary algorithms have been successfully applied to a large number of single and multiobjective optimisation problems in both stationary and dynamic environments. However simple evolutionary algorithms have the tendency to converge to local optima. The maintenance of a diversity population can help prevent this premature convergence by ensuring that the search space is adequately explored. 

Population diversity is also important for multiobjective and dynamic problems. In these types of problems it is crucial that the search space is sufficiently explored in order to either find multiple optima or to track the movement of an optima as the success criteria changes. To help maintain a diverse population there exist several diversity promotion mechanisms.

\section{Aims}
The aim of this project is to study the performance and behaviour of different evolutionary algorithms on dynamic optimisation problems. Specifically to compare a variety of diversity mechanisms in order to see which mechanisms perform best in a variety of scenarios.

\section{Overview of Dissertation}

This chapter has provided a very brief introduction to evolutionary algorithms  and the general aim of this project. Chapter 2 will provide a more in depth overview of evolutionary algorithms, diversity mechanisms as well as how to test and measure the performance of said algorithms. Chapter 3 will discuss the specific aims of the project such as which algorithms shall be tested and on what problem(s). Chapter 4 will explain the coding behind the various algorithms, problems and measures. Chapter 5 will present the results of the tests and discuss them. Chapter 6 will conclude the report and suggest further work.

\chapter{Literature survey}

\section{Evolutionary Algorithms}

Evolutionary Algorithms are a class of algorithms that copy natural evolutionary principles. They are popular tools in the areas of search, optimisation, machine learning and for solving design problems \cite{Whitley}. There are many different variants of evolutionary algorithms, all of which are based on the same basic principle. Given a population of individuals the environmental pressure causes natural selection (survival of the fittest) and this causes a rise in the fitness of the population \cite[p.15]{EibenSmith}. Generally evolutionary algorithms start with a randomly generated initial population, the members of which are evaluated by a fitness function. Based on this fitness candidates are chosen to populate the next generation (the chosen candidates may not be the ones with the highest fitness, however the higher the fitness the higher the likelihood of an individual being selected) by using crossover (or recombination) and/or mutation operators on them. The crossover operator takes two or more candidates (the parents) and generates one or more new candidates (the children). The mutation operator is applied to one candidate and results in one new candidate. The new candidates then compete with with the old ones for a place in the next generation. This process is iterated until a candidate with high enough fitness (a solution) is found or a set computational limit is reached. The general scheme of an evolutionary algorithm is shown in pseudo-code and via flowchart in algorithms 1 and figure 2.1 respectively.
%\begin{figure}[ht]
%	\centering
%	\includegraphics[width=10cm]{figures/figure1.png}
%	\caption{Pseudo-code for an evolutionary algorithm}
%	\label{fig:graph}
%\end{figure}

\begin{algorithm}
	\caption{A simple Evolutionary Algorithm}
	\begin{algorithmic}
		
		\State Initialise population with random candidate solutions
		\State Evaluate each candidate
		\Repeat
		\State Select Parents
		\State Recombine pairs of Parents
		\State Mutate the resulting Offspring
		\State Evaluate new candidates
		\State Select individuals for the next generation
		\Until{} Termination condition is satisfied
	\end{algorithmic}
\end{algorithm}

\begin{figure}[h]
	\centering
	\includegraphics[width=10cm]{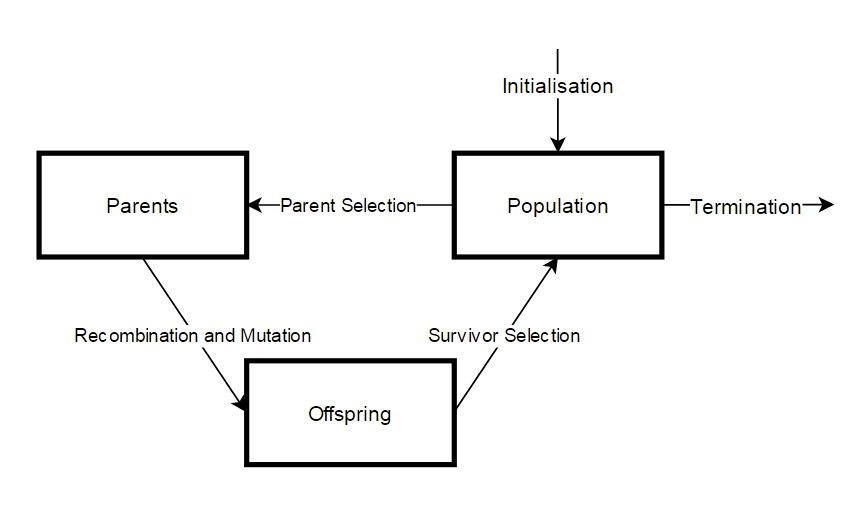}
	\caption{An evolutionary algorithm as a flow chart}
	\label{fig:graph}
\end{figure}

\section{Diversity}

In evolutionary algorithms the term diversity indicates dissimilarities of individuals. Diversity is best represented visually, figure 2.2 is an example of a population with a high level of diversity vs a population with low level of diversity. 

\begin{figure}[h]
	\centering
	\subfloat[Population with High Diversity]{\includegraphics[width=0.5\textwidth,height=5cm]{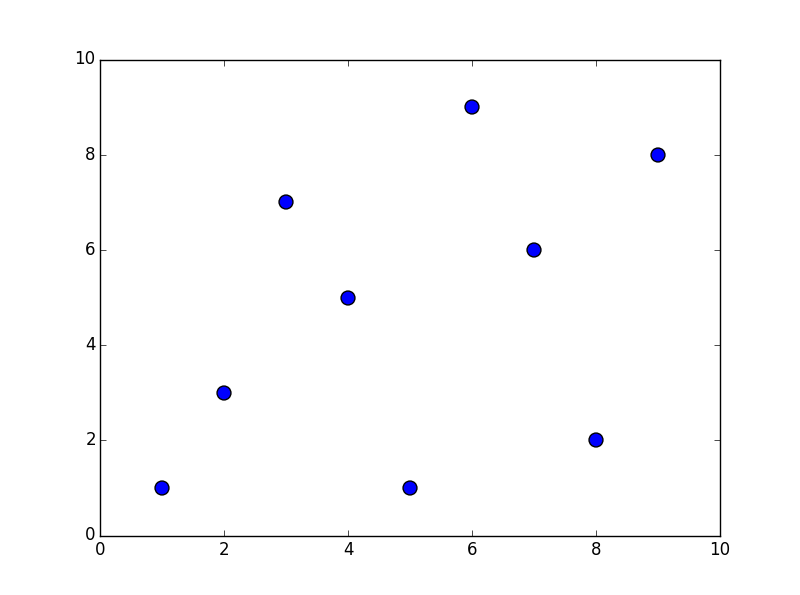}\label{fig:f1}}
	\hfill
	\subfloat[Population with Low Diversity]{\includegraphics[width=0.5\textwidth,height=5cm]{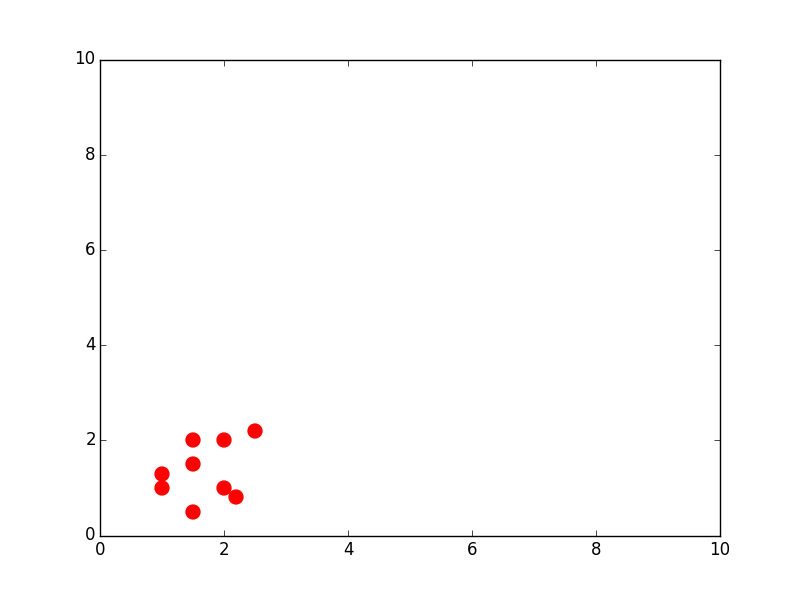}\label{fig:f2}}
	\caption{High and Low Population Diversity in a 2D space}
\end{figure}

Diversity is an important characteristic of a population as without it there is a risk of the individual with the highest fitness taking over the whole population before the fitness landscape has been properly explored \cite{DirkDiversity}. Maintaining diversity in a population is of particular importance when dealing with multiobjective and dynamic problems. Both types of problems need for the entire fitness landscape to be searched as multiobjective problems have several optima located throughout the landscape and in dynamic problems the optima moves from place to place. In order to maintain a population's diversity there are multiple diversity mechanisms that can be used. Techniques for diversifying a population typically reduce selection pressure, selection noise or operator disruption (or some combination of the three) \cite{Mahfoud95nichingmethods}. Diversity mechanisms can be split into two groups, niching techniques and non-niching techniques. 

The diversity of individuals and populations can be measured either in the genotype or in the phenotype space. When the phenotype or genotype are made up of a fixed number of parameters p, the tools of mathematical analysis and cluster analysis can be applied to obtain a measure of diversity \cite{theodoridis2009pattern}. However, it is usually the case that the genotypic space is easier to measure as it is generally much simpler \cite{DBLP:journals/ec/MattiussiWF04}. Principal genotypic measures are entropy, and more commonly, pair-wise Hamming distance \cite{DBLP:conf/ae/MorrisonJ01}. Pairwise Hamming Distance is defined as:

$$D(P)=\sum_{i=1}^{n-1} \sum_{j=i+1}^{n}d_h(i_j,i_k)$$

where $d_h(i_j,i_k)$ is the hamming distance between two individuals and $n$ is the number of individuals in the population.

Population diversity is much more efficiently computed using the moment of inertia method \cite{DBLP:conf/ae/MorrisonJ01}. When calculating the moment of inertia of binary strings, each bit is assumed to be an independent dimension. In this case the coordinates of the centroid, $(c_1,c_2,c_3,\ldots,c_L)$ of $P$ bit strings of length $L$ are computed as:

$$ c_i=\frac{\sum_{j=1}^{j=P} x_{ij}}{P} $$

and the moment of inertia about the centroid is:

$$ I=\sum_{i=1}^{i=L} \sum_{j=1}^{j=P} (x_{ij} - c_i)^2 $$

where $s_{ij}$ is the bit in position $i$ of the $j^{th}$ string and $c_i$ is the $i^{th}$ coordinate of the centroid.

It can be proved that the moment of inertia is the same as the pair-wise Hamming distance divided by the size of the population.

\section{Selective Pressure}

The term selective pressure is used to describe the tendency to select only the best individuals from the current generation to propagate the next one. A certain amount of selective pressure is necessary to ensure that an evolutionary algorithm reaches an optimum. However, too much selective pressure can cause the levels of genetic diversity within a population to drop, this increases the chance that the global optimum will be missed and the algorithm will converge to a local optimum. On the other hand, too little selective pressure will prevent the algorithm from converging in a reasonable time. It is necessary to have a good balance between selective pressure and genetic diversity so that evolutionary algorithms will converge to the global optimum within a reasonable time.

\section{Niching}

Niching algorithms are characterised by their capabilities of maintaining stable sub-populations (niches)\cite{mumford2009computational}. They excel at solving multi-objective optimisation problems.

\subsection{Fitness Sharing}

Fitness sharing is a diversity mechanism based on the idea that all the individuals in a particular niche have to share the resources available, as in nature. This means the number of individuals in a certain niche is controlled by sharing their fitness before selection occurs, this encourages the population to diversify as there is a fitness penalty in densely populated areas. Practically the mechanism works by measuring the distance $d(i, j)$ of every possible pair of individuals $i$ and $j$ in the population. The fitness $F$ of each individual $i$ is then adjusted according to the number of individuals inside its sharing radius $\sigma_{share}$ using a power law distribution \cite[p.163]{EibenSmith}\cite{fitnessSharing}.

$$F'(i)=\frac{F(i)}{\sum_{j} sh(d(i,j))}$$

where the sharing function $sh(d)$ is a function of the distance $d$ given by

\[sh(d)= 
	\begin{cases}
		1-(d/ \sigma_{share})^\alpha & \text{if } d\leq\sigma_{share}\\
		0 							 & \text{otherwise}\\
	\end{cases}
\]

The sharing radius, $\sigma_{share}$, defines the niche size. The value of $\sigma_{share}$ decides how many niches can be maintained and the granularity with which different niches can be discriminated. Individuals within the sharing radius of another will reduce that individuals fitness proportionally with how close the two individuals are. For problems in binary space the distance is usually measured by using Hamming distance. Figure 2.3 shows how a sharing radius affects the fitness value of the individual in the center, a larger sharing radius will lead to larger fitness penalties, as more individuals are inside the niche.  

\begin{figure}[h]
	\centering
	\includegraphics[width=8cm]{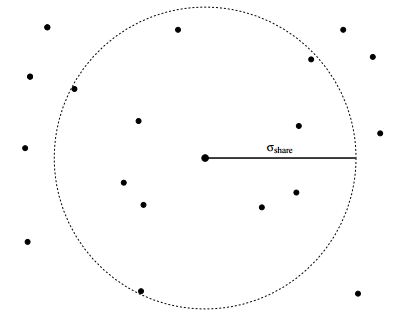}
	\caption{How the value of  $\sigma_{share}$ affects niche size }
	\label{fig:graph}
\end{figure}

\subsection{Clearing}

The clearing method \cite{DBLP:conf/icec/Petrowski96} is similar to fitness sharing as it also encourages diversity by limiting the resources in the environment. Unlike fitness sharing where the resources are split amongst all individuals in a niche, clearing allocates the resources only to the best member(s) of the subpopulation. The difference in population distribution between clearing and fitness sharing can be seen in figure 2.4.
\begin{figure}[H]
	\centering
	\subfloat[Fitness Sharing]{\includegraphics[width=0.5\textwidth,height=5cm]{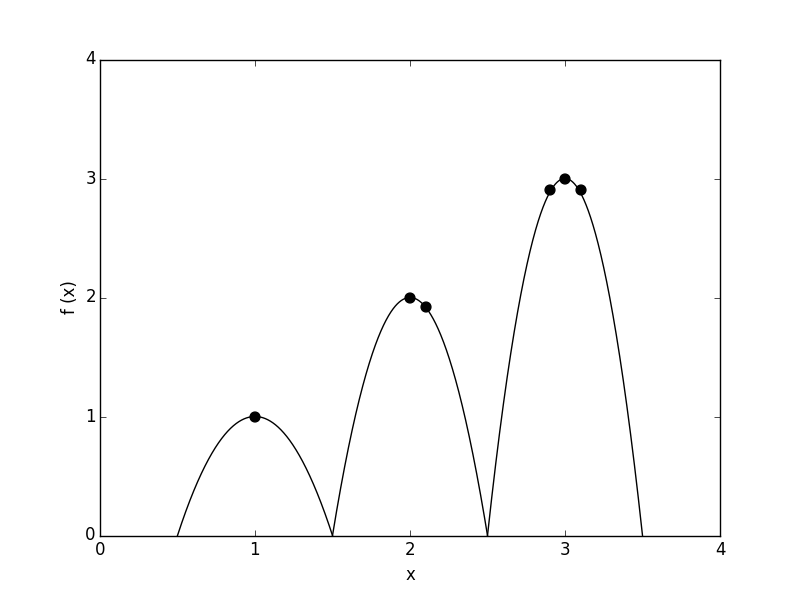}\label{fig:f1}}
	\hfill
	\subfloat[Clearing]{\includegraphics[width=0.5\textwidth,height=5cm]{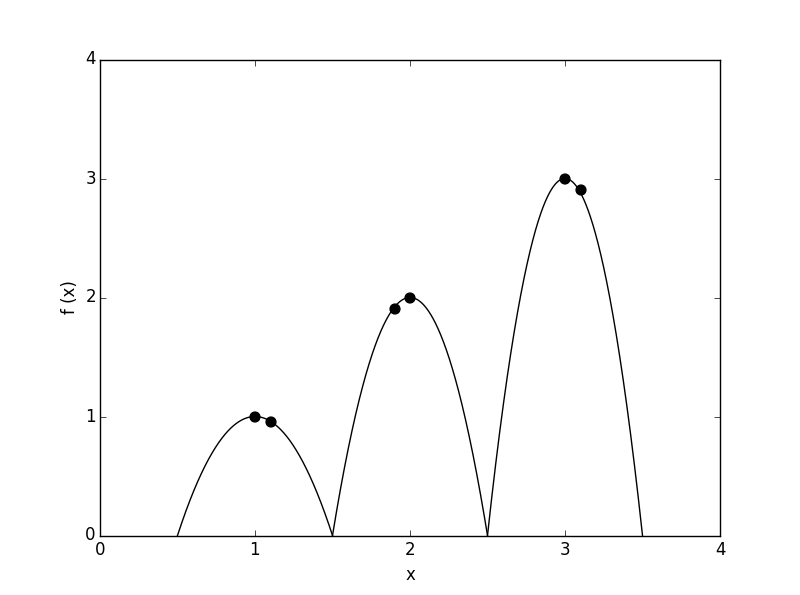}\label{fig:f2}}
	\caption{Optimal population distribution under fitness sharing (left) and clearing (right). Fitness sharing puts individuals on peaks with proportion to their fitness. Clearing splits the population evenly between all peaks.}
\end{figure}
In practise, each niche has a maximum capacity $k$ which specifies the number of individuals that can be within a certain niche. Clearing allocates the available resources to the $k$ best individuals in a niche and removes the others from the population. As with fitness sharing, the clearing method uses a dissimilarity measure between individuals to determine if they belong in the same subpopulation. If the distance between a pair of individuals is less than a dissimilarity threshold $\sigma_s$ (clearing radius) then those individuals are members of the same niche. Figure 2.3 can also be used to demonstrate the clearing radius, the difference being that only the best individuals inside the radius keep their fitness instead of proportionally decreasing all individuals fitnesses.

%\begin{figure}[H]
%	\centering
%	\includegraphics[width=10cm]{figures/figure5.png}
%	\caption{Optimal population distribution under fitness sharing (top) and clearing (bottom). Fitness sharing puts individuals on peaks with proportion to their fitness. Clearing splits the population evenly between all peaks.}
%	\label{fig:graph}
%\end{figure}

\subsection{Crowding}
The crowding algorithm was first proposed by De Jong in 1975 \cite{DeJong:1975:ABC:907087}. In the crowding algorithm a small portion of the population reproduce and die each generation. Each newly generated offspring replaces an existing member of the population, preferably the most similar one \cite{Mahfoud95nichingmethods}. To approximately achieve the replacement of the most similar member, a sample of the original population is taken and the new member replaces the most closely related member of the sample. Specifically,the mechanism is as follows. A percentage of the population, specified by the $generation gap$ (G), is chosen via fitness proportionate selection to undergo crossover and mutation. G x n individuals from the population are chosen to die (to be replaced by the new offspring). Each offspring finds the member it replaces as follows. A random sample of CF individuals is taken from the population where CF is called the $crowding factor$. Of the CF individuals the one most similar to the offspring being inserted gets replaced. Similarity is defined using bitwise (genotypic) matching. 

There are multiple variants of crowding designed to improve on the original with an improved version being deterministic crowding \cite{DBLP:conf/ppsn/1992}. Deterministic crowding uses preselection to reduce the amount of computation being done. The algorithm relies on the fact that the offspring are probably similar to their parents. After crossover and mutation the offspring and the parents compete with each offspring competing against the most similar parent for entry into the population. Algorithm 2 shows deterministic crowding in practise. 

\begin{algorithm}
	
	\caption{Deterministic Crowding}
	\begin{algorithmic}
		\Repeat
		\State Select two parents $p_1$ and $p_2$, randomly without replacement
		\State Apply crossover, yielding $c_1$ and $c_2$
		\State Apply mutation/other operators, yielding $c'_1$ and $c'_2$
		\If {$Distance(p_1,c'_1)+Distance(p_2,c'_2) <= Distance(p_1,c'_2)+Distance(p_2,c'_1)$ }
			\If {$fitness(c'_1)>fitness(p_1)$} replace $p_1$ with $c'_1$ \EndIf
			\If {$fitness(c'_2)>fitness(p_2)$} replace $p_2$ with $c'_2$ \EndIf
		\Else
			\If {$fitness(c'_2)>fitness(p_1)$} replace $p_1$ with $c'_2$ \EndIf
			\If {$fitness(c'_1)>fitness(p_2)$} replace $p_2$ with $c'_1$ \EndIf
		\EndIf
		\Until{}
	\end{algorithmic}
\end{algorithm}

\newpage
\section{Non Niching}
Non niching methods promote diversity in ways outside of maintaining sub-populations. Many non niching methods use restricted mating techniques. Restricted mating techniques promote diversity within a population by applying conditions to restrict or encourage the selection of which individuals will mate

\subsection{Removal of Genotype}

A simple diversity preservation method is to not allow genotypic duplicates within the population \cite{Friedrich:2009:ADM:1668000.1668003}. This means that offspring are tested to ensure that they aren’t an identical copy of an already existing member of the population. A similar method is to not allow phenotypic (fitness)  duplicates (individuals with the same fitness as an existing member of the population) into the population.

\subsection{Incest Prevention}

Incest prevention \cite{eshelman:icga91} promotes restricted mating between dissimilar individuals. Generally, when similar individuals are mated the offspring usually doesn't produce significant new information about the search space. In practise incest prevention only allows two parents to mate if their Hamming distance is above a given threshold. This threshold lowers after a generation where no offspring was accepted into the population.

\subsection{Island Models}

In island models the population is split into several subpopulations that are are evolved in parallel on separate “islands”. Islands evolve independently most of the time, but members from each island will periodically migrate to other islands. This migration of these individuals allows the different islands to communicate and compete with one another. This helps to coordinate the search by focusing on the most promising areas of the search space \cite{kacprzyk2015springer}. As the subpopulations are kept separate it is possible to use different algorithmic parameters or even different algorithms on different islands. This property is very useful as it can allow for the generation for a much more diverse set of solutions to be generated than could be achieved by running any single algorithm.

\subsection{Religion Based EA (RBEA)}

The RBEA \cite{DBLP:conf/ppsn/ThomsenRK00} is based on the religious principles of having children, believing in no other religion and converting non-believers, this allows for information to spread through a population the same way religion has done in the real world. Each religion is considered a subpopulation whose members can only mate amongst themselves (generating offspring of the same religion). Information can be transferred between the subpopulations by the method of religious conversion.

\section{Test Problems}

Evolutionary algorithms have been been successfully applied to a wide variety of static optimisation problems. However, many real world problems are set in dynamic environments. Thus, in order to measure the performance of evolutionary algorithms it is important to test them using dynamic optimisation problems. In general these dynamic problems can be split into two types. 

In the first type of problem, the environment will change between multiple static problems or multiple states of the same problem. An example of this type of problem is the dynamic knapsack problem \cite{DBLP:conf/ppsn/LewisHR98} where the knapsacks weight capacity varies over time (usually oscillating between two values). Yang proposed a method \cite{1299951} of converting binary-encoded stationary problems into dynamic ones using a bitwise XOR operator. His method changes the the fitness landscape but maintains the number of optima and the fitness value of the optima.

In the second type of problem the environment changes continuously, usually by making small changes to the shape of the previous environment. Examples of this would be the moving peaks benchmark \cite{785502} and the dynamic bit-matching problem \cite{yang2013evolutionary}. 

\subsection{Dynamic Bit Matching Problem}
In the dynamic bit-matching problem, an algorithm needs to find solutions with the lowest hamming distance to a target string which can change over time. This problems is relatively easy to define, easy to analyse and especially well suited to binary representations\cite{branke2012evolutionary}. The frequency of changes can be easily set and the severity of the changes can also be altered in a straightforward way. Severity can be increased or decreased by varying how many bits are altered at every change. This problem is limited in that it is unimodal, this means that a hill climbing algorithm would probably be more efficient than an evolutionary algorithm \cite{branke2012evolutionary}.

\subsection{Moving Peaks Benchmark}

The moving peaks benchmark consists of a multi-dimensional landscape consisting of several peaks, where the height, width and position of each peak is altered slightly every time a change in the environment occurs. The function is highly customisable allowing for the parameters affecting the number, height and shape of the peaks as well as the rate of change in the environment amongst other things \cite{movingPeaks}. The customisability allows for many different scenarios to be created to challenge an algorithm in a variety of ways. Success on the moving peaks benchmark is measures in an algorithms ability to locate and track the highest peak in the landscape and the variety of ways it can be set up mean that an algorithm may perform well on one configuration of this benchmark and poorly on another.

\newpage
\section{Measuring Performance}

In order to analyse the effects different diversity mechanisms have on evolutionary algorithms the performance needs to be measured, a variety of methods exist to do this.

\subsection{Running Fitness} 
A simple measure of performance would be to take the maximum fitness at each generation, this will show how an algorithms performs at all times in a run. The average or minimum fitness could also be recorded to give further information about the population.

\subsection{Likelihood of Optimality}
Assuming that an evolutionary algorithm was executed for $k$ generations in each of $n$ runs and $m$ is the number of runs that reached an optimal solution within those $k$ generations. The likelihood of optimality $Lopt(k)$ measure of that algorithm is the estimated probability $m/n$ \cite{Sugihara97measuresfor}.

\subsection{Average Fitness Value}
Assuming that an evolutionary algorithm was executed for $k$ generations in each of $n$ runs. The average fitness value $\bar{f}(k)$ of that algorithm at the $k$th generation is the average of the best fitness values obtained within $k$ generation in n runs \cite{Sugihara97measuresfor}. 

\subsection{Likelihood of Evolution Leap}
A generation is considered a leap if the solution it produces has higher fitness than any solution previously obtained before the generation. if an evolutionary algorithm was executed for $k$ generation in each of $n$ runs and $l$ is the average number of leaps within $k$ generations. The likelihood of evolution leap $Lel(k)$ for that algorithm at generation $k$ is the estimated probability $l/n$ (this is not an explicit measure of solution quality but it can still be of use in deciding how many generations to run an algorithm for) \cite{Sugihara97measuresfor}.

\subsection{Best-of-Generation}
The best-of-generation measure is calculated by averaging the highest fitness values for a generation of many individual runs on the same problem together. This can be used via a performance curve to show a whole picture of how an algorithm has performed. However, as the performance curve is not quantitative, it is challenging to compare the final outcome of different algorithms and to see if any statistically significant difference exists between them.

An improvement on this measure is the Average best-of-generation where the best-of-generation is averaged over all the generations. It is formulated as:

$$ \bar{F}_{BOG} = \frac{1}{G} \sum_{i=1}^{G} (\frac{1}{N} \sum_{j=1}^{N} F_{BOG_{ij}}) $$

where $g$ is the number of generations, $n$ is the number of runs and $F_{BOG_{ij}}$ is the best-of-generation fitness of generation $i$ of run $j$ \cite{yang2013evolutionary}.

\subsection{Offline Performance}
The offline performance of an algorithm is defined as the average, over a number of periods (a period is the time interval between two landscape changes), of the best solution found within the same period \cite{DBLP:journals/soco/Ben-RomdhaneAK13}.

$$ offline=\frac{1}{H} \sum_{k=1}^{H} f(BestSoFar_k) $$

The offline performance represents the algorithms ability to track a moving optima.

\subsection{Optimisation Accuracy}

The optimisation accuracy (also called the relative error) is defined as:

$$ accuracy_t = \frac{f(BOG_t)-Min_t}{Max_t-Min_t} $$

where $BOG_t$ is the best solution in the population at time $t$, $Max_t$ is the best possible fitness value in the search space and $Min_t$ is the worst possible fitness value in the search space. The accuracy ranges from 0 to 1 with a value of 1 and 0 representing the best and worst possible values, respectively \cite{DBLP:conf/ppsn/Weicker02}.

\subsection{Stability}

The stability of an algorithm is a measure of how well it is able to recover from a change in the environment. An algorithm is said to be stable it maintains its accuracy from one time step to another. The stability at generation $g$ is defined as:

$$ stability_g = max(0, accuracy_g-accuracy_{g-1}) $$

This measure ranges from 0 to 1, with a value close to 0 meaning high stability \cite{DBLP:conf/ppsn/Weicker02}.
\chapter{Requirements and Analysis}

The overall aim of this project is to produce results that show the performance of a variety of DMs on dynamic optimisation problems set in a binary space. 

\section{Diversity Mechanisms}

As discussed in the literature there are a variety of ways to encourage diversity in evolutionary algorithms. This project will look at some/all of the following diversity mechanisms in conjunction with a simple evolutionary algorithm:
\begin{itemize}
	\item Fitness Sharing
	\item Clearing
	\item Deterministic Crowding
	\item Incest Prevention
	\item Island Models	
	\item Removal of Genotype
\end{itemize}
In order to keep the  investigation fair all algorithmic parameters that can be kept constant shall be, these parameters are:
\begin{itemize}
	\item Population Size
	\item Offspring Size
	\item Crossover Rate
	\item Mutation Rate	
\end{itemize}

The RBEA algorithm will not be looked at in this dissertation as it requires the world to be represented as a grid \cite{DBLP:conf/ppsn/ThomsenRK00}. This is unfortunately incompatible with this project as high dimension binary spaces cannot be represented in this way.

\section{Benchmark Problem(s)}

The problem that will be used to benchmark these algorithms will be the moving peaks problem. The moving peaks problem was chosen as it is a highly customisable function which will allow for the testing of multiple scenarios by only changing a few parameters. To achieve this variety in problems without using moving peaks would require the use of multiple benchmarks. 

The tests will start of simple and can then be made more difficult to see how the different algorithms cope with increasingly complicated environments. For example, an early test can be to track a slowly moving single peaks. By then increasing the number of peaks in the function each algorithm can be tested to see how it copes when tracking multiple different simultaneously. An even more challenging test would be to then vary the heights of the peaks during the run to see how the algorithms perform with a shifting global optima.

Formally the moving peaks problem set in binary space can be represented:

\noindent
\\
Each peak $P_1,P_2,\ldots,P_k $ can be described by:
\begin{itemize}
	\item $ Position_i \in (0,1)^n $ 
	\item $ Height_i \in \mathbb{N} $
\end{itemize}
This is shown visually in figure 3.1:
\begin{figure}[H]
	\centering
	\includegraphics[width=12cm]{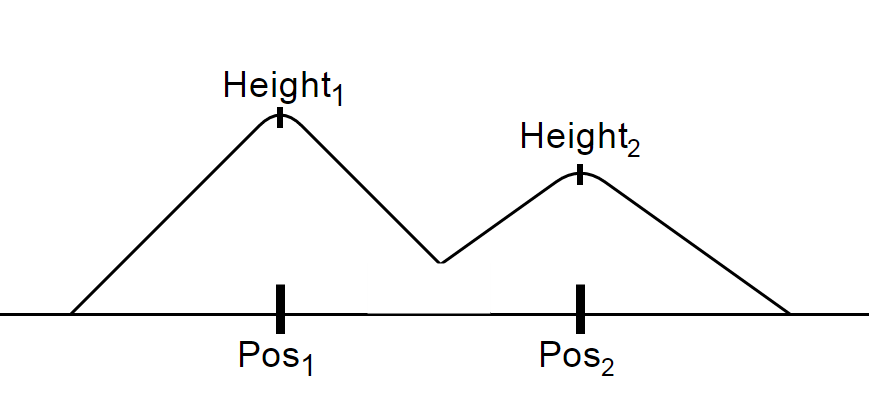}
	\caption{Binary Peaks search space}
	\label{fig:graph}
\end{figure}
The peaks can be transformed using a bitwise XOR operator as previously discussed \cite{1299951} or by using a mathematical function to alter the values of individual bits.

The fitness of any individual in the search space can be calculated with the formula:

$$ fitness(x) = \underset{y \in Peaks}{\max} (Height_y - Hamming(x,y))  $$

\section{Measuring Performance}

As we will be dealing with problems with shifting optima offline performance will be used to measure the success that the different algorithms have at tracking the optima.

As the project is investigating the effect that diversity mechanisms have on performance a measure of diversity will be taken. The moment of inertia measure will be taken, this was chosen due to its faster computation speed compared to other measures that would give the same information. 

The last thing to be measured will be the maximum, average and minimum fitness of the population at each generation. This information will allow the visualisation of the algorithms performance to be displayed on a graph.

\section{Distributed Evolutionary Algorithms in Python (DEAP)}

DEAP is a python framework designed for the rapid implementation and testing of evolutionary algorithms \cite{DEAP}. DEAP seeks to make algorithms explicit and data structures transparent, as opposed to the more common black-box frameworks. The framework is open source under an LGPL license.
DEAP will be used for the implementation section of this report as its transparent data structures will make it easier to extend than other frameworks. The framework also has basic implementations of evolutionary algorithms and a variety of benchmarking problems in both continuous and binary spaces.
\newpage
\section{Requirements}

\begin{center}
	\captionof{table}{Project Requirements}
	\begin{tabular}{|| p{13cm} | p{2cm}||} 
		\hline
		Requirement & Priority \\ [0.5ex] 
		\hline\hline
		Code the moving peaks problem & Mandatory  \\ 
		\hline
		Code a Basic evolutionary algorithm & Mandatory  \\ 
		\hline
		Code the Fitness Sharing Algorithm & Mandatory \\
		\hline
		Code the Deterministic Crowding Algorithm & Mandatory  \\
		\hline
		Code the Island Model Algorithm & Mandatory  \\  
		\hline
		Code the Clearing Algorithm & Desirable  \\  
		\hline
		Code the Incest Prevention Algorithm & Desirable  \\
		\hline
		Code the Removal of Genotype Algorithm & Desirable  \\
		\hline
		Test coded algorithms on a static problem & Mandatory  \\  
		\hline
		Test coded algorithms on a moving single peak & Mandatory  \\  
		\hline
		Test coded algorithms on a multiple moving peaks & Mandatory  \\  
		\hline
		Run coded algorithms on multiple moving peaks with variable height & Mandatory  \\  
		\hline
		Run coded algorithms on moving peaks where the movement speed of the peaks is increased & Optional  \\
		\hline
		Run coded algorithms on any other coded problems & Optional  \\
		\hline
		Calculate offline performance for all runs & Mandatory  \\
		\hline
		Calculate maximum, average and minimum fitness for all runs & Mandatory  \\
		\hline
		Calculate Best-of-Generation fitness for all runs & Optional  \\
		\hline
		Produce tables for all calculated results & Mandatory  \\
		\hline
		Produce graphs for all calculated results & Mandatory  \\
		\hline
	\end{tabular}
\end{center}

The mandatory requirements are those that form the core of the project and if they are not completed, the project will be a failure. The desirable requirements are things that would notably improve the project, but if they are not completed the project would still have met its intended purpose but it will be narrow in scope. The optional requirements are things that would be nice, but not completing them would have little impact on the success of the project.
\chapter{Design}

This chapter will cover the design of the previously mentioned algorithms, problems and measures. Everything will be programmed in python using the DEAP framework.

\section{Algorithms}

In all of the following algorithms, unless otherwise stated the Crossover, Mutation and Selection functions are as follows:

Crossover will be "Two Point Crossover" where two random points are chosen on the parents bitstrings and the bits are exchanged between these points to create the children. Figure 4.1 demonstrates how two point crossover works on example bitstrings.

\begin{figure}[H]
	\centering
	\includegraphics[width=11cm]{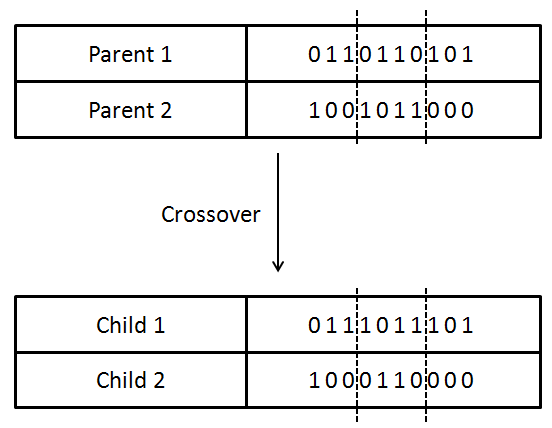}
	\caption{Two Point Crossover}
	\label{fig:graph}
\end{figure}

Mutation will be "Flip Bit Mutation" where each bit in the parent bitstring has an independent probability to flip which produces the child, this is shown in figure 4.2. For this project the probability of a bit flipping is set to $1/stringLength$.

\begin{figure}[H]
	\centering
	\includegraphics[width=11cm]{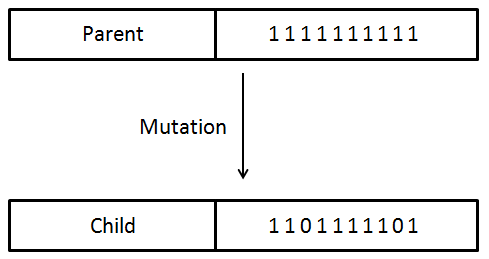}
	\caption{Flip Bit Mutation}
	\label{fig:graph}
\end{figure}

Selection will be done via "Tournament Selection" where a random selection of individuals will be entered into a tournament where the one with the highest fitness will be selected for the next generation. This process will repeat until enough individuals have been chosen to create the new population. For this project the tournament size will be set to 3. Algorithm 3 shows how tournament selection works in practise.

\begin{algorithm}
	
	\caption{Tournament Selection}
	\begin{algorithmic}
	\State Let $T$ be tournament size
	\State Let $N$ be number of individuals to select
	\While {Number of individuals Selected $< N$}
	\State Randomly select $T$ individuals
	\State Select the best individual
	\EndWhile
	\end{algorithmic}
\end{algorithm}

\subsection{Offspring Generation}
In all of the subsequent algorithms offspring will be generated using either the above crossover function or the mutation function. The number of offspring generated using each method is not exact but is instead based on probability. For the algorithms in this project the probability of an individual being generated using crossover is $p(crossover)=0.65$  and the probability of an individual being created by mutation is $p(mutation)=0.35$. The algorithm used to generate new individuals is Algorithm 4, in this project $p(crossover) + p(mutation)=1$, this means that the else state in Algorithm 4 is never used.

\begin{algorithm}[h]
	
	\caption{Offspring Generation}
	\begin{algorithmic}
		\Ensure $p(crossover) + p(mutation) <= 1$
		\State Let $\lambda$ be number of offspring to create
		\While {Number of offspring created $< \lambda$}
		\State Generate a random number $r$ between 0 and 1
		\If {$r<p(crossover)$}
			\State Randomly select two parents and generate offspring using crossover
		\ElsIf {$r<p(crossover)+ p(mutation)$}
			\State Randomly select a parent and generate offspring using crossover
		\Else
			\State Randomly select a member of the population which becomes an offspring
		\EndIf
		\EndWhile
	\end{algorithmic}
\end{algorithm}

\subsection{Basic Evolutionary Algorithm}
Algorithm 5 is the eaMuPlusLambda algorithm from the deap framework, it is just a standard evolutionary algorithm without any diversity preserving mechanisms. This algorithm will be used as a base that the diversity mechanisms will be added too. It will be used on its own as a baseline measure to show the performance of the various diversity mechanisms.

\begin{algorithm}[h]
	\caption{eaMuPlusLambda}
	\begin{algorithmic}
		\State Let $g=0$
		\State Generate an initial random population of individuals of size $\mu$
		\State Evaluate the fitness of the individuals
	    \While {$g <= MaxGen$}
		\State Generate $\lambda$ offspring using Offspring Generation
		\State Evaluate the fitness of the offspring
		\State Generate the new population, selecting from the offspring and the old population 
		\State Let $g=g+1$
		\EndWhile
	\end{algorithmic}
\end{algorithm}	

\subsection{Fitness Sharing}

Fitness sharing reduces the fitness of an individual proportionally to how close it is to others in the population. When using fitness sharing the population and offspring have to be evaluated together since the fitness of an individual is dependant upon its location in relation to the population. The formulas defining fitness sharing are in chapter 2. Algorithms 6 shows how Algorithm 5 will be altered to enact fitness sharing.

\begin{algorithm}[H]
	
	\caption{eaMuPlusLambda with Fitness Sharing}
	\begin{algorithmic}
		\State Let $g=0$
		\State Generate an initial random population of individuals of size $\mu$
		\State Evaluate the fitness of the individuals
		\While {$g <= MaxGen$}
		\State Generate $\lambda$ offspring using Offspring Generation
		\State Evaluate the fitness of the offspring and the population using fitness sharing
		\State Generate the new population, selecting from the offspring and the old population 
		\State Let $g=g+1$
		\EndWhile
	\end{algorithmic}
\end{algorithm}

\subsection{Clearing}

When using clearing the fitness of the population must be calculated twice, once to get the raw fitness and a second time to remove the weak individuals from the population, this adjustment is shown in Algorithm 7

\begin{algorithm}[H]
	
	\caption{eaMuPlusLambda with Clearing}
	\begin{algorithmic}
		\State Let $g=0$
		\State Generate an initial random population of individuals of size $\mu$
		\State Evaluate the fitness of the individuals
		\While {$g <= MaxGen$}
		\State Generate $\lambda$ offspring using Offspring Generation
		\State Evaluate the raw fitness of the offspring and the population
		\State Use Clearing to re-evaluate the fitness of the offspring and the population
		\State Generate the new population, selecting from the offspring and the old population 
		\State Let $g=g+1$
		\EndWhile
	\end{algorithmic}
\end{algorithm}	

Clearing is similar to fitness sharing but instead of lowering the fitness of every individual in a niche it instead allocates full fitness to the best members of the niche and zero fitness to the others. The size of the niches is set using a niche cap (which is the maximum number of individuals that can be in a niche) and a clearing radius (which defines how similar individuals need to be to be in the same niche). Algorithm 8 shows how clearing works, the parameters $radius$ and $cap$ are the clearing radius and the niche cap respectively.
\newpage

\begin{algorithm}[h]
	
	\caption{Clearing}
	\begin{algorithmic}
		\Function{Clearing}{$radius, cap$}
		\State Sort the population(P) by fitness value in decreasing order
		\For {$i=0$ to $len(pop)$}
		\If {$Fitness(P[i])>0$}
		\State $numWinner=1$
		\For {$j=i+1$ tos $len(pop)$}
		\If {$Fitness(P[j]) > 0$ and $Distance(P[i], P[j]) < radius$}
		\If {$numWinners<cap$} $numWinners+=1$
		\Else {} $Fitness(P[j]) = 0$ 
		\EndIf
		\EndIf
		\EndFor
		\EndIf
		\EndFor
		\EndFunction
	\end{algorithmic}
\end{algorithm}

\subsection{Deterministic Crowding}

Algorithm 9 shows how Algorithm 5 has been changed to implement deterministic crowding. As deterministic crowding requires all individuals to be created by crossover the offspring generation must also be altered, how it is altered is shown in Algorithm 10. The deterministic crowding algorithm used for selection is Algorithm 2 which can be found in chapter 2.

\begin{algorithm}[H]
	
	\caption{eaMuPlusLambda with Deterministic Crowding}
	\begin{algorithmic}
		\State Let $g=0$
		\State Generate an initial random population of individuals of size $\mu$
		\State Evaluate the fitness of the individuals
		\While {$g <= MaxGen$}
		\State Generate $\lambda$ offspring using Offspring Generation, remembering which offspring
		 were \State produced by which parents.
		\State Evaluate the fitness of the offspring.
		\State Generate the new population, using deterministic crowding to select between the \State parents and their offspring.
		\State Let $g=g+1$
		\EndWhile
	\end{algorithmic}
\end{algorithm}	

\begin{algorithm}[h]
	
	\caption{Offspring Generation for Deterministic Crowding}
	\begin{algorithmic}
		\State Let $\lambda$ be number of offspring to create 
		\While {Number of offspring created $< \lambda$}
		\State Generate a random number $r$ between 0 and 1
		\State Randomly select two parents and generate offspring using crossover
		\If {$r<p(mutation)$}
		\State Mutate the offspring
		\EndIf
		\EndWhile
	\end{algorithmic}
\end{algorithm}

\newpage
\subsection{Incest Prevention}

To prevent incest you must change how the parents are selected for crossover. In the other algorithms the parents are chosen at random but in incest prevention parents are only paired up with suitable mates. Which pairs of individuals are suitable to mate is decided by a minimum distance threshold between them. The main algorithm is the same as Algorithm 5, but the offspring generation function for choosing which individuals to crossover has been altered, this is shown in Algorithms 11 and 12.

\begin{algorithm}[h]
	
	\caption{Offspring Generation with Incest Prevention}
	\begin{algorithmic}
		\Ensure $p(crossover) + p(mutation) <= 1$
		\State Let $\lambda$ be number of offspring to create
		\While {Number of offspring created $< \lambda$}
		\State Generate a random number $r$ between 0 and 1
		\If {$r<p(crossover)$}
		\State Use Incest Prevention to select two parents, then generate offspring using crossover
		\ElsIf {$r<p(crossover)+ p(mutation)$}
		\State Randomly select a parent and generate offspring using crossover
		\Else
		\State Randomly select a member of the population which becomes an offspring
		\EndIf
		\EndWhile
	\end{algorithmic}
\end{algorithm}

\begin{algorithm}[H]
	
	\caption{Incest Prevention}
	\begin{algorithmic}
		\State Select a random individual from the population 
		\State Get the members of the population above the distance threshold
		\If {There are no individuals above the threshold}
		\State Select another random individual as the partner And reduce the threshold slightly
		\Else {} Randomly select a partner from the suitable members
		\EndIf
	\end{algorithmic}
\end{algorithm}

If the population becomes less diverse then the distance threshold is lowered, this is an attempt to pair up the least similar individuals in a similar population.

\newpage
\subsection{Removal of Genotype}%genotypes

Altering Algorithm 5 to prevent genotype duplicates is simple. The selection function is changed to only allow one individual of each genotype into the population.

\begin{algorithm}[H]
	
	\caption{eaMuPlusLambda with Genotype Removal}
	\begin{algorithmic}
		\State Let $g=0$
		\State Generate an initial random population of individuals of size $\mu$
		\State Evaluate the fitness of the individuals
		\While {$g <= MaxGen$}
		\State Generate $\lambda$ offspring using Offspring Generation
		\State Evaluate the fitness of the offspring
		\State Generate the new population, selecting only unique individuals from the offspring and \State the old population 
		\State Let $g=g+1$
		\EndWhile
	\end{algorithmic}
\end{algorithm}	

\subsection{Island Models}

In island models the population is split onto several "islands" where each subpopulation evolves separately from the others, periodically members will "migrate" from one island to another. Each individual island uses the steps from Algorithm 5 to evolve its population. Algorithm 14 shows how the island model will be implemented. 

\begin{algorithm}[H]
	
	\caption{eaMuPlusLambda with Island Model}
	\begin{algorithmic}
		\State Let $g=0$
		\State Concurrently for each of the islands generate an initial random population of individuals of size $\mu/numOfIslands$
		\State Evaluate the fitness of the individuals
		\While {$g <= MaxGen$}
		\For {$i$ in $numOfIslands$}
		\State Generate $\lambda/numOfIslands$ offspring using Offspring Generation
		\State Evaluate the fitness of the offspring
		\State Generate the new population, selecting from the offspring and the old population 
		\EndFor
		\If {$g=migrationGeneration$} migrate individuals using ring migration
		\EndIf
		\State Let $g=g+1$
		\EndWhile
	\end{algorithmic}
\end{algorithm}

Ring migration is when the same number of emigrants are chosen in each of the subpopulations and then moved between them in a circle. Emigrants from island one immigrate to island two, emigrants from island two immigrate to island three etc. This is shown visually in figure 4.3. 

\begin{figure}[H]
	\centering
	\includegraphics[width=11cm]{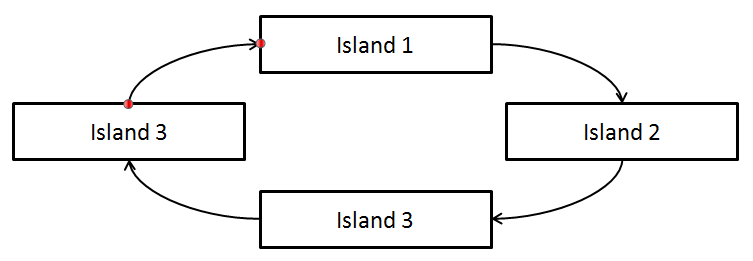}
	\caption{Ring Mutation}
	\label{fig:graph}
\end{figure}
\section{Problems}

\subsection{OneMax}
Problem one will be a single stationary peak represented using $OneMax$. The fitness function $OneMax:\{1\}^n$ is defined as:

$$OneMax(x) = \sum_{i=1}^{n} x[i]$$

$OneMax$ is a simple problem that needs no population diversity to solve successfully. While it is primarily being used to ensure all of the algorithms are working correctly this problem will also provide a baseline performance for how well each algorithm converges to a maxima. 

\subsection{TwoMax}
Problem two will have two stationary peaks represented using $TwoMax$. The fitness function $TwoMax:\{0,1\}^n$ is defined as:

$$TwoMax(x) = \max \{\sum_{i=1}^{n} x[i],\sum_{i=1}^{n}(1-x[i])\}$$

$TwoMax$ is another simple problem where the primary aim is to locate both peaks. This problem is a good basic test of each algorithms ability to promote diversity in a population. Without a large enough diversity the population will converge to a single peak, if this happens the algorithm will be considered to have failed to solve the problem.

\subsection{Moving Peak}
Problem three is a single moving peak, introducing movement into the problem will test the algorithms ability to track a shifting optima. The moving peaks problem is described in chapter 3, the fitness function is:

$$ fitness(x) = \underset{y \in Peaks}{\max} (Height_y - Hamming(x,y))  $$

The same fitness function will be used on all of the problems after this one.
\\
\\
\noindent
This problem is a test of how well an algorithm can reconverge to a peak after it has moved. While a high diversity level is not needed to succeed on this problem some diversity may help algorithms find the peak faster after it moves.

\subsection{Moving Peaks}
Problem four will have multiple moving peaks where each peak has a static height but the global optima will always be the same peak. This problem will test if an algorithm can find multiple peaks and then reconverge to them as they move. 

\subsection{Changing Height Peaks}
Problem five will have multiple static peaks where each peak has a moving height. This problem is a test of the algorithms ability to move from one peak to another when the global optima changes peak. To obtain success an algorithm will need to maintain subpopulations at each peak in order to quickly adapt to the height changes.

\subsection{Moving Peaks with variable height}
Problem six will have multiple moving peaks where each peak has a moving height. This problem will test the algorithms ability to track multiple shifting optima where the global optima also changes. It is a combination of the previous two problems and will be very challenging. In order to attain success on this problem an algorithm will need to be able to maintain subpopulations on multiple moving peaks, such that when a peaks height increases there are already individuals located there.

\section{Measures}

There are three areas of performance that need to be measured to show a complete picture of each algorithms performance. A running or "online" measure that will show how the algorithms perform at each generation. A overall or "offline" measure that will assign an overall value to an algorithms performance. And finally a measure of population diversity so that it can be shown if an algorithm is able to find multiple peaks as well as showing the effect diversity has on performance.

To measure the running performance of the algorithms I will be taking measures of the minimum, average and maximum fitness at each generation. The measure that will be used to show the overall performance of each algorithm will be "offline performance". To measure population diversity the moment of inertia will be taken.

Multiple measures are required as a single measure can be misleading. For example if an algorithm fails to find both peaks on $TwoMax$ then its offline performance will be the same as it was for $OneMax$ (which will likely be good due to the simple nature of $OneMax$). However, if a diversity measure is taken then we can see that the algorithm failed to find both peaks and this information can put the offline performance into context. 

\subsection{Minimum, Average and Maximum}

Calculating these values for most of the algorithms is trivial but fitness sharing and clearing produce a problem. As fitness sharing and clearing encourage diversity by altering the fitness values of members of the population a normalised fitness must also be recorded at each generation. This is to show an accurate comparison between the quality of solutions found by these algorithms and the ones that do not alter fitness values.

\subsection{Diversity}

The diversity measure that is being taken is moment of inertia diversity, it is calculated using:

$$ I=\sum_{i=1}^{i=L} \sum_{j=1}^{j=P} (x_{ij} - c_i)^2 $$

where $s_{ij}$ is the bit in position $i$ of the $j^{th}$ string and $c_i$ is the $i^{th}$ coordinate of the centroid

$$ c_i=\frac{\sum_{j=1}^{j=P} x_{ij}}{P} $$

\subsection{Offline Performance}

The offline performance represents the algorithms overall performance and is defined as:

$$ offline=\frac{1}{H} \sum_{k=1}^{H} f(BestSoFar_k) $$

Where $BestSoFar$ is the best solution found in the current fitness landscape.

The offline performance shows how well an algorithm can adapt to changes in the environment. As with the minimum, average and maximum measures, offline performance will be calculated using normalised fitness values where necessary.

\chapter{Results}

The Chapter will show the performance of all the algorithms on each of the problems. All algorithms will have a $\mu$ (population size) of 50 and a $\lambda$ (number of offspring) of 30, all individuals will be represented by a   binary string of length 100. The offspring will be generated with crossover percentage of $p(crossover)=0.65$ and mutation percentage $p(mutation) =0.35$. In the island model algorithm the values of $\mu$ and $\lambda$ are divided by the number of islands in order to keep the overall population the same size as the other algorithms. All results are generated with the same random seed and are averaged over 30 runs in order to ensure fair and correct results.

\section{OneMax}

This problem was initialised with a maxima at $\{1,1,\dots\}$. The maximum fitness that can be achieved is 100.

\begin{figure}[h]
	\centering
	\subfloat[Diversity]{\includegraphics[width=0.5\textwidth,height=5cm]{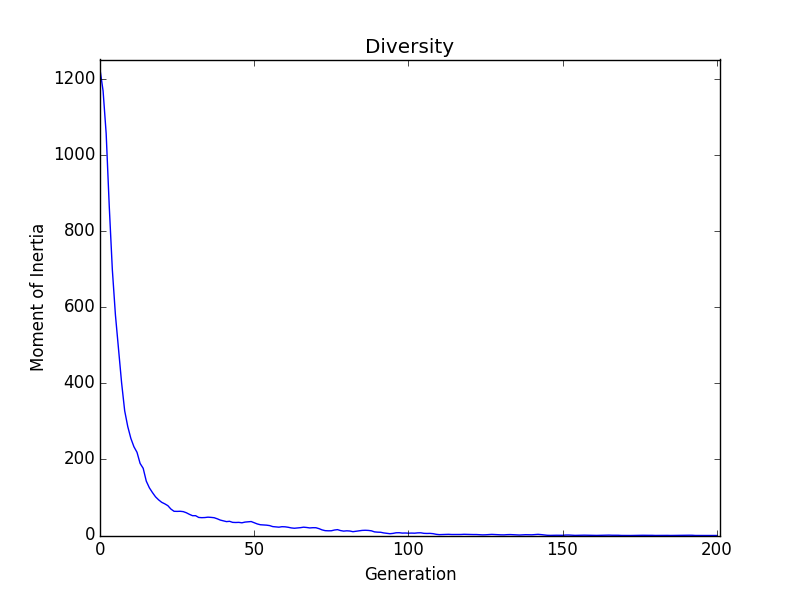}\label{fig:f1}}
	\hfill
	\subfloat[Fitness]{\includegraphics[width=0.5\textwidth,height=5cm]{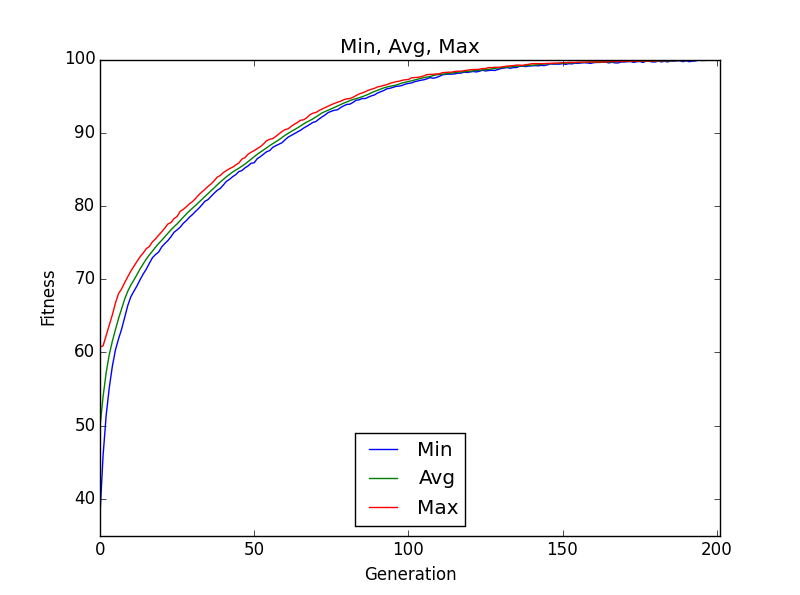}\label{fig:f2}}
	\caption{Performance of Basic on Onemax}
\end{figure}

From the graphs in figure 5.1 we can see the expected solution for $OneMax$. The population diversity is initially high due to the population being made up of randomly generated individuals, it then rapidly decreases as the population converges to the optima. The fitness is initially low, again due to the random initial population, but it steadily increases as the algorithm converges to the peak and then stays at the peak until the end of the run.

\begin{figure}[h]
	\centering
	\subfloat[Diversity]{\includegraphics[width=0.5\textwidth,height=5cm]{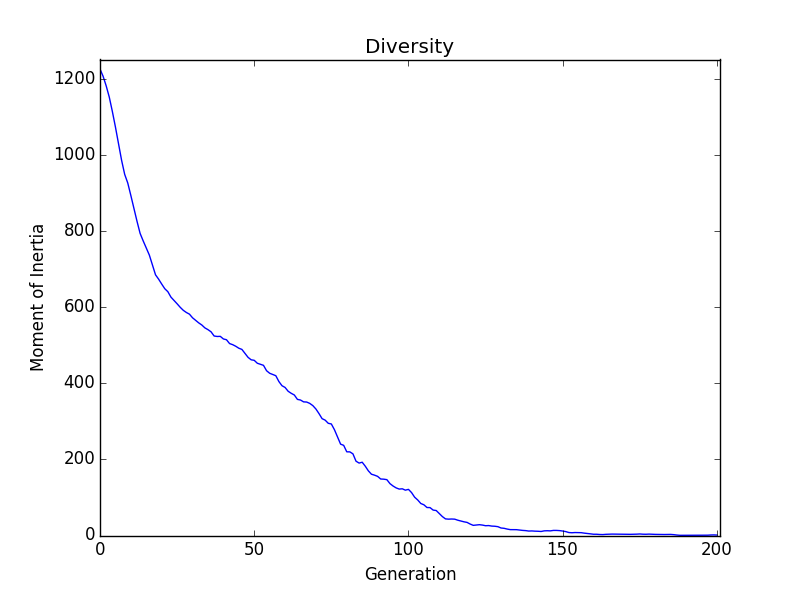}\label{fig:f1}}
	\hfill
	\subfloat[Fitness]{\includegraphics[width=0.5\textwidth,height=5cm]{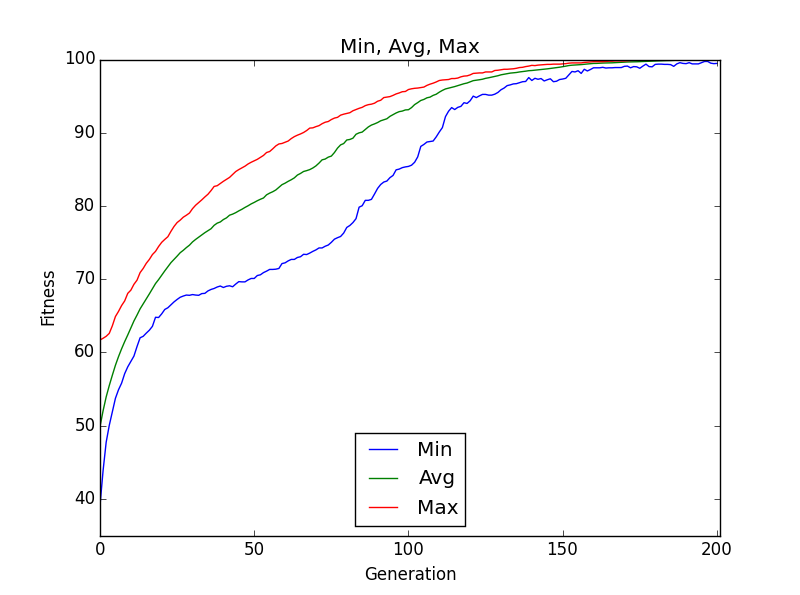}\label{fig:f2}}
	\caption{Performance of Clearing on Onemax}
\end{figure}

\begin{figure}[h]
	\centering
	\subfloat[Diversity]{\includegraphics[width=0.5\textwidth,height=5cm]{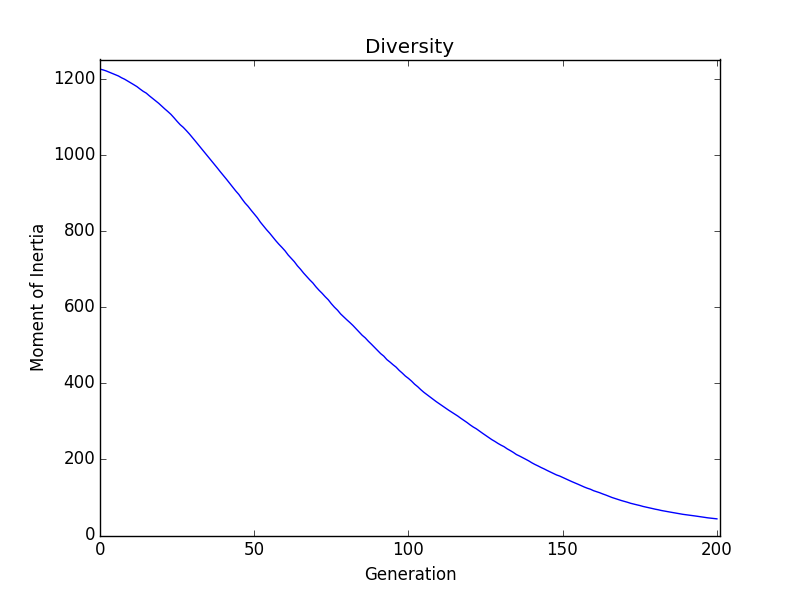}\label{fig:f1}}
	\hfill
	\subfloat[Fitness]{\includegraphics[width=0.5\textwidth,height=5cm]{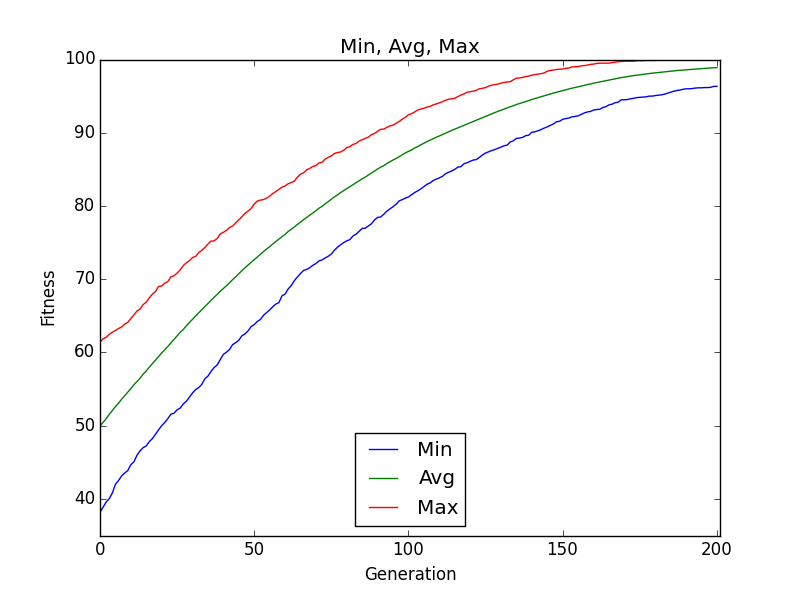}\label{fig:f2}}
	\caption{Performance of Crowding on Onemax}
\end{figure}

\begin{figure}[H]
	\centering
	\subfloat[Diversity]{\includegraphics[width=0.5\textwidth,height=5cm]{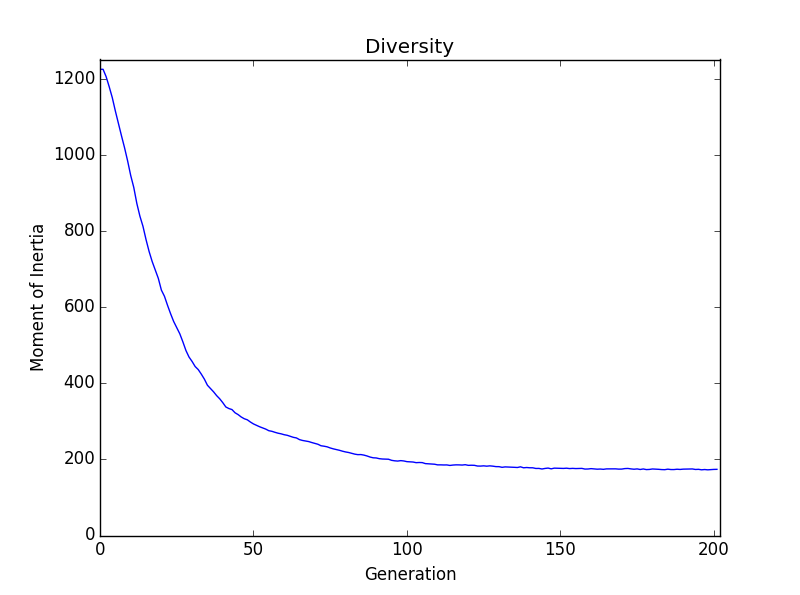}\label{fig:f1}}
	\hfill
	\subfloat[Fitness]{\includegraphics[width=0.5\textwidth,height=5cm]{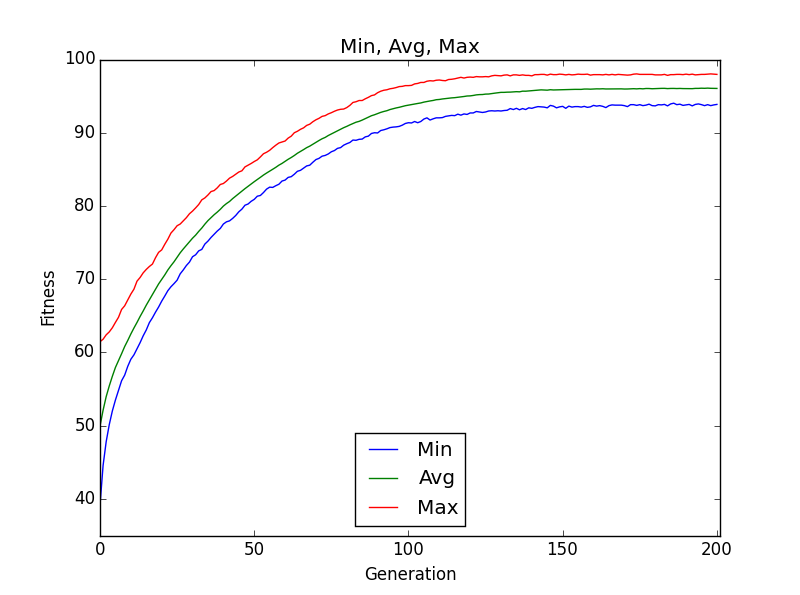}\label{fig:f2}}
	\caption{Performance of Fitness Sharing on Onemax}
\end{figure}

Figure 5.4 and Table 5.1 show that fitness sharing is the only algorithm to fail to find the optima. It gets very near and then levels out just before reaching the peak. This may be due to the nature of fitness sharing, in that as all of the population was in such a small area of the search space the fitness penalties received were so large that the population failed to improve further.

\begin{figure}[H]
	\centering
	\subfloat[Diversity]{\includegraphics[width=0.5\textwidth,height=5cm]{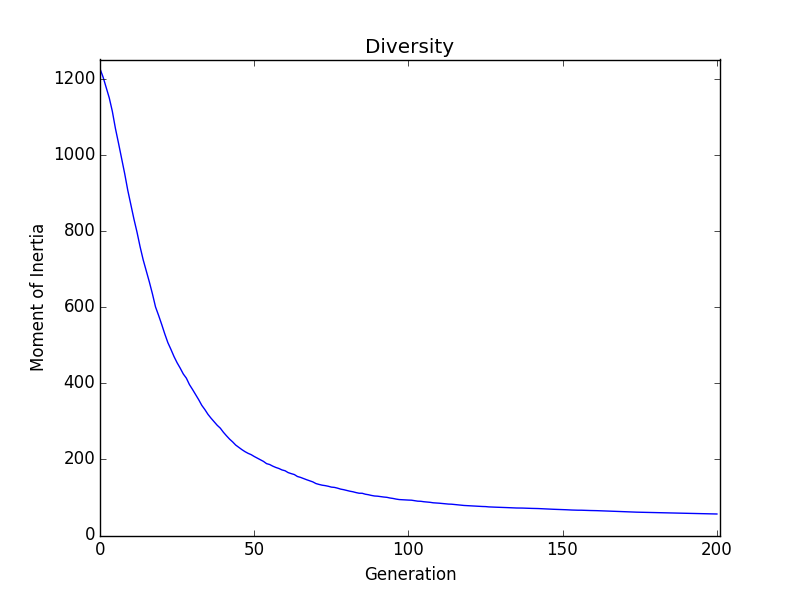}\label{fig:f1}}
	\hfill
	\subfloat[Fitness]{\includegraphics[width=0.5\textwidth,height=5cm]{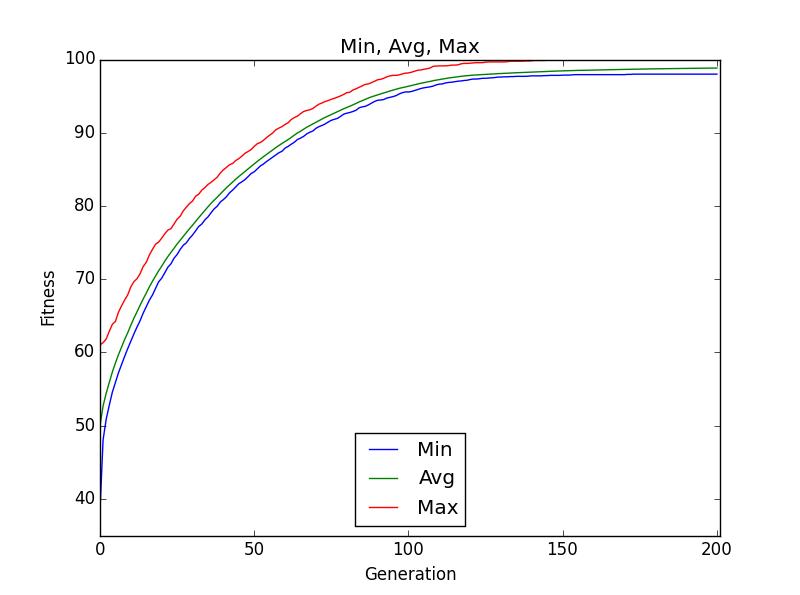}\label{fig:f2}}
	\caption{Performance of Removal of Genotype on Onemax}
\end{figure}

\begin{figure}[H]
	\centering
	\subfloat[Diversity]{\includegraphics[width=0.5\textwidth,height=5cm]{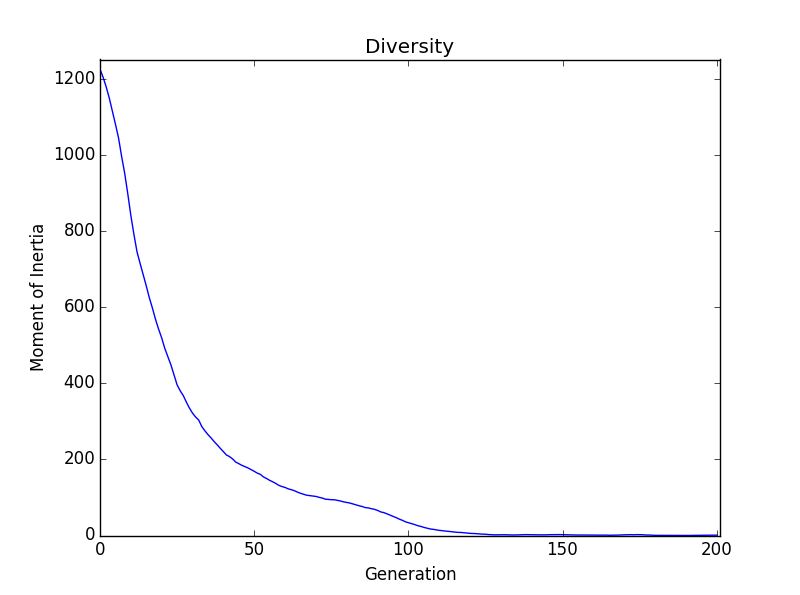}\label{fig:f1}}
	\hfill
	\subfloat[Fitness]{\includegraphics[width=0.5\textwidth,height=5cm]{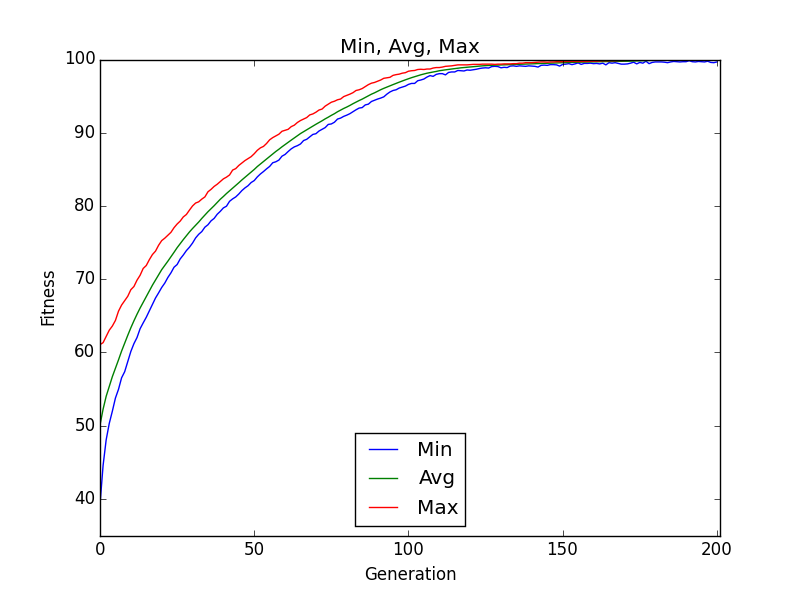}\label{fig:f2}}
	\caption{Performance of Incest Prevention on Onemax}
\end{figure}

\begin{figure}[H]
	\centering
	\subfloat[Diversity]{\includegraphics[width=0.5\textwidth,height=5cm]{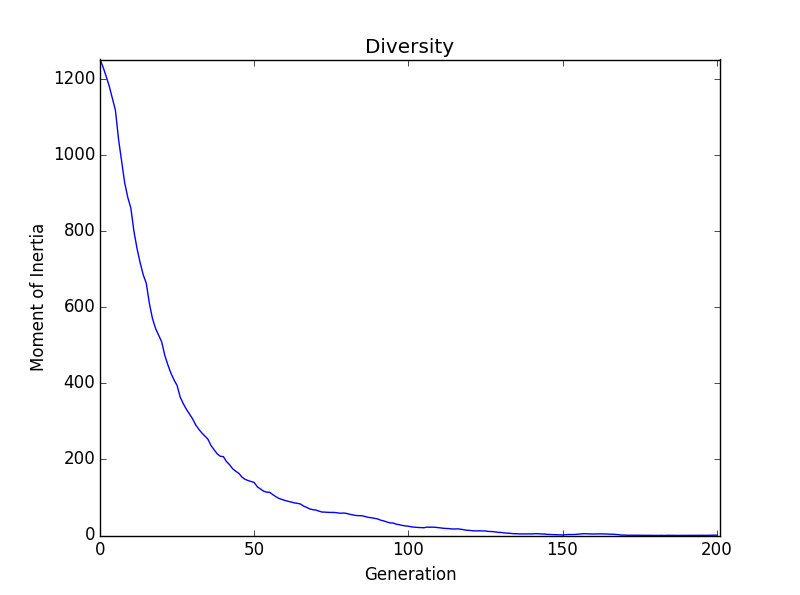}\label{fig:f1}}
	\hfill
	\subfloat[Fitness]{\includegraphics[width=0.5\textwidth,height=5cm]{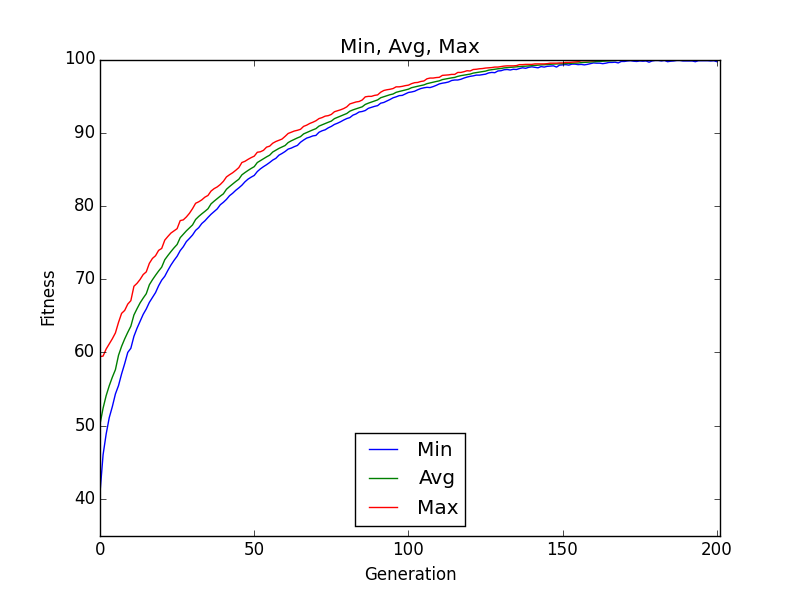}\label{fig:f2}}
	\caption{Performance of Islands Models on Onemax}
\end{figure}

\begin{figure}[H]
	\centering
	\subfloat[Diversity]{\includegraphics[width=0.7\textwidth,height=5cm]{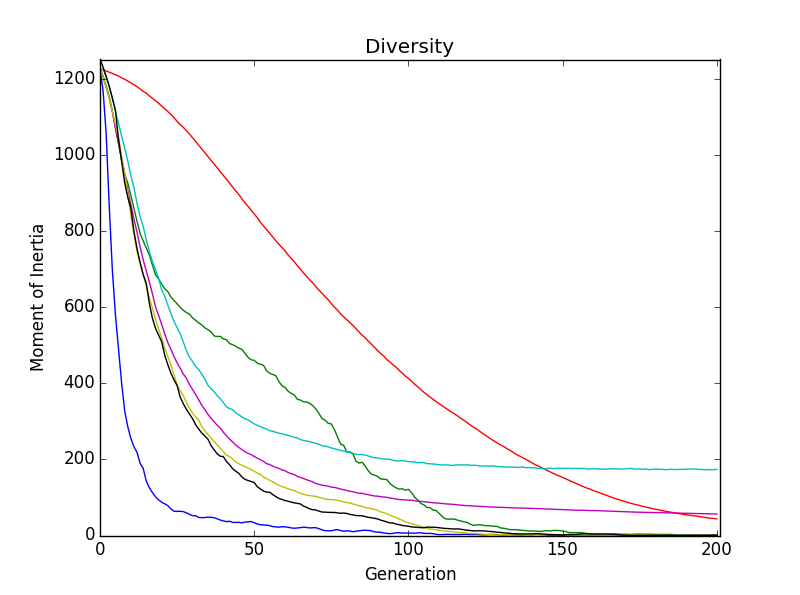}\label{fig:f1}}
%	\hfill
	\subfloat[Legend]{\includegraphics[width=0.3\textwidth,height=5cm]{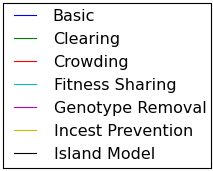}\label{fig:f2}}
	\\
	\hspace*{-4.8cm}
	\subfloat[Fitness]{\includegraphics[width=0.7\textwidth,height=5cm]{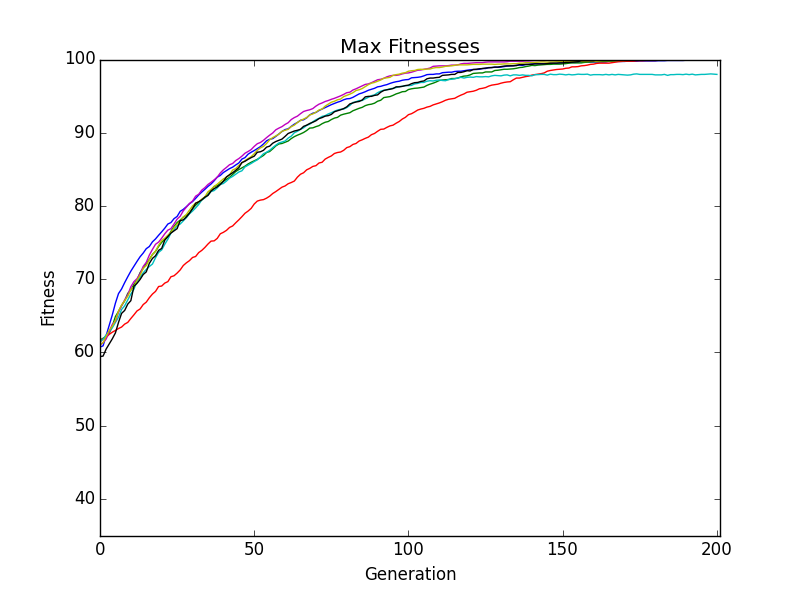}\label{fig:f2}}
	\caption{Performance of All Algorithms on Onemax}
\end{figure}

\begin{center}
	\captionof{table}{Offline Performance and Maximum Achieved Fitness on Onemax}
	\begin{tabular}{| p{2cm} | p{1cm} | p{1.5cm} | p{1.5cm} | p{1.5cm} | p{1.5cm} | p{1.7cm} | p{1.5cm} |} 
		\hline
		Algorithm & Basic & Clearing & Crowding & Fitness Sharing & Genotype Removal & Incest Prevention & Island Model \\ \hline
		Offline Performance & 92.807 & 91.503 & 88.335 & 91.361 & 92.722 & 92.392 & 92.083 \\ \hline
		Maximum Achieved Fitness & 100 & 100 & 100 & 98 & 100 & 100 & 100 \\ \hline
		\hline
	\end{tabular}
\end{center}

This problem can be considered a test problem to ensure that the algorithms are working. As can be seen in the results all of the algorithms are able to locate the maxima except for fitness sharing which came very close. The only difference between the other algorithms is the variation in speed in which each algorithm finds $OneMax$. The basic, genotype removal, incest prevention and island model algorithms show near identical performance and are the fastest to locate the maxima. Clearing and crowding are slightly slower, this is likely due to these algorithms having higher population diversity which caused the slower convergence.

\section{TwoMax}

This problem was initialised with two maxima at $\{0,0,\dots\}$ and $\{1,1,\dots\}$. The maximum fitness that can be achieved is 100 and it can be achieved at either maxima.

\begin{figure}[H]
	\centering
	\subfloat[Diversity]{\includegraphics[width=0.5\textwidth,height=5cm]{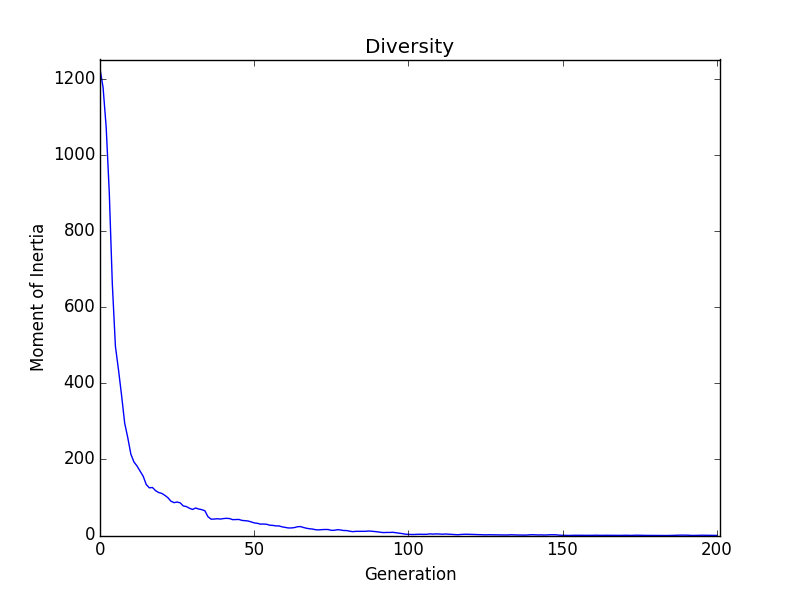}\label{fig:f1}}
	\hfill
	\subfloat[Fitness]{\includegraphics[width=0.5\textwidth,height=5cm]{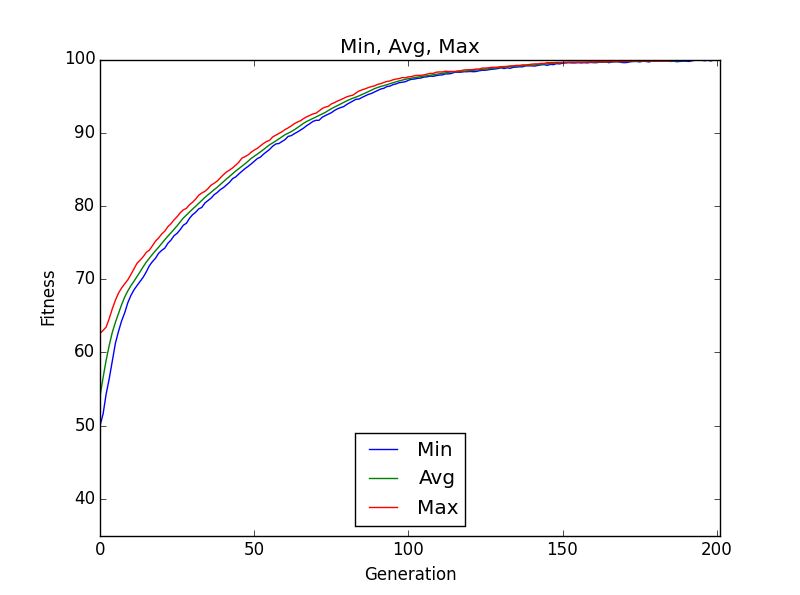}\label{fig:f2}}
	\caption{Performance of Basic on Twomax}
\end{figure}

The results in figure 5.9 show that the basic algorithms performs the same on $TwoMax$ as on $OneMax$. This similar performance is due to the low population diversity which causes convergence to only a single peak.

\begin{figure}[H]
	\centering
	\subfloat[Diversity]{\includegraphics[width=0.5\textwidth,height=5cm]{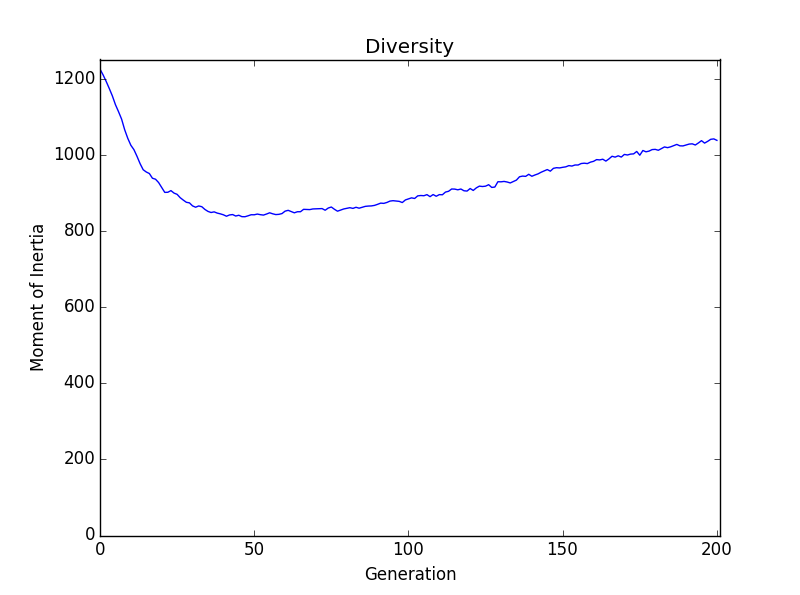}\label{fig:f1}}
	\hfill
	\subfloat[Fitness]{\includegraphics[width=0.5\textwidth,height=5cm]{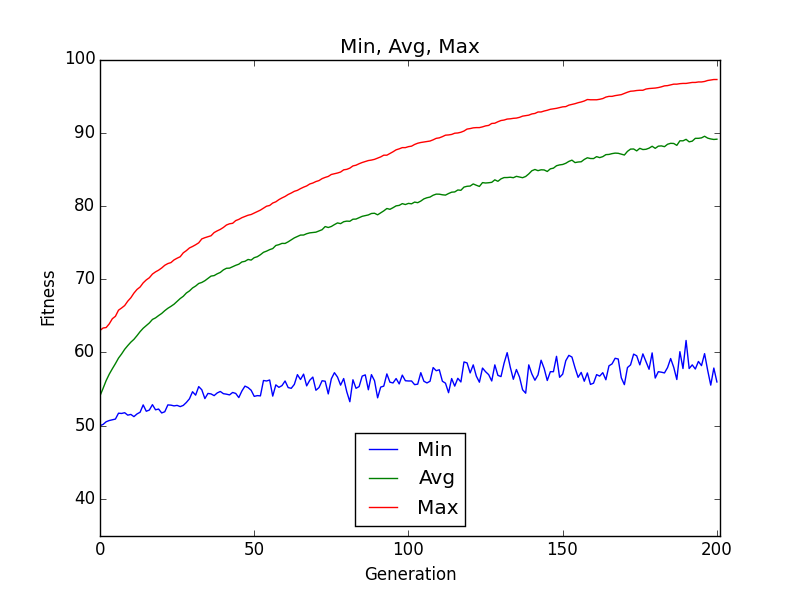}\label{fig:f2}}
	\caption{Performance of Clearing on Twomax}
\end{figure}

The clearing algorithm shows an example of good performance on $TwoMax$ (figure 5.10), the high population diversity allows it to have individuals near both peaks. However, the slower convergence is again an issue as the algorithm fails to locate an exact optima.

\begin{figure}[H]
	\centering
	\subfloat[Diversity]{\includegraphics[width=0.5\textwidth,height=5cm]{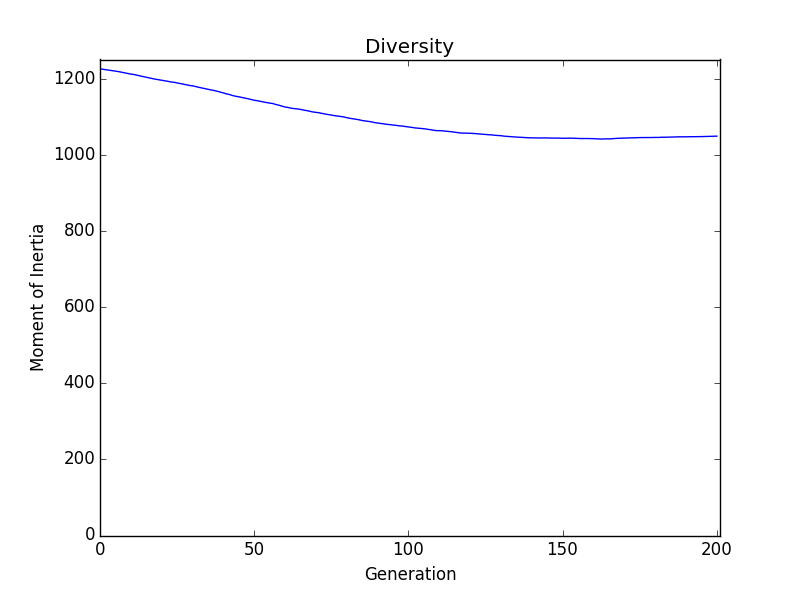}\label{fig:f1}}
	\hfill
	\subfloat[Fitness]{\includegraphics[width=0.5\textwidth,height=5cm]{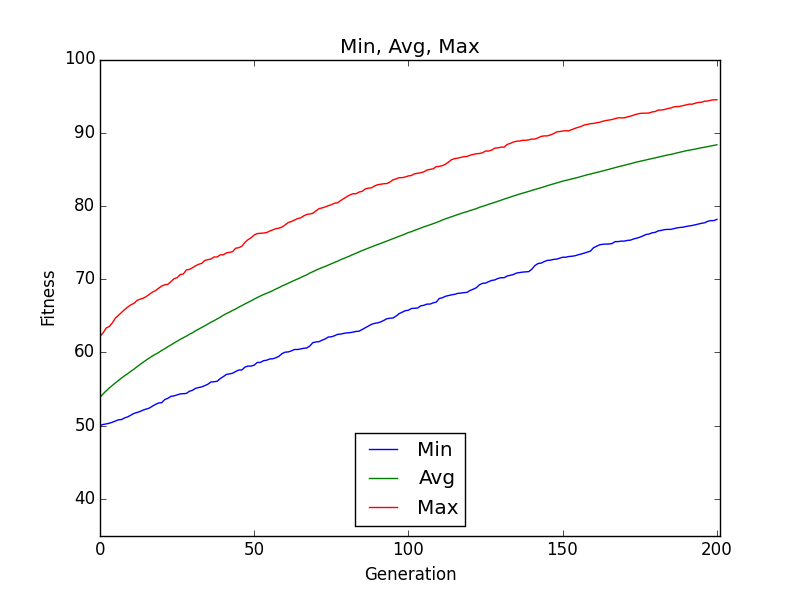}\label{fig:f2}}
	\caption{Performance of Crowding on Twomax}
\end{figure}

\begin{figure}[H]
	\centering
	\subfloat[Diversity]{\includegraphics[width=0.5\textwidth,height=5cm]{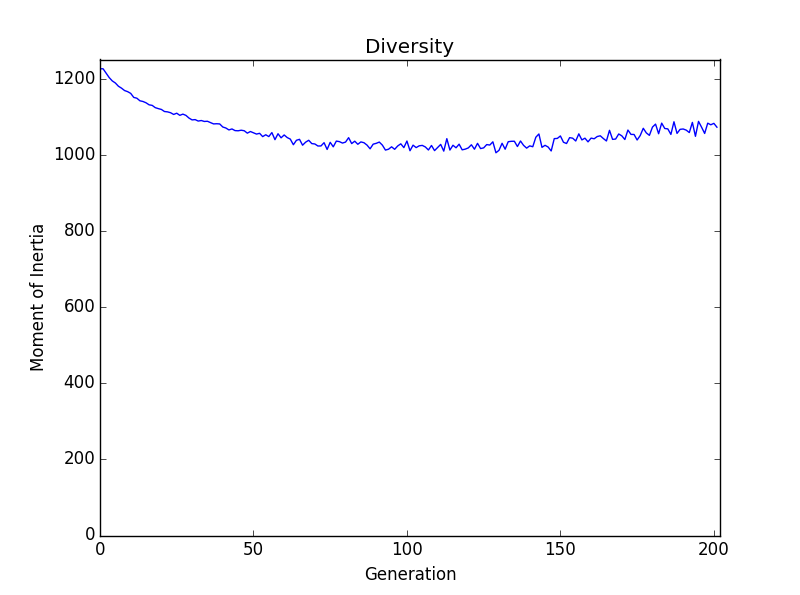}\label{fig:f1}}
	\hfill
	\subfloat[Fitness]{\includegraphics[width=0.5\textwidth,height=5cm]{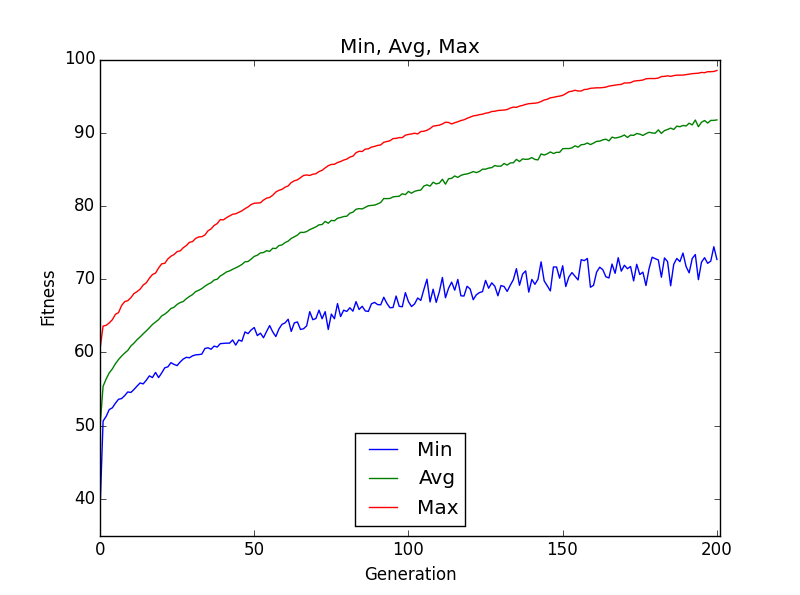}\label{fig:f2}}
	\caption{Performance of Fitness Sharing on Twomax}
\end{figure}

\begin{figure}[H]
	\centering
	\subfloat[Diversity]{\includegraphics[width=0.5\textwidth,height=5cm]{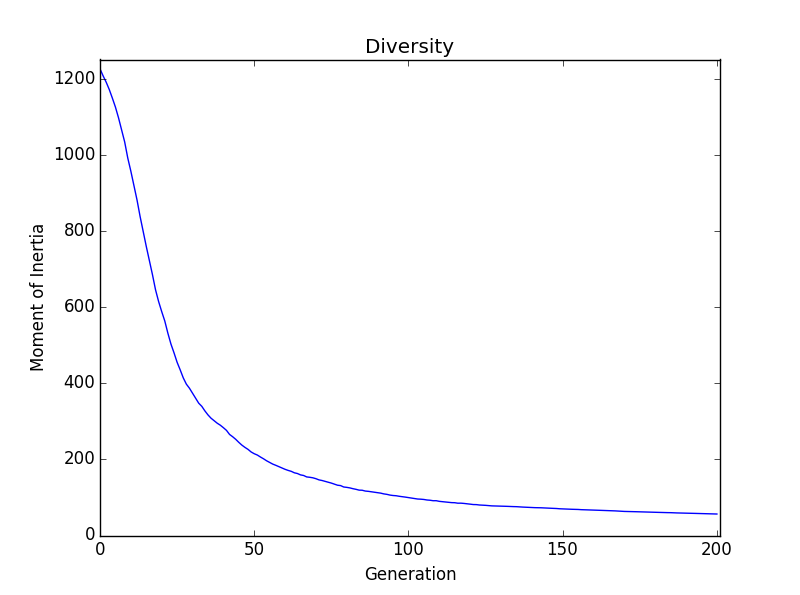}\label{fig:f1}}
	\hfill
	\subfloat[Fitness]{\includegraphics[width=0.5\textwidth,height=5cm]{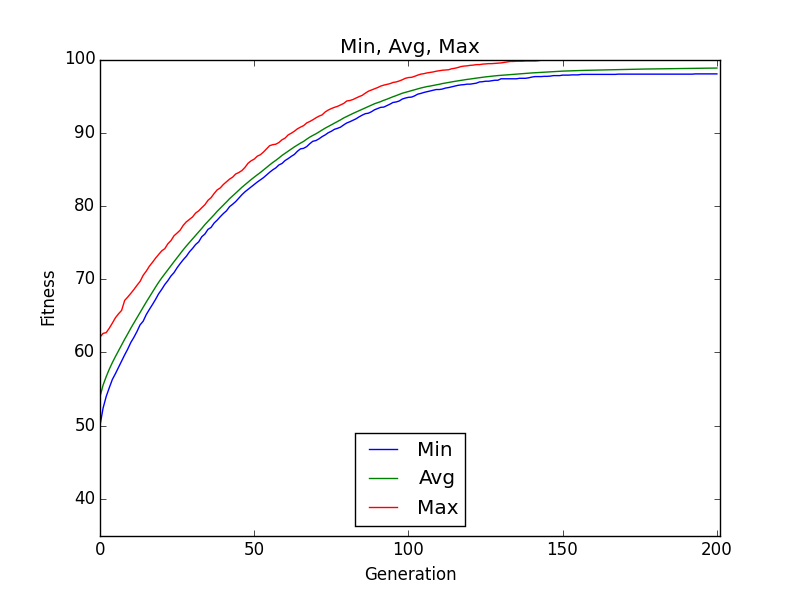}\label{fig:f2}}
	\caption{Performance of Removal of Genotype on Twomax}
\end{figure}

\begin{figure}[H]
	\centering
	\subfloat[Diversity]{\includegraphics[width=0.5\textwidth,height=5cm]{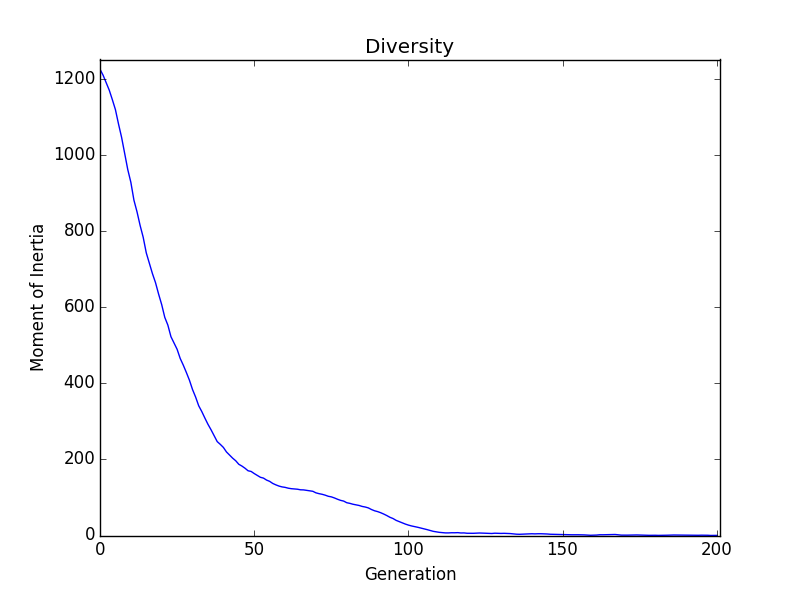}\label{fig:f1}}
	\hfill
	\subfloat[Fitness]{\includegraphics[width=0.5\textwidth,height=5cm]{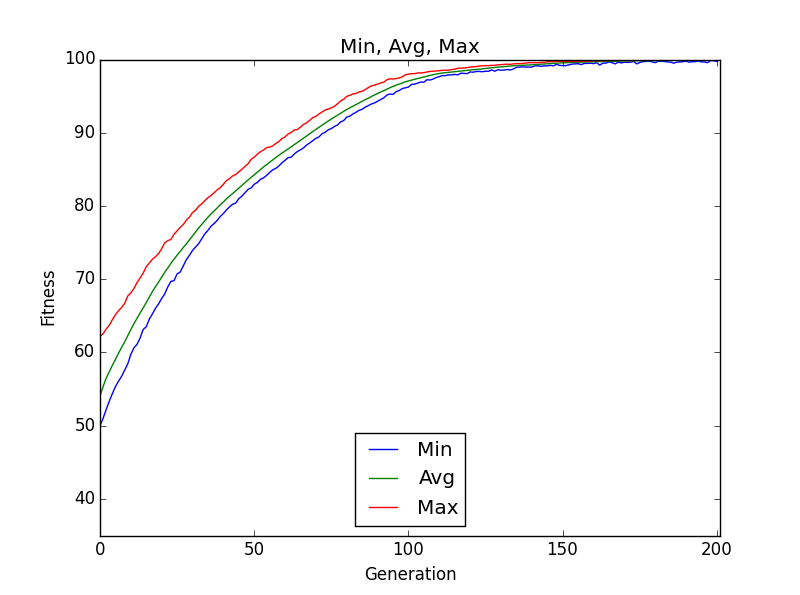}\label{fig:f2}}
	\caption{Performance of Incest Prevention on Twomax}
\end{figure}

\begin{figure}[H]
	\centering
	\subfloat[Diversity]{\includegraphics[width=0.5\textwidth,height=5cm]{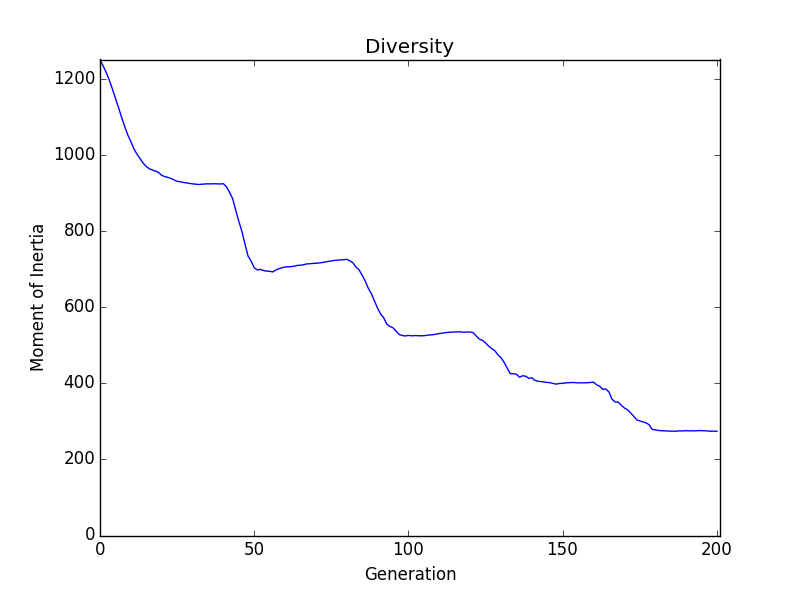}\label{fig:f1}}
	\hfill
	\subfloat[Fitness]{\includegraphics[width=0.5\textwidth,height=5cm]{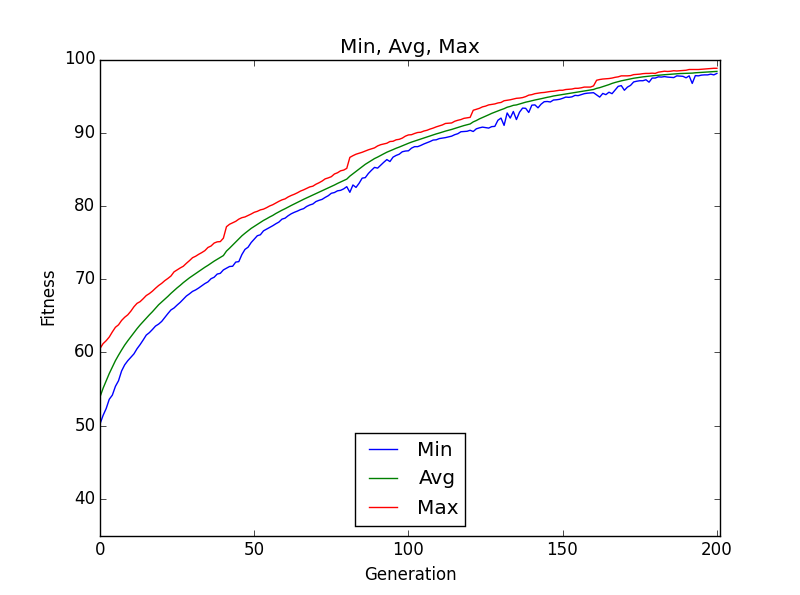}\label{fig:f2}}
	\caption{Performance of Islands Models on Twomax}
\end{figure}

Figure 5.15 shows how migrating individuals between islands affects a populations  diversity and fitness, every time a migration occurs there is a fall in diversity and an increase in fitness.

\begin{figure}[H]
	\centering
	\subfloat[Diversity]{\includegraphics[width=0.7\textwidth,height=5cm]{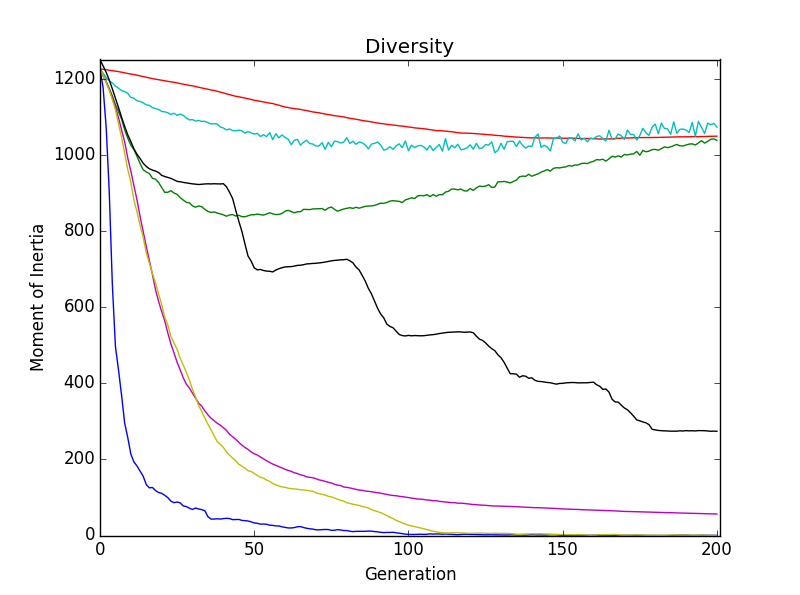}\label{fig:f1}}
	%	\hfill
	\subfloat[Legend]{\includegraphics[width=0.3\textwidth,height=5cm]{figures/legend.png}\label{fig:f2}}
	\\
	\hspace*{-4.8cm}
	\subfloat[Fitness]{\includegraphics[width=0.7\textwidth,height=5cm]{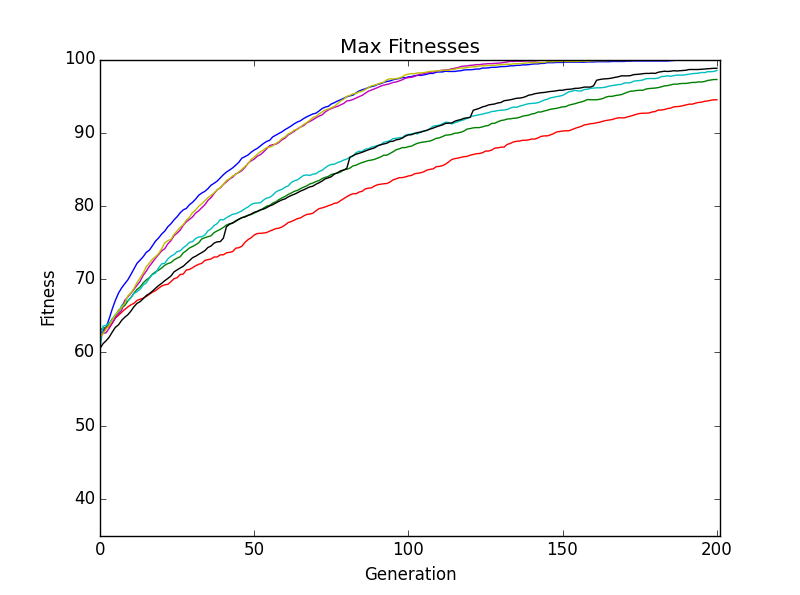}\label{fig:f2}}
	\caption{Performance of All Algorithms on Twomax}
\end{figure}

\begin{center}
	\captionof{table}{Offline Performance and Maximum Achieved Fitness on Twomax}
	\begin{tabular}{| p{2cm} | p{1cm} | p{1.5cm} | p{1.5cm} | p{1.5cm} | p{1.5cm} | p{1.7cm} | p{1.5cm} |}
		\hline
		Algorithm & Basic & Clearing & Crowding & Fitness Sharing & Genotype Removal & Incest Prevention & Island Model \\ \hline
		Offline Performance & 92.558 & 85.865 & 82.5695 & 87.276 & 91.956 & 92.091 & 87.783 \\ \hline
		Maximum Achieved Fitness & 100 & 97 & 95 & 99 & 100 & 100 & 99 \\ \hline
		Finds Both Peaks & No & Yes & Yes & Yes & No & No & Partially \\ \hline
		\hline
	\end{tabular}
\end{center}

$TwoMax$ is a test of how well each algorithm can promote diversity in a population. The algorithms (clearing, crowding and fitness sharing) that performed worse on $OneMax$ are the only algorithms capable of finding both peaks on the problem. The island model was the best of the rest as it maintained diversity for a longer period. However it eventually converged to a single peak. 

The problem of slower convergence that was demonstrated by these algorithms on $OneMax$ is still visible here. Although the algorithms were capable of maintaining subpopulations near both peaks they failed to converge to an optima during the run.

The other algorithms: basic, genotype removal and incest prevention were successful at finding an optima. However, this is not considered a success as the aim of $TwoMax$ is to locate both peaks.

These differences are clearly visible on the graphs in figure 5.16, the algorithms with higher diversity have lower maximum fitness, and vice versa.

\section{One Moving Peak}

This problem was initialised with one peak at $\{0,0,\dots\}$ with height 100. At generations $150$ and $300$ the peak moves using a bit flip function with probability of flipping $p=0.1$ but the height remains constant. 

\begin{figure}[H]
	\centering
	\subfloat[Diversity]{\includegraphics[width=0.5\textwidth,height=5cm]{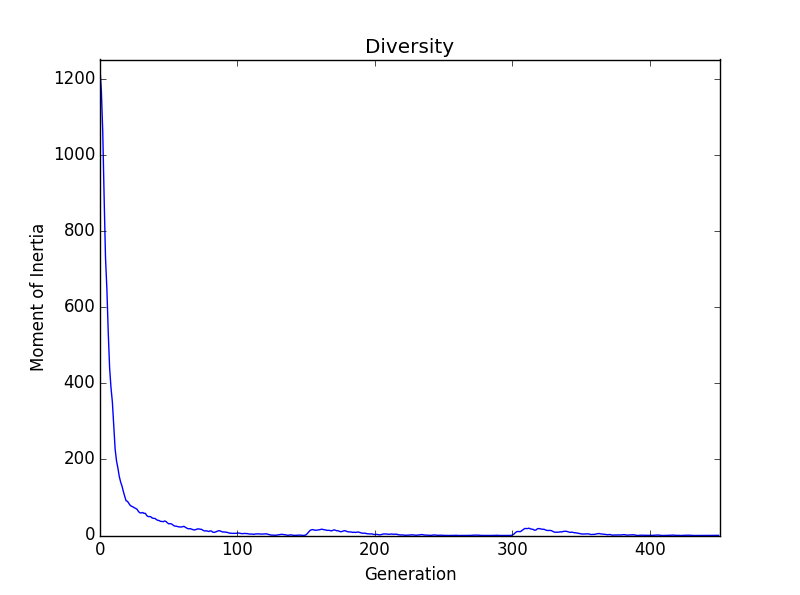}\label{fig:f1}}
	\hfill
	\subfloat[Fitness]{\includegraphics[width=0.5\textwidth,height=5cm]{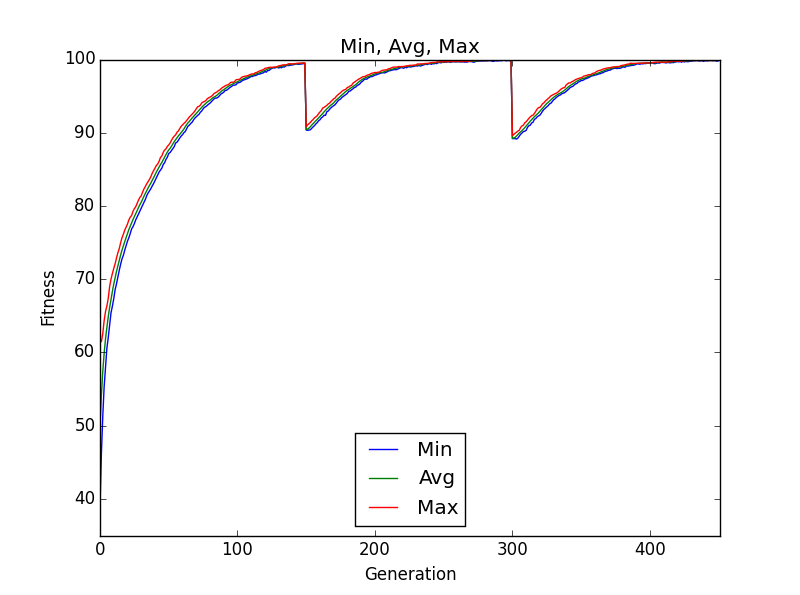}\label{fig:f2}}
	\caption{Performance of Basic on One Moving Peak}
\end{figure}

In the graphs in figure 5.17 the effects of moving the peak are clearly visible. The algorithm converges to the peak as in previous problems and when the peak moves there is a drop in fitness. Every time the peak moves the algorithm reconverges too the new peak location.

\begin{figure}[H]
	\centering
	\subfloat[Diversity]{\includegraphics[width=0.5\textwidth,height=5cm]{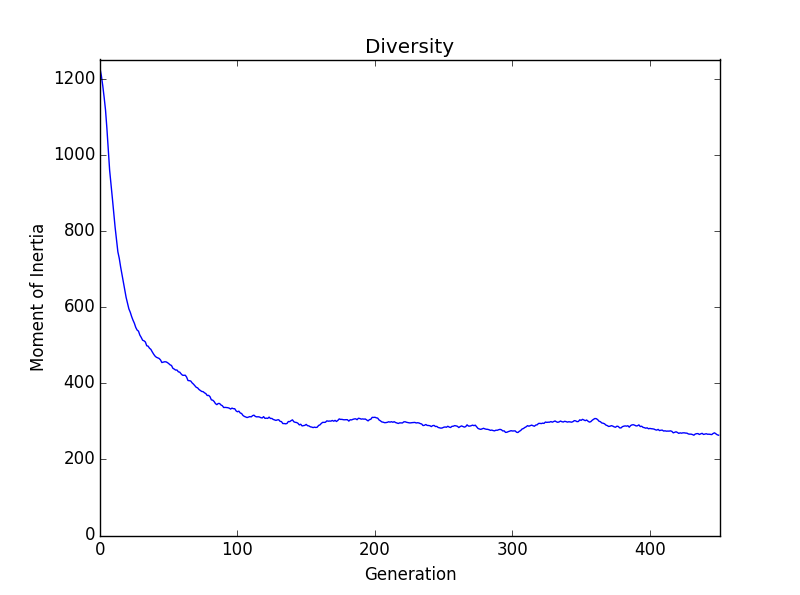}\label{fig:f1}}
	\hfill
	\subfloat[Fitness]{\includegraphics[width=0.5\textwidth,height=5cm]{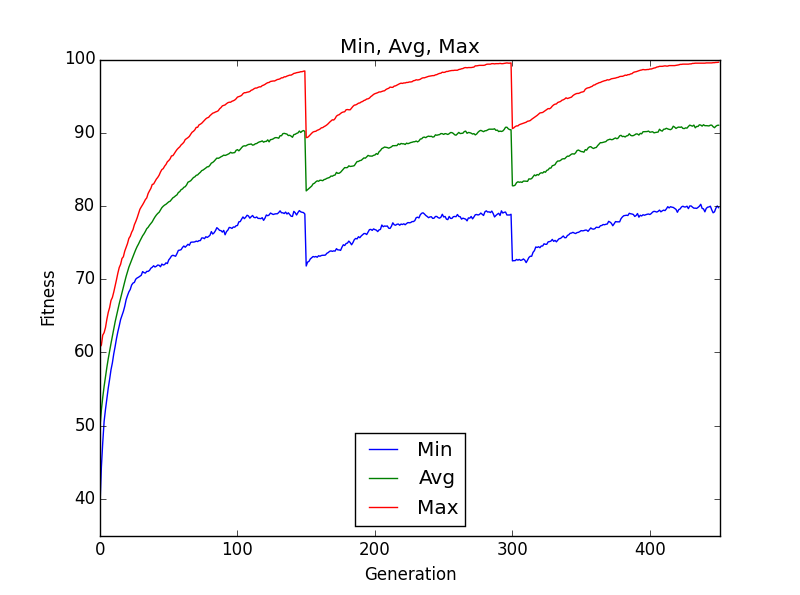}\label{fig:f2}}
	\caption{Performance of Clearing on One Moving Peak}
\end{figure}

\begin{figure}[H]
	\centering
	\subfloat[Diversity]{\includegraphics[width=0.5\textwidth,height=5cm]{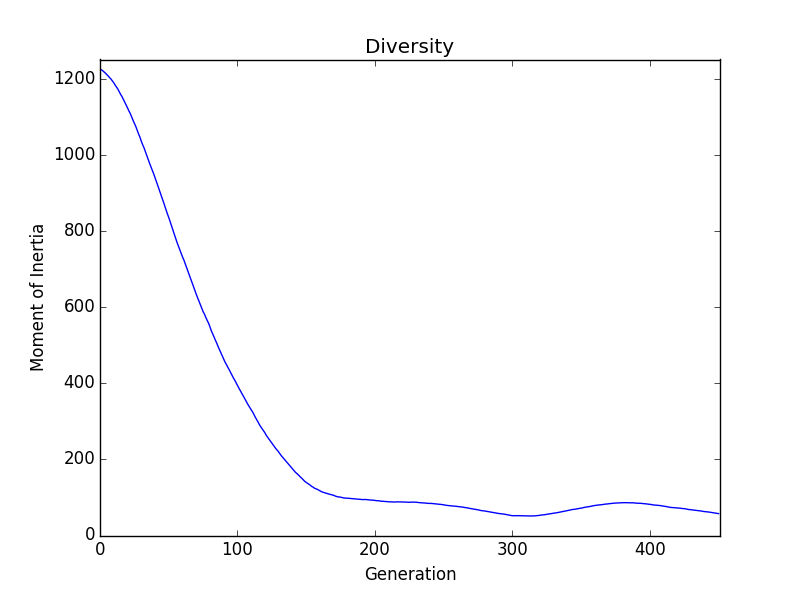}\label{fig:f1}}
	\hfill
	\subfloat[Fitness]{\includegraphics[width=0.5\textwidth,height=5cm]{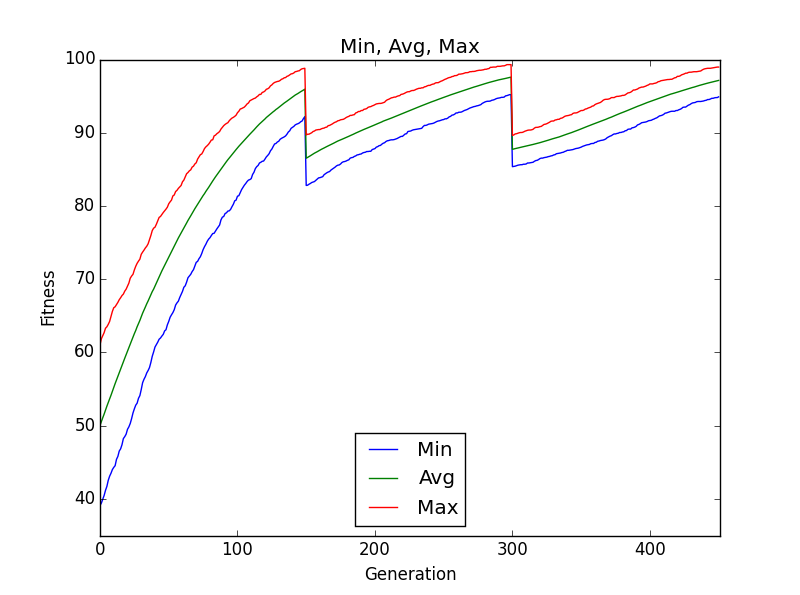}\label{fig:f2}}
	\caption{Performance of Crowding on One Moving Peak}
\end{figure}

\begin{figure}[H]
	\centering
	\subfloat[Diversity]{\includegraphics[width=0.5\textwidth,height=5cm]{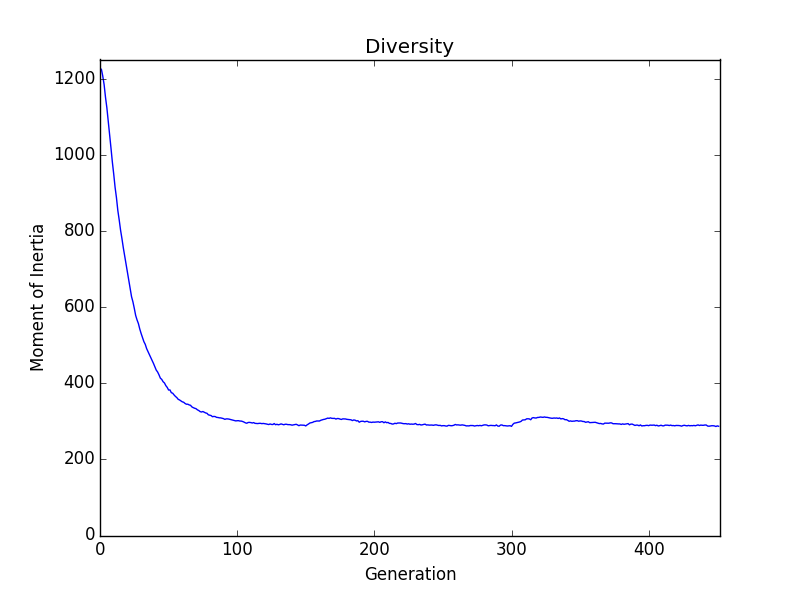}\label{fig:f1}}
	\hfill
	\subfloat[Fitness]{\includegraphics[width=0.5\textwidth,height=5cm]{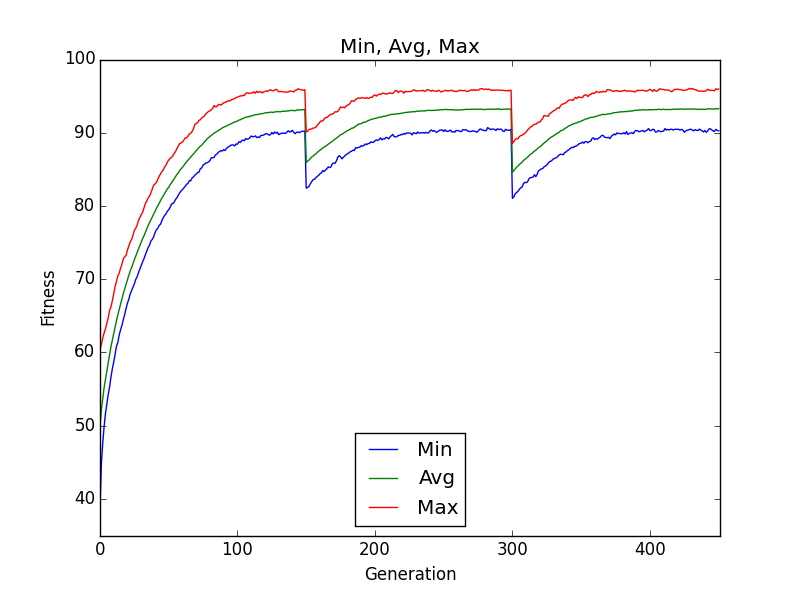}\label{fig:f2}}
	\caption{Performance of Fitness Sharing on One Moving Peak}
\end{figure}

Fitness sharing is again the only algorithm to fail to find the exact peak (figure 5.20 (b)), this is probably due to it having a higher level of diversity compared to the other algorithms. However this diversity does seem to allow it to have smaller fitness drops and recover more quickly than the other algorithms when the peak moves.

\begin{figure}[H]
	\centering
	\subfloat[Diversity]{\includegraphics[width=0.5\textwidth,height=5cm]{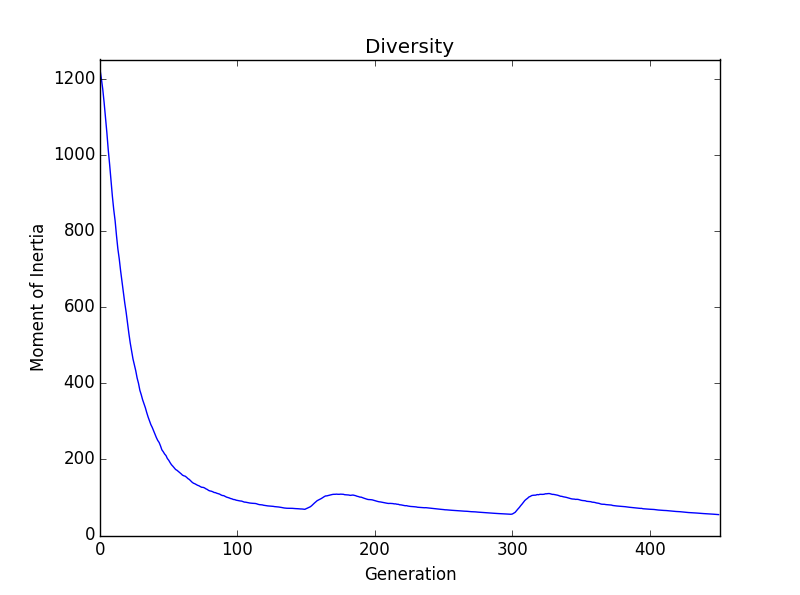}\label{fig:f1}}
	\hfill
	\subfloat[Fitness]{\includegraphics[width=0.5\textwidth,height=5cm]{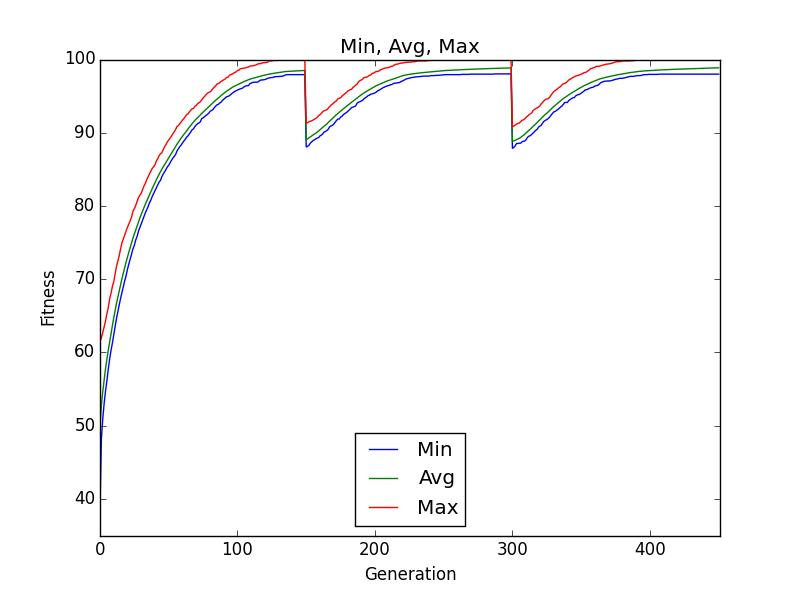}\label{fig:f2}}
	\caption{Performance of Removal of Genotype on One Moving Peak}
\end{figure}

\begin{figure}[H]
	\centering
	\subfloat[Diversity]{\includegraphics[width=0.5\textwidth,height=5cm]{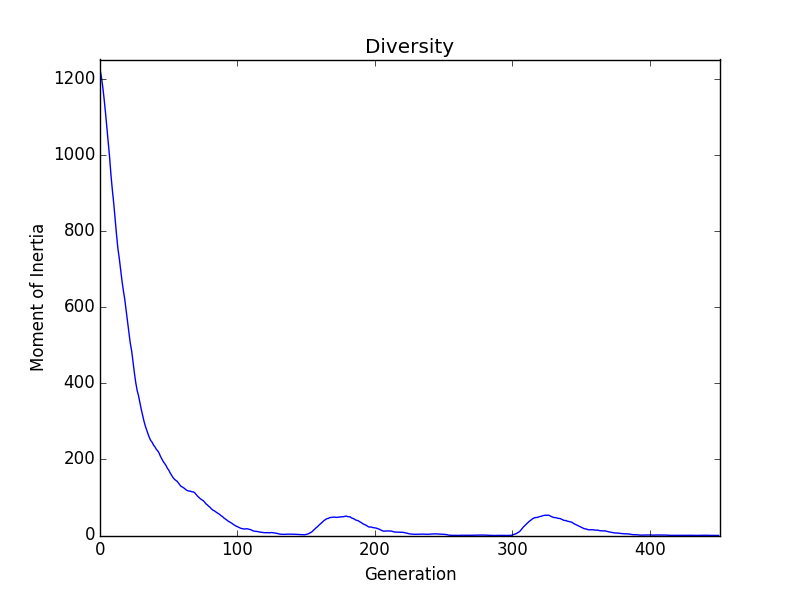}\label{fig:f1}}
	\hfill
	\subfloat[Fitness]{\includegraphics[width=0.5\textwidth,height=5cm]{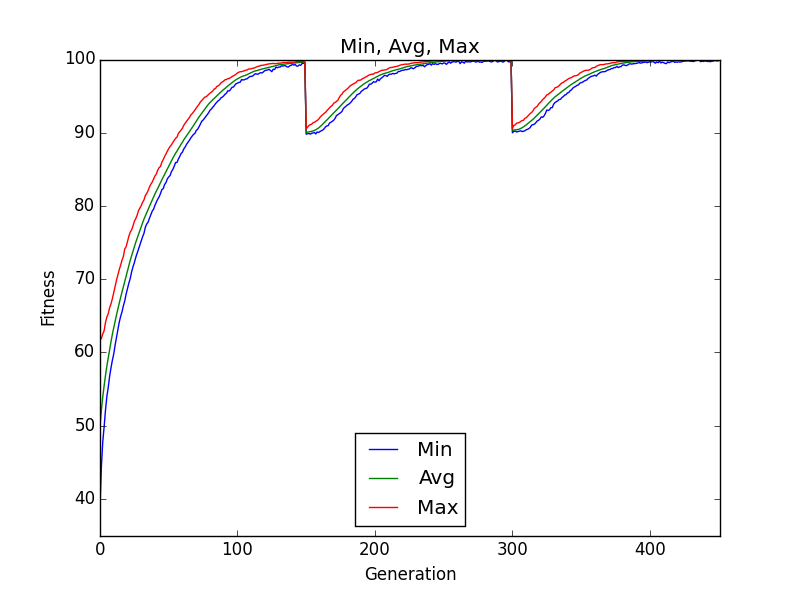}\label{fig:f2}}
	\caption{Performance of Incest Prevention on One Moving Peak}
\end{figure}

\begin{figure}[H]
	\centering
	\subfloat[Diversity]{\includegraphics[width=0.5\textwidth,height=5cm]{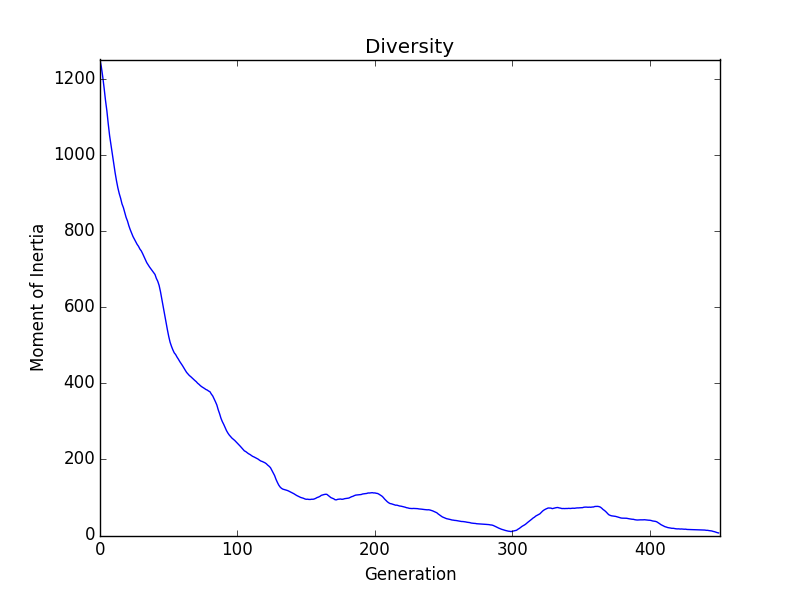}\label{fig:f1}}
	\hfill
	\subfloat[Fitness]{\includegraphics[width=0.5\textwidth,height=5cm]{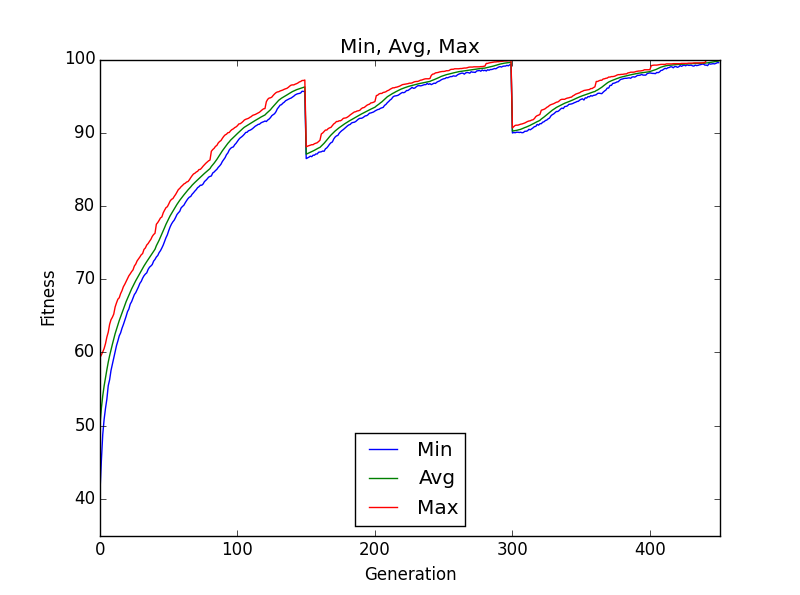}\label{fig:f2}}
	\caption{Performance of Islands Models on One Moving Peak}
\end{figure}

\begin{figure}[H]
	\centering
	\subfloat[Diversity]{\includegraphics[width=0.7\textwidth,height=5cm]{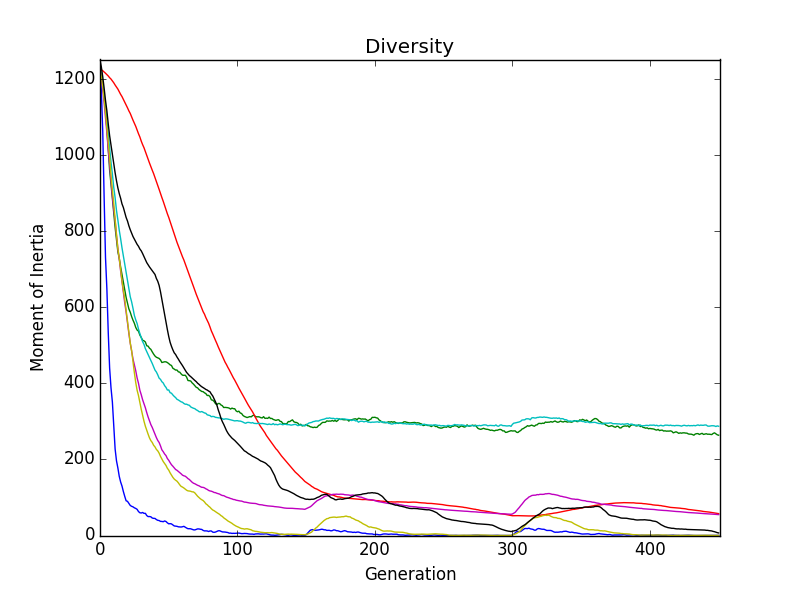}\label{fig:f1}}
	%	\hfill
	\subfloat[Legend]{\includegraphics[width=0.3\textwidth,height=5cm]{figures/legend.png}\label{fig:f2}}
	\\
	\hspace*{-4.8cm}
	\subfloat[Fitness]{\includegraphics[width=0.7\textwidth,height=5cm]{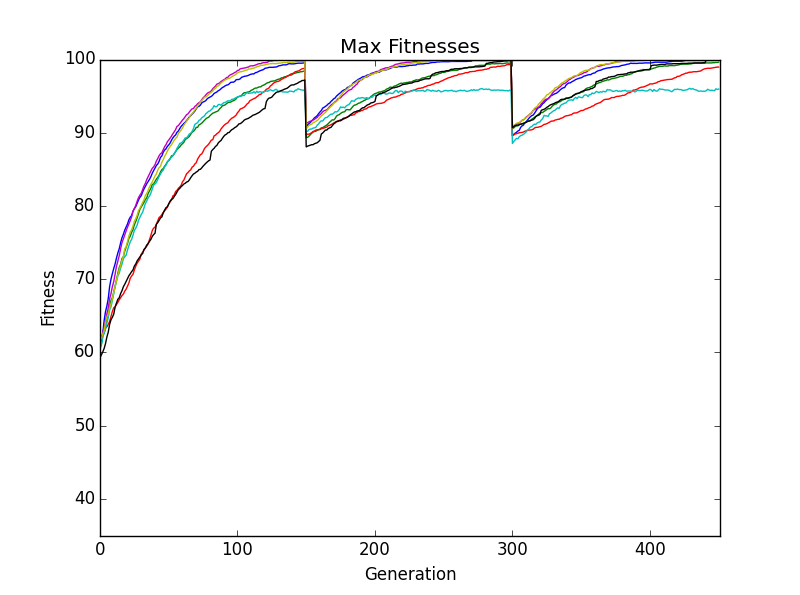}\label{fig:f2}}
	\caption{Performance of All Algorithms on One Moving Peak}
\end{figure}

\newpage
\begin{center}
	\captionof{table}{Offline Performance and Maximum Achieved Fitness on One Moving Peak}
	\begin{tabular}{| p{2cm} | p{1cm} | p{1.5cm} | p{1.5cm} | p{1.5cm} | p{1.5cm} | p{1.7cm} | p{1.5cm} |}
		\hline
		Algorithm & Basic & Clearing & Crowding & Fitness Sharing & Genotype Removal & Incest Prevention & Island Model \\ \hline
		Offline Performance & 95.384 & 93.647 & 91.613 & 93.087 & 95.604 & 95.358 & 92.939 \\ \hline
		Maximum Achieved Fitness & 100\newline 100\newline 100 & 98\newline 100\newline 100 & 99\newline 99\newline 99 & 96\newline 96\newline 96 & 100\newline 100\newline 100& 100\newline 100\newline 100& 97\newline 100\newline 100 \\ \hline
		\hline
	\end{tabular}
\end{center}
The maximum achieved fitness has a value for each time the peaks move separated by line breaks (this will be the same for all future results tables).

This was the first dynamic test the algorithms faced and all of them were able to successfully track the peak as it moves. Looking at figure 5.24 (a) where all of the algorithms diversities are plotted it seems that this problem split the algorithms into three groups. Basic, genotype removal and incest prevention have a sharp fall in diversity and then remain at a low diversity for the rest of the run with small increases when the peak moves. 

The second group consists of fitness sharing and clearing, initially these algorithms have a decrease at a comparably rate to the first group. However, they level off earlier and stay at a higher population diversity. With fitness sharing this diversity allows for small fitness drops and faster reconvergence but this is not the case in clearing. This is likely due to a combination of fitness sharing not reaching the optima and of how fitness sharing would keep all individuals near the peak but with fitness penalties. In clearing if too much of the population is near the peak many individuals will have 0 fitness, this seems to lead to having less useful diversity on a single peak. 

The last group consisting of crowding and island model have a slower fall in diversity than the previous algorithms, but eventually drop to the same diversity levels as the algorithms in the first group.\newpage

\section{Two Moving Peaks}

This problem was initialised with two peaks at $\{0,0,\dots\}$ with height 100 and $\{1,1,\dots\}$ with height 90. At generations $150$ and $300$ both peaks move using a bit flip function with probability of flipping $p=0.1$ but their heights remain constant. This means that peak 0 is always the global maxima. 

\begin{figure}[H]
	\centering
	\subfloat[Diversity]{\includegraphics[width=0.5\textwidth,height=5cm]{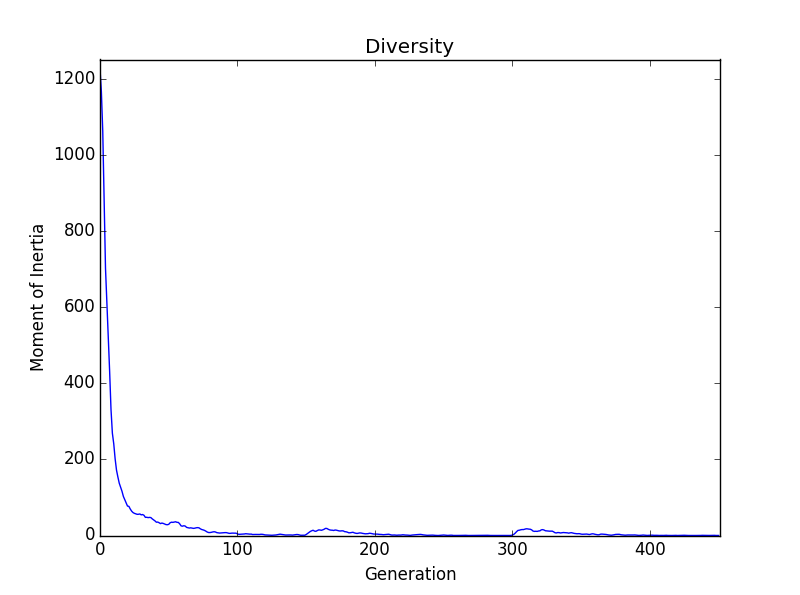}\label{fig:f1}}
	\hfill
	\subfloat[Fitness]{\includegraphics[width=0.5\textwidth,height=5cm]{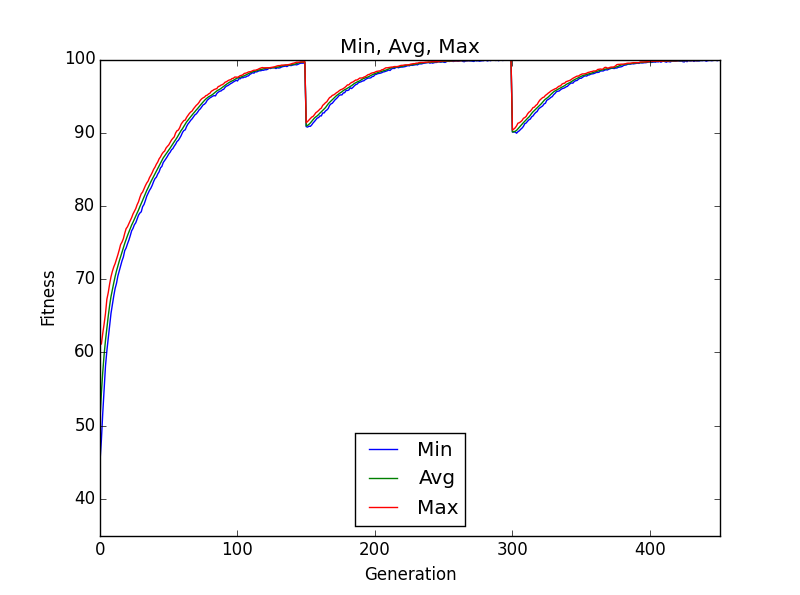}\label{fig:f2}}
	\caption{Performance of Basic on Two Moving Peaks}
\end{figure}

Figure 5.25 shows that as with $TwoMax$ the basic algorithm fails to find both peaks and behaves just as it did on the single moving peak problem (figure 5.17).

\begin{figure}[H]
	\centering
	\subfloat[Diversity]{\includegraphics[width=0.5\textwidth,height=5cm]{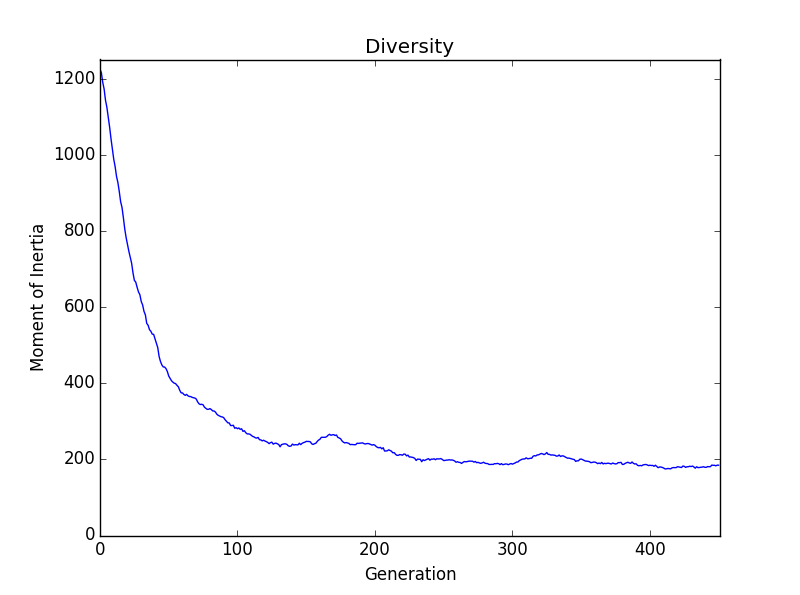}\label{fig:f1}}
	\hfill
	\subfloat[Fitness]{\includegraphics[width=0.5\textwidth,height=5cm]{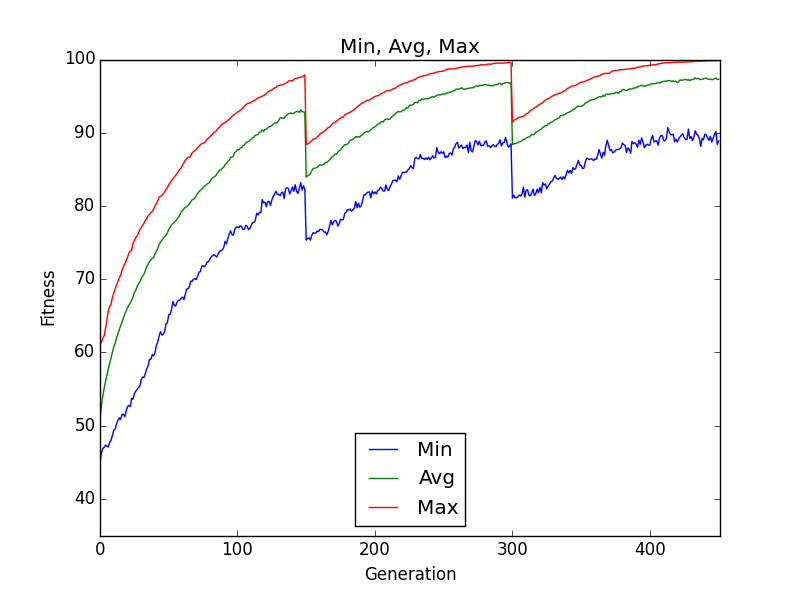}\label{fig:f2}}
	\caption{Performance of Clearing on Two Moving Peaks}
\end{figure}

\begin{figure}[H]
	\centering
	\subfloat[Diversity]{\includegraphics[width=0.5\textwidth,height=5cm]{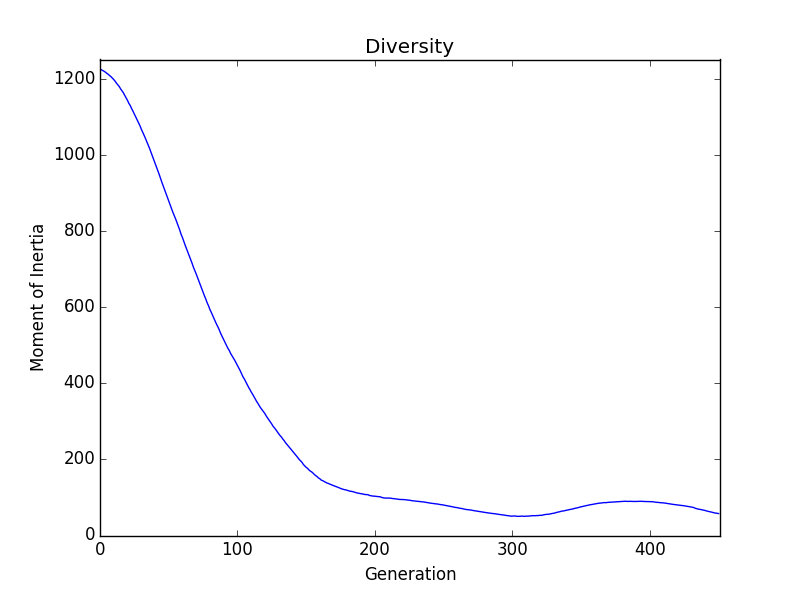}\label{fig:f1}}
	\hfill
	\subfloat[Fitness]{\includegraphics[width=0.5\textwidth,height=5cm]{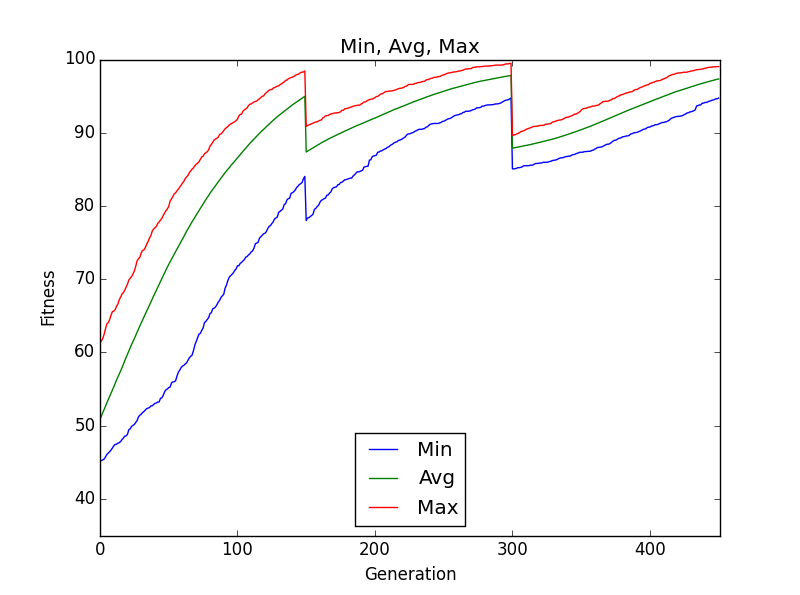}\label{fig:f2}}
	\caption{Performance of Crowding on Two Moving Peaks}
\end{figure}

\begin{figure}[H]
	\centering
	\subfloat[Diversity]{\includegraphics[width=0.5\textwidth,height=5cm]{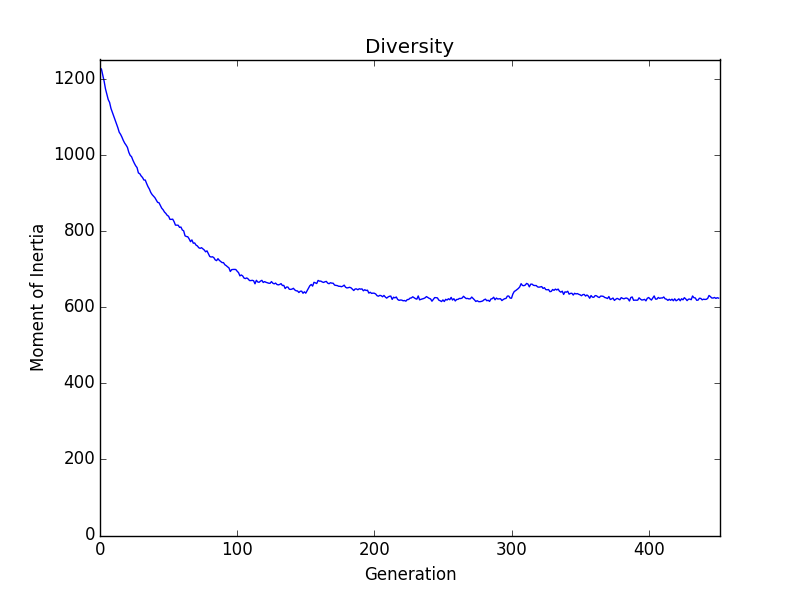}\label{fig:f1}}
	\hfill
	\subfloat[Fitness]{\includegraphics[width=0.5\textwidth,height=5cm]{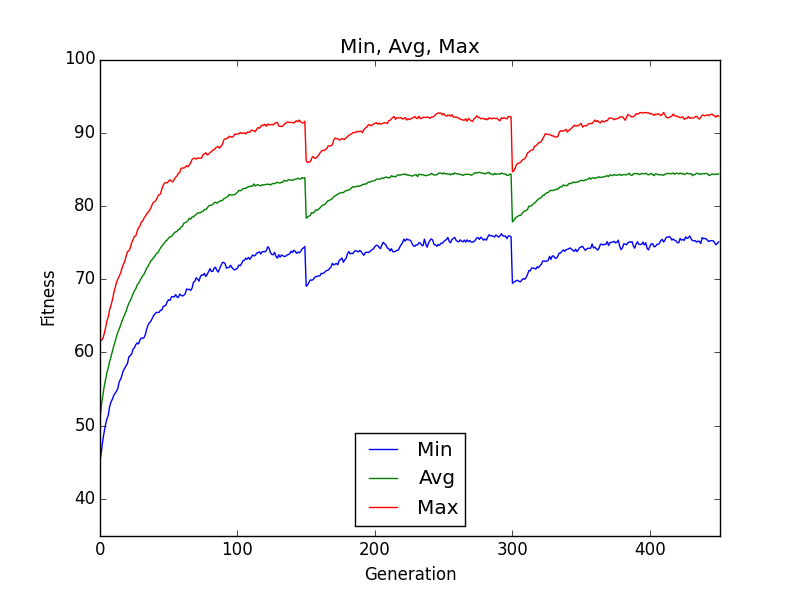}\label{fig:f2}}
	\caption{Performance of Fitness Sharing on Two Moving Peaks}
\end{figure}

In figure 5.28 we can see that fitness sharing maintains a higher diversity than the other algorithms on this problem. Unfortunately this is not due to both peaks being found but due to a wide variety of individuals all on the higher peak. 

\begin{figure}[H]
	\centering
	\subfloat[Diversity]{\includegraphics[width=0.5\textwidth,height=5cm]{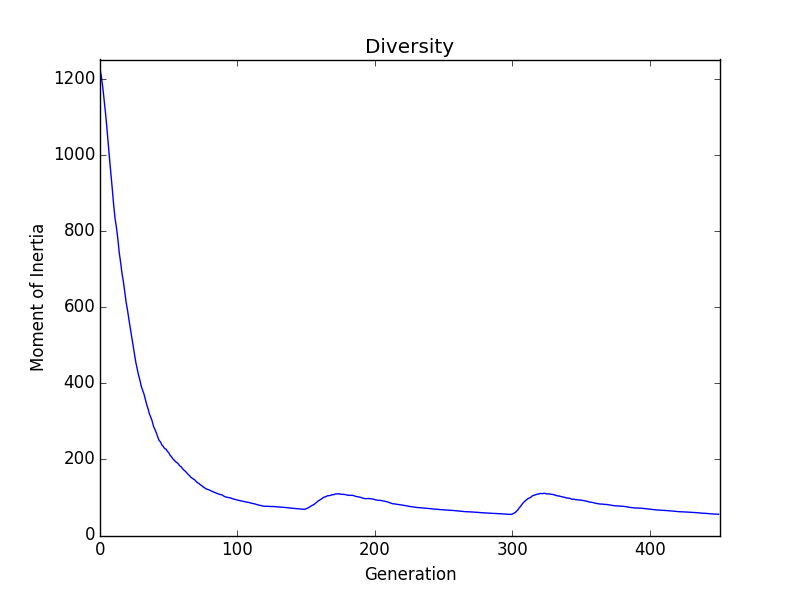}\label{fig:f1}}
	\hfill
	\subfloat[Fitness]{\includegraphics[width=0.5\textwidth,height=5cm]{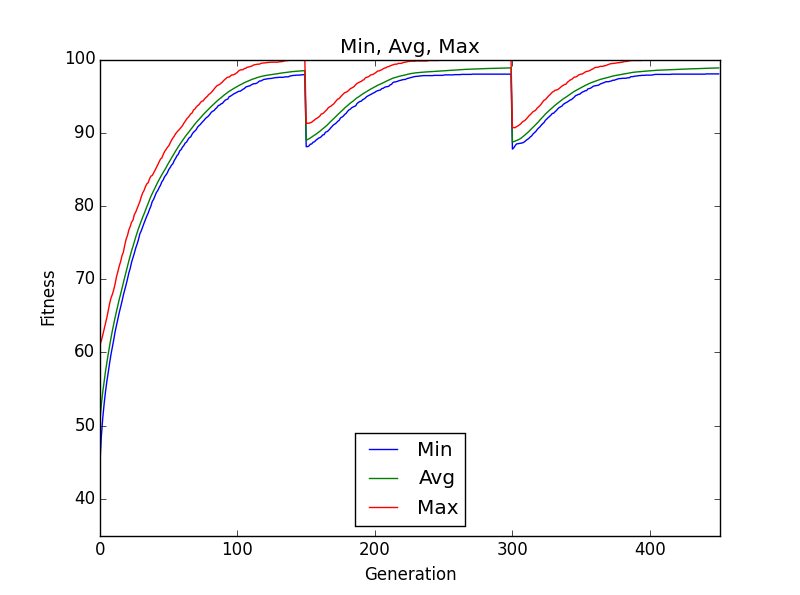}\label{fig:f2}}
	\caption{Performance of Removal of Genotype on Two Moving Peaks}
\end{figure}

\begin{figure}[H]
	\centering
	\subfloat[Diversity]{\includegraphics[width=0.5\textwidth,height=5cm]{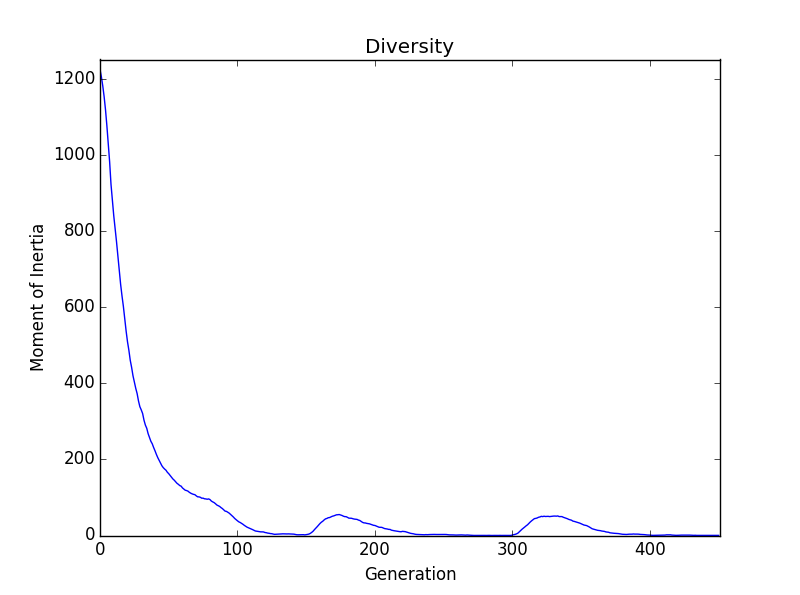}\label{fig:f1}}
	\hfill
	\subfloat[Fitness]{\includegraphics[width=0.5\textwidth,height=5cm]{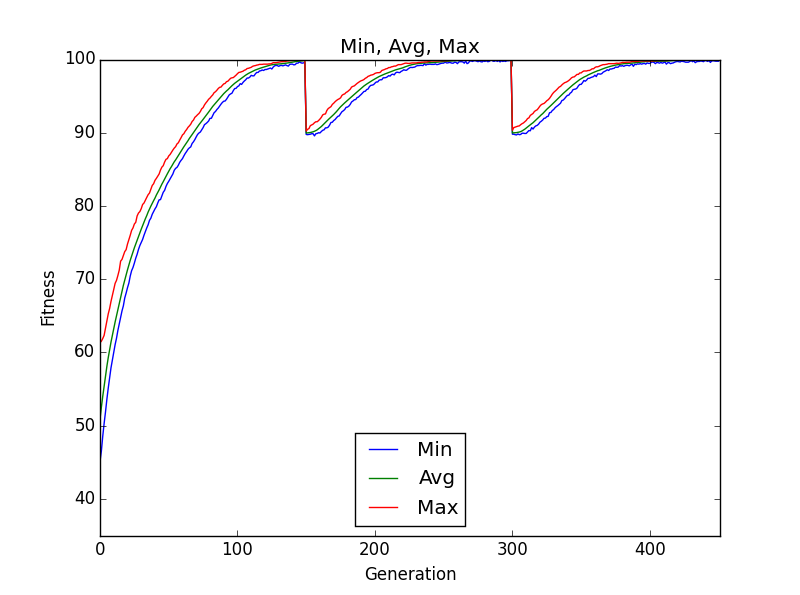}\label{fig:f2}}
	\caption{Performance of Incest Prevention on Two Moving Peaks}
\end{figure}

\begin{figure}[H]
	\centering
	\subfloat[Diversity]{\includegraphics[width=0.5\textwidth,height=5cm]{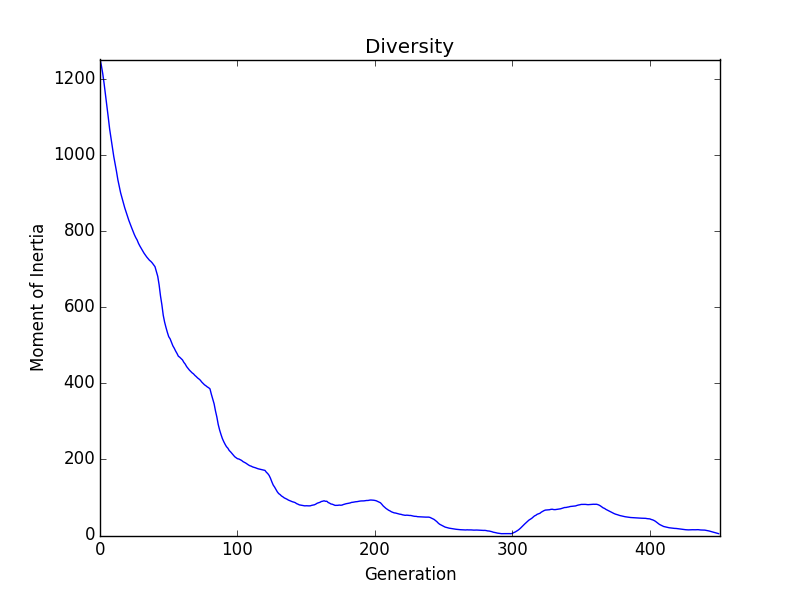}\label{fig:f1}}
	\hfill
	\subfloat[Fitness]{\includegraphics[width=0.5\textwidth,height=5cm]{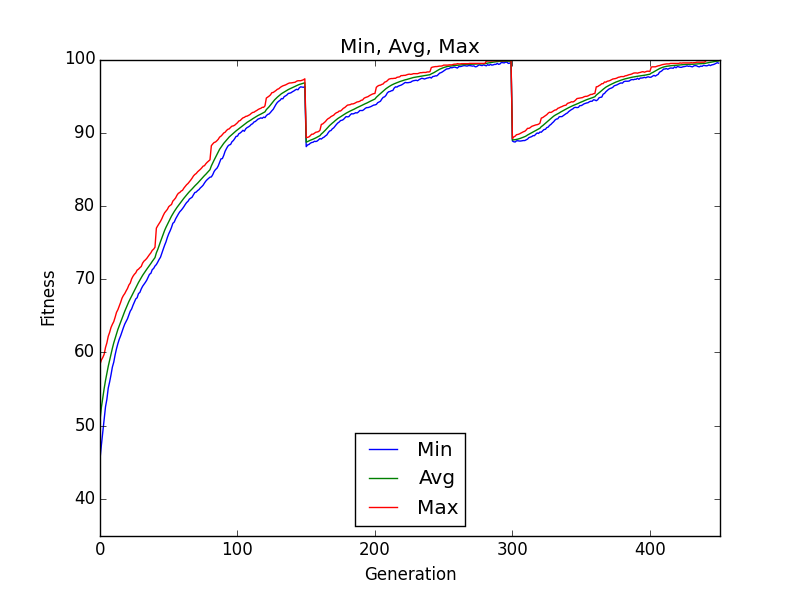}\label{fig:f2}}
	\caption{Performance of Islands Models on Two Moving Peaks}
\end{figure}

\begin{figure}[H]
	\centering
	\subfloat[Diversity]{\includegraphics[width=0.7\textwidth,height=5cm]{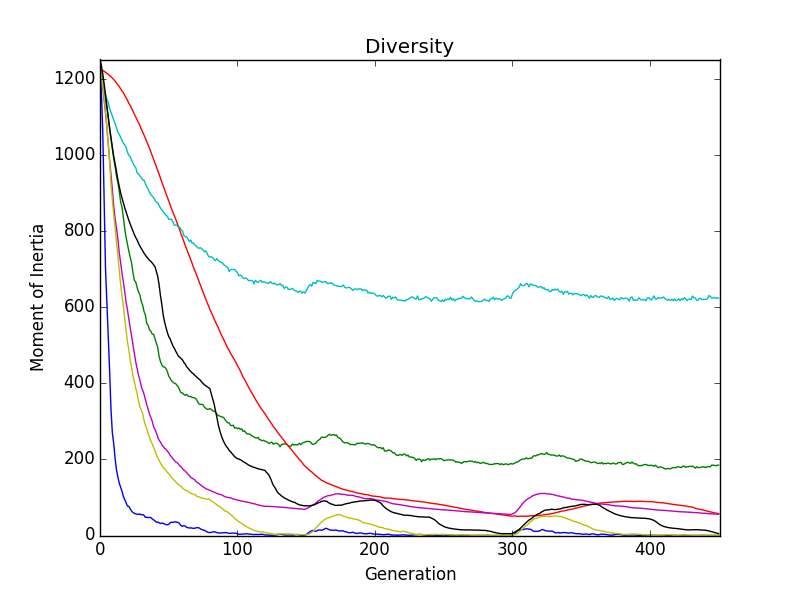}\label{fig:f1}}
	%	\hfill
	\subfloat[Legend]{\includegraphics[width=0.3\textwidth,height=5cm]{figures/legend.png}\label{fig:f2}}
	\\
	\hspace*{-4.8cm}
	\subfloat[Fitness]{\includegraphics[width=0.7\textwidth,height=5cm]{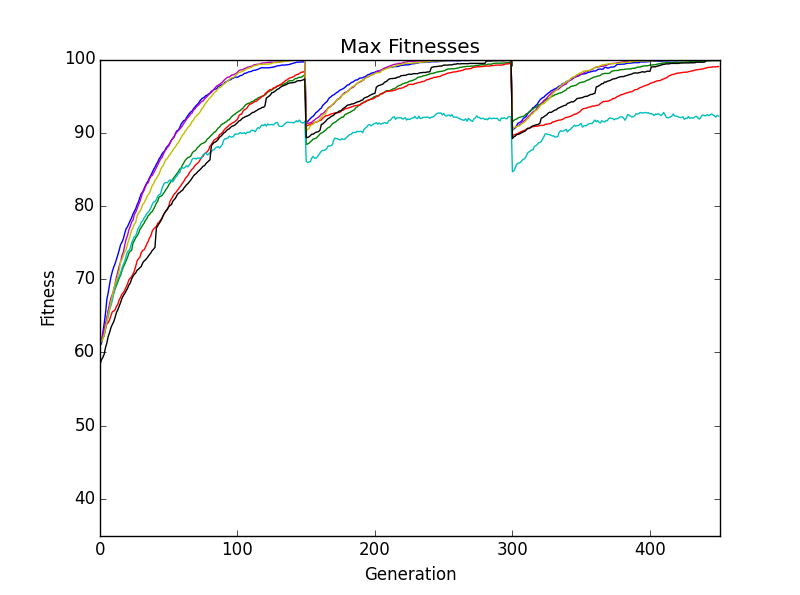}\label{fig:f2}}
	\caption{Performance of All Algorithms on Two Moving Peaks}
\end{figure}

\begin{center}
	\captionof{table}{Offline Performance and Maximum Achieved Fitness on Tne Moving Peaks}
	\begin{tabular}{| p{2cm} | p{1cm} | p{1.5cm} | p{1.5cm} | p{1.5cm} | p{1.5cm} | p{1.7cm} | p{1.5cm} |}
		\hline
		Algorithm & Basic & Clearing & Crowding & Fitness Sharing & Genotype Removal & Incest Prevention & Island Model \\ \hline
		Offline Performance & 95.594 & 93.185 & 91.803 & 90.117 & 95.438 & 95.182 & 93.082 \\ \hline
		Maximum Achieved Fitness & 100\newline 100\newline 100 & 98\newline 100\newline 100 & 98\newline 99\newline 99 & 92\newline 92\newline 93 & 100\newline 100\newline 100 & 100\newline 100\newline 100 & 97\newline 100\newline 100 \\ \hline
		Finds Both Peaks & No & No & No & No & No & No & No \\ \hline
		\hline
	\end{tabular}
\end{center}

All of the algorithms failed to solve this problem, with only fitness sharing having any notable amount of population diversity. While the previous results from $TwoMax$ had indicated that the basic, genotype removal, incest prevention and island model algorithms would be unable to locate both peaks it is unexpected that the other algorithms would also fail to do so.

An idea as to why clearing, crowding and fitness sharing failed to locate the second peak is that early in the runs, while the population was still diverse enough to not be penalised (in clearing and fitness sharing), the population began to move up the taller peak. When the diversity mechanisms started to have a sizeable effect on the populations they were already on a single peak and were unable to jump across to the other one. This can be seen in the fitness sharing graphs (5.28) as the mechanism does produce a diverse population but the diversity comes from having many low fitness individuals on the higher peak. Having a larger independent probability of a bit flipping in the mutation operator might allow individuals stuck on one peak to jump to the other one.

In order to test this hypothesis I increased the probability of mutation in the bit flip function from $1/stringLength$ to $10/stringLength$. The results below show the diversity of 30 runs plotted on a graph but they are not averaged.

\begin{figure}[H]
	\centering
	\subfloat[p=$1/stringLength$]{\includegraphics[width=0.5\textwidth,height=5cm]{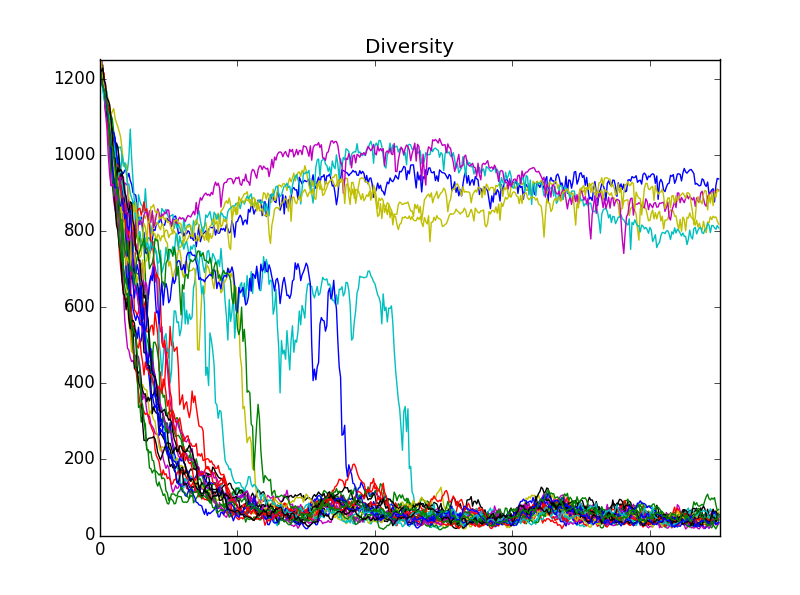}\label{fig:f1}}
	\hfill
	\subfloat[p=$10/stringLength$]{\includegraphics[width=0.5\textwidth,height=5cm]{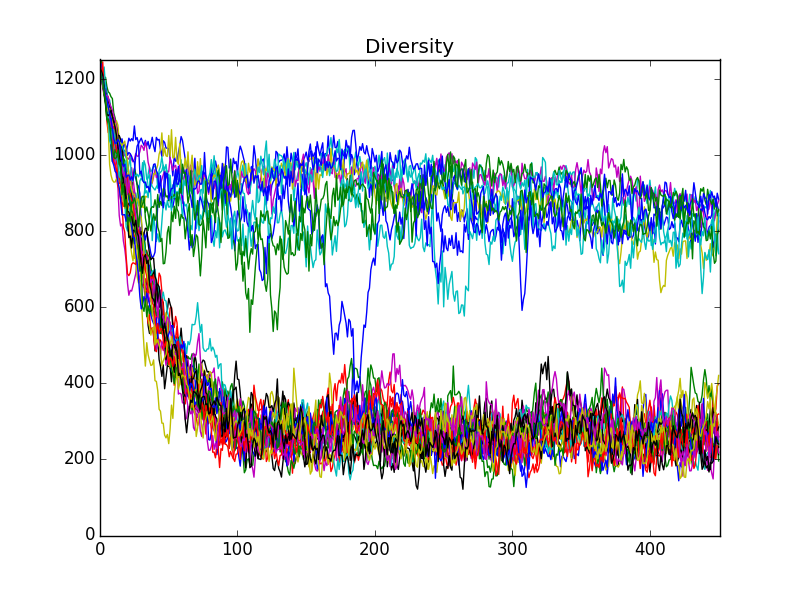}\label{fig:f2}}
	\caption{Diversity on Two Moving Peaks with different bit flip Probabilities}
\end{figure}

The graphs in figure 5.33 show that enough runs to be significant are able to locate both peaks when the bit flip probability is increased. While this is outside of the project scope as all variables that can be are kept constant, this information is interesting and could show the need for further work to be carried out.

\section{Height Changing Peaks}

This problem was initialised with two peaks at $\{0,0,\dots\}$ and $\{1,1,\dots\}$ both with height 100. At generation $150$ peak 0's height is reduced to 80. At generation $300$ peak 0's height is increased back to 100 and peak 1's height is reduced to 80. This means that the maximum fitness is always 100 but the peak where the global optima is located changes. 

\begin{figure}[H]
	\centering
	\subfloat[Diversity]{\includegraphics[width=0.5\textwidth,height=5cm]{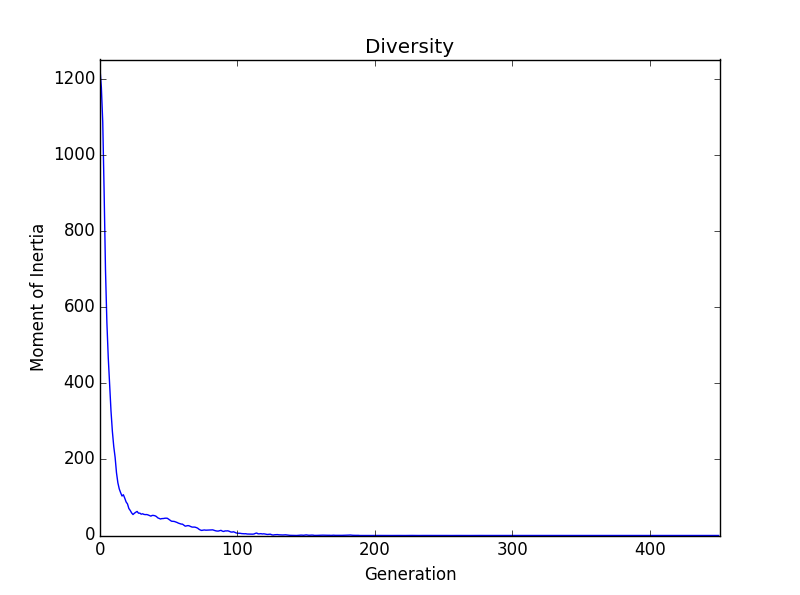}\label{fig:f1}}
	\hfill
	\subfloat[Fitness]{\includegraphics[width=0.5\textwidth,height=5cm]{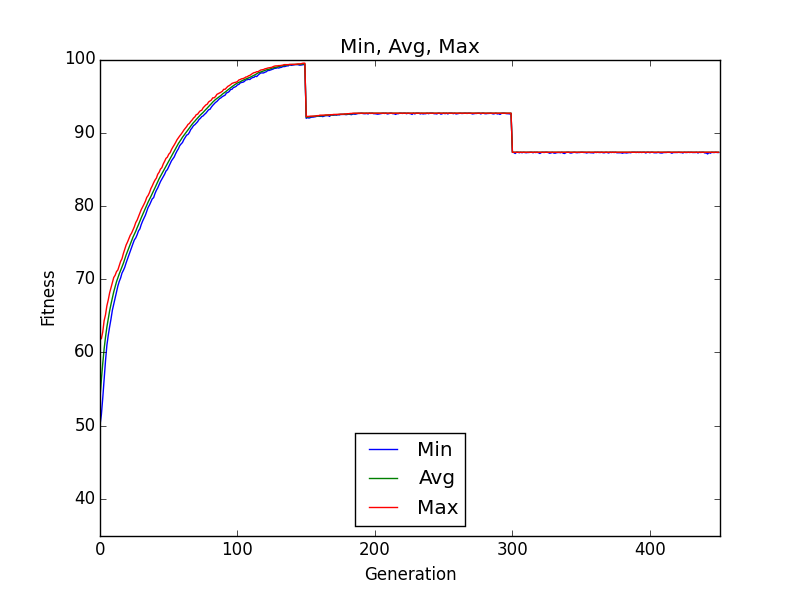}\label{fig:f2}}
	\caption{Performance of Basic on Height Changing Peaks}
\end{figure}

In figure 5.34 you can see how the basic algorithm finds one of the peaks which are both at height 100 and then when the peaks height changes the fitness of the population can be seen to instantly drop. These graphs do not show the expected performance for this algorithm, I expected that that fitness would rise to 100 and then drop to 90 for the rest of the graph. This was the expected result as in the case that peak 0 is initially converged too the algorithm will have fitness 100 then 80 then 100, if peak 1 was converged too the fitness would be 100, 100, 80. When these fitnesses are averaged together they would give 100, 90, 90. However the fitness drops to slightly above 90  the first time the peaks height changes and then slightly below 90 after the second change. This is just down to random chance as the population initially converged to peak 1 on a few more of the 30 runs that the average was generated over. If the average was created using more runs then the predicted result would likely be shown.

\begin{figure}[H]
	\centering
	\subfloat[Diversity]{\includegraphics[width=0.5\textwidth,height=5cm]{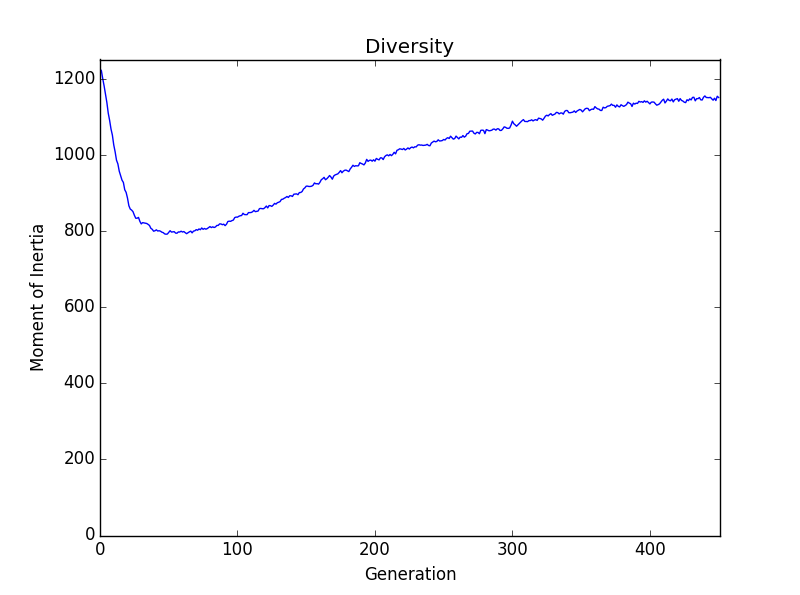}\label{fig:f1}}
	\hfill
	\subfloat[Fitness]{\includegraphics[width=0.5\textwidth,height=5cm]{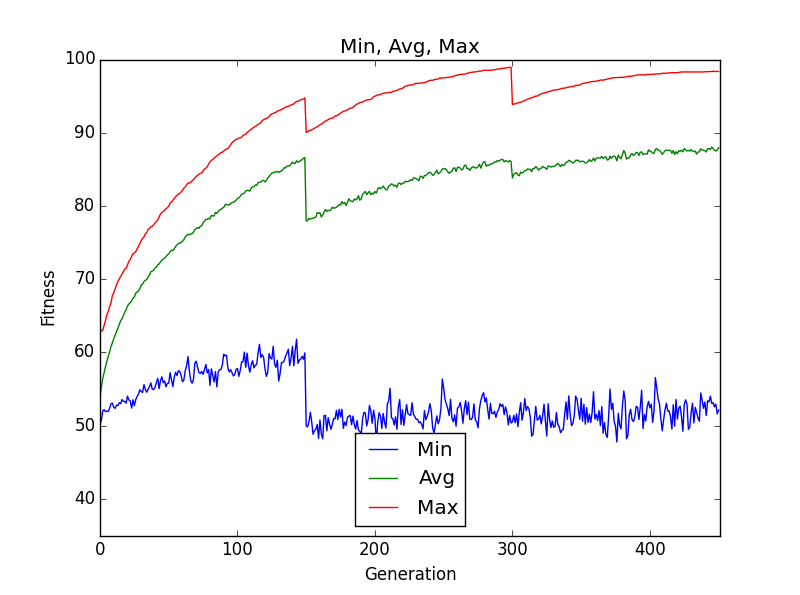}\label{fig:f2}}
	\caption{Performance of Clearing on Height Changing Peaks}
\end{figure}

Figure 5.35 shows that on this problem clearing is able to locate both peaks and as such is able to repeatedly find the optima after the peaks heights change.

\begin{figure}[H]
	\centering
	\subfloat[Diversity]{\includegraphics[width=0.5\textwidth,height=5cm]{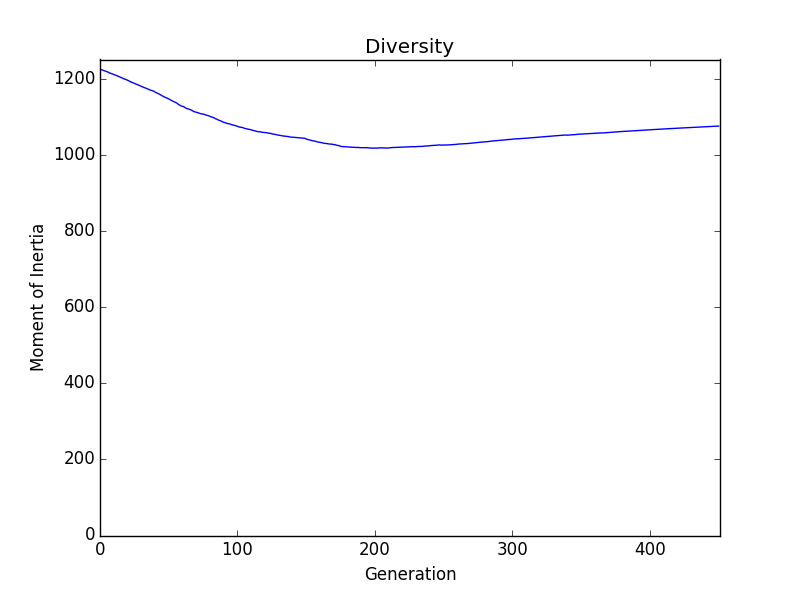}\label{fig:f1}}
	\hfill
	\subfloat[Fitness]{\includegraphics[width=0.5\textwidth,height=5cm]{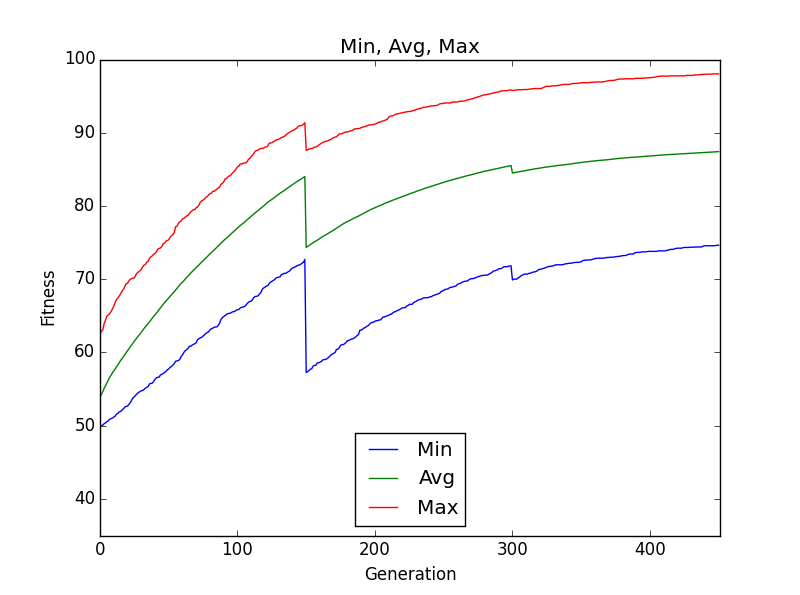}\label{fig:f2}}
	\caption{Performance of Crowding on Height Changing Peaks}
\end{figure}

\begin{figure}[H]
	\centering
	\subfloat[Diversity]{\includegraphics[width=0.5\textwidth,height=5cm]{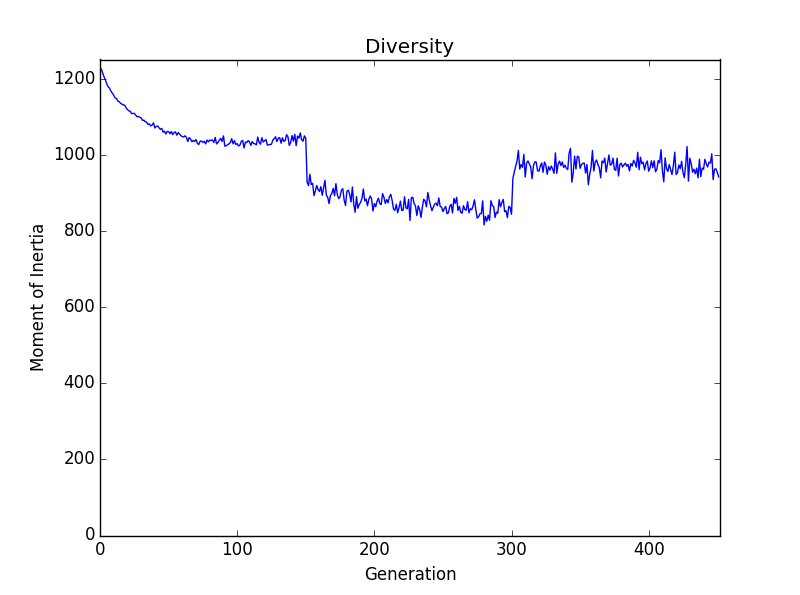}\label{fig:f1}}
	\hfill
	\subfloat[Fitness]{\includegraphics[width=0.5\textwidth,height=5cm]{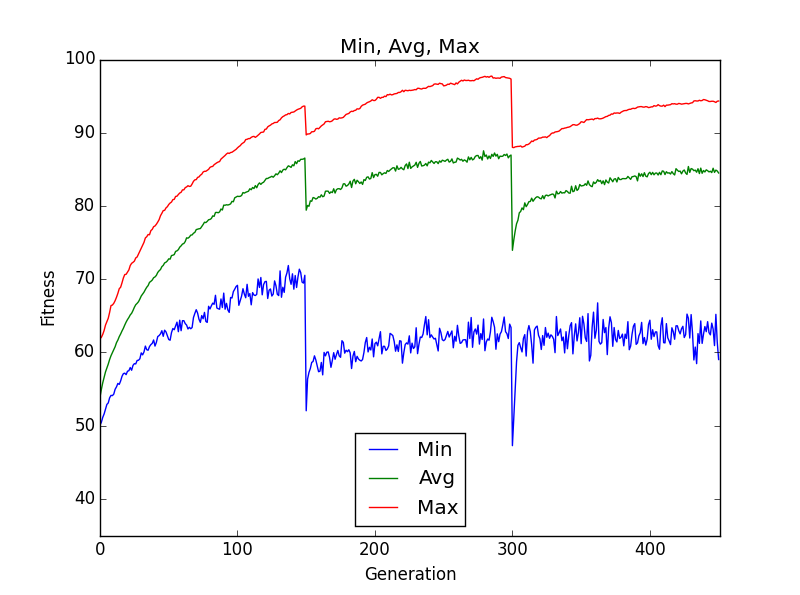}\label{fig:f2}}
	\caption{Performance of Fitness Sharing on Height Changing Peaks}
\end{figure}

Fitness sharing is also able to find both peaks on this problem but is unable to converge to the exact optima. The population diversity graph (figure 5.37 (b)) for this population is interesting as there are notable shifts in the diversity when the peaks change. This seems to be caused by the subpopulations at each peak being unequal in size. When one peak becomes the optima then more of the population move to that peak and the diversity drops, when the peaks swap later in the run some of the population move back to the other peak and the subpopulation sizes become more even again. This being the reason would explain why the same effect is not seen in the clearing algorithm as the subpopulation sizes are capped in that algorithm.

\begin{figure}[H]
	\centering
	\subfloat[Diversity]{\includegraphics[width=0.5\textwidth,height=5cm]{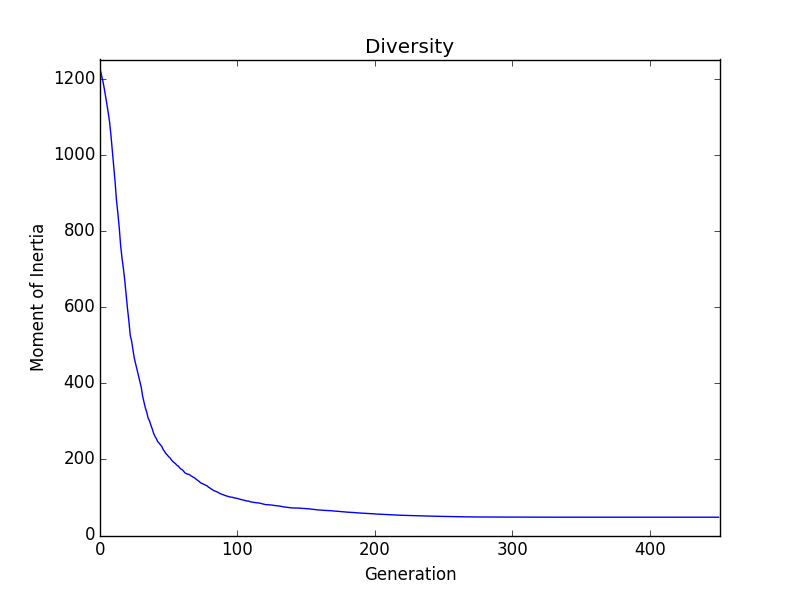}\label{fig:f1}}
	\hfill
	\subfloat[Fitness]{\includegraphics[width=0.5\textwidth,height=5cm]{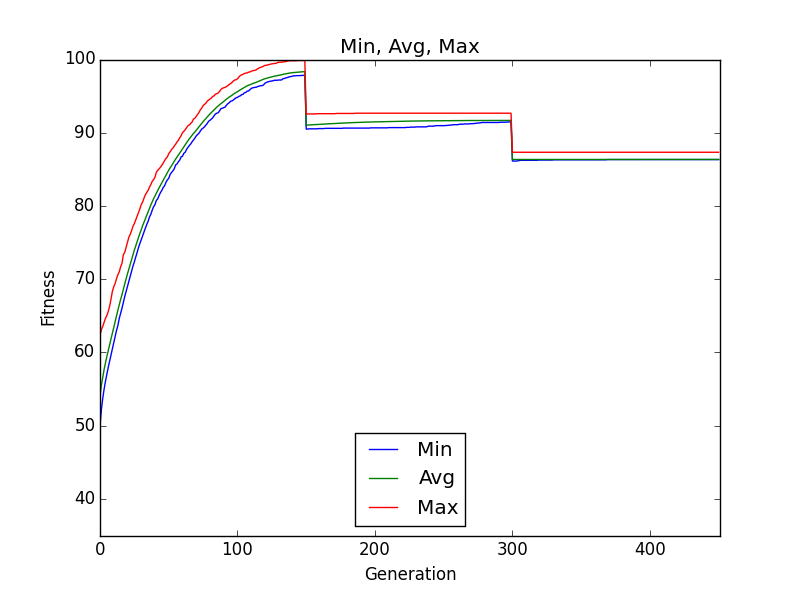}\label{fig:f2}}
	\caption{Performance of Removal of Genotype on Height Changing Peaks}
\end{figure}

\begin{figure}[H]
	\centering
	\subfloat[Diversity]{\includegraphics[width=0.5\textwidth,height=5cm]{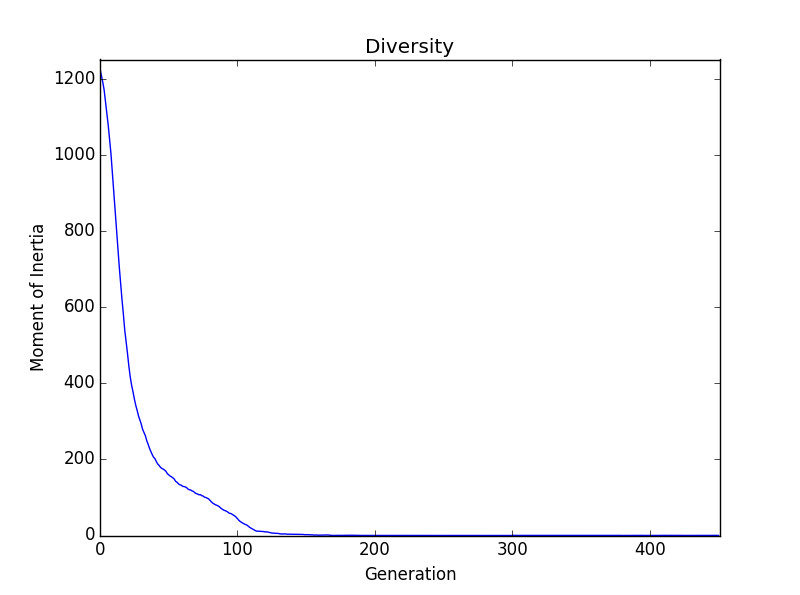}\label{fig:f1}}
	\hfill
	\subfloat[Fitness]{\includegraphics[width=0.5\textwidth,height=5cm]{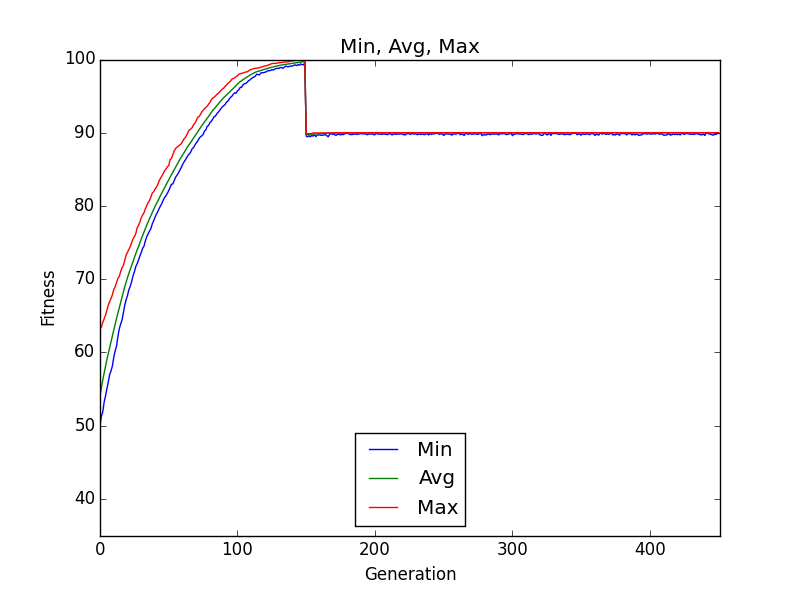}\label{fig:f2}}
	\caption{Performance of Incest Prevention on Height Changing Peaks}
\end{figure}

Figure 5.39 shows the expected performance that was mentioned when discussing the basic algorithms performance on this problem. This would suggest that incest prevention converged to each peak an equal number of times over all the runs.

\begin{figure}[H]
	\centering
	\subfloat[Diversity]{\includegraphics[width=0.5\textwidth,height=5cm]{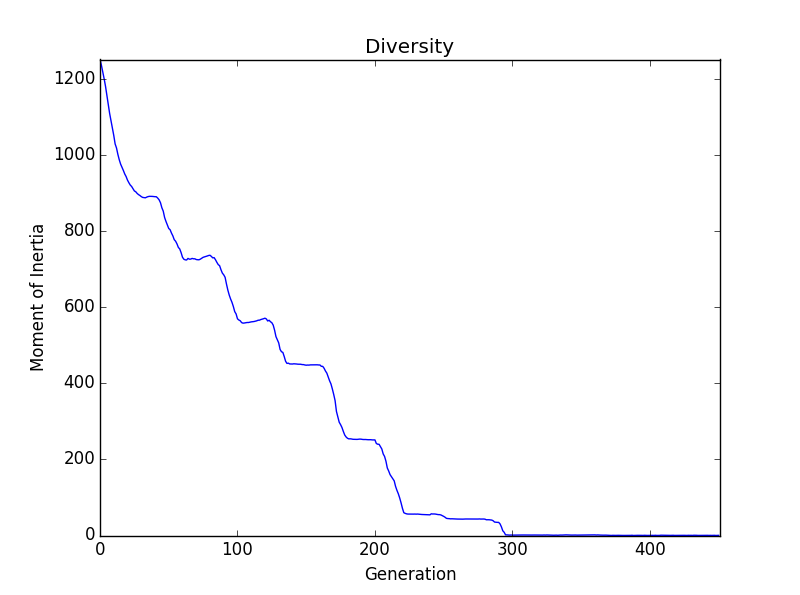}\label{fig:f1}}
	\hfill
	\subfloat[Fitness]{\includegraphics[width=0.5\textwidth,height=5cm]{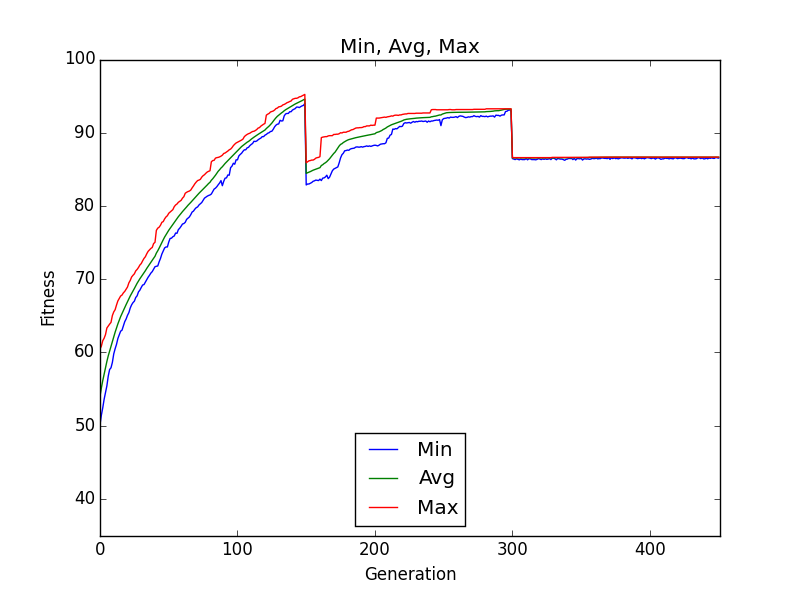}\label{fig:f2}}
	\caption{Performance of Islands Models on Height Changing Peaks}
\end{figure}

\begin{figure}[H]
	\centering
	\subfloat[Diversity]{\includegraphics[width=0.7\textwidth,height=5cm]{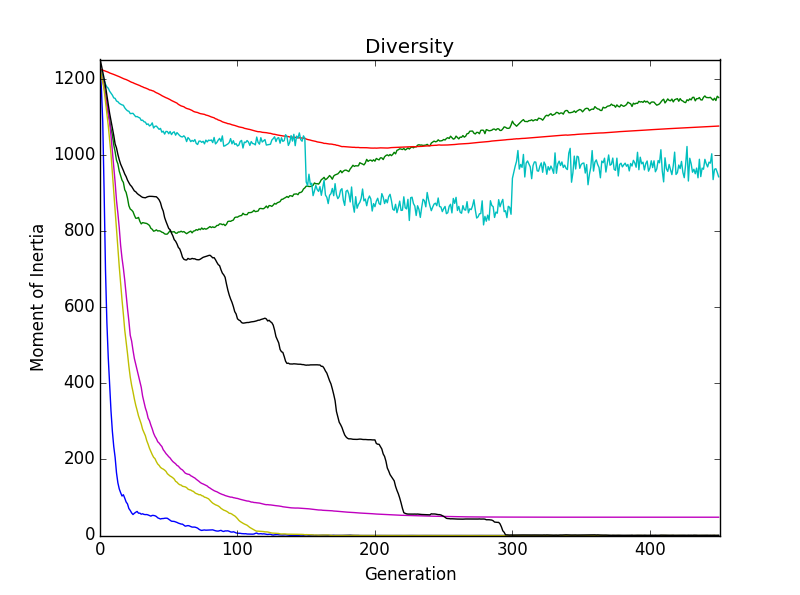}\label{fig:f1}}
	%	\hfill
	\subfloat[Legend]{\includegraphics[width=0.3\textwidth,height=5cm]{figures/legend.png}\label{fig:f2}}
	\\
	\hspace*{-4.8cm}
	\subfloat[Fitness]{\includegraphics[width=0.7\textwidth,height=5cm]{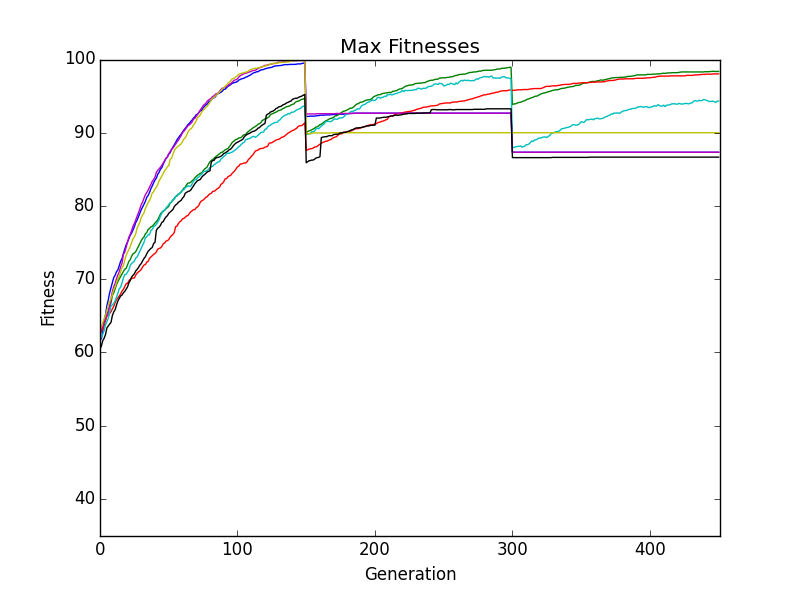}\label{fig:f2}}
	\caption{Performance of All Algorithms on Height Changing Peaks}
\end{figure}
\newpage
\begin{center}
	\captionof{table}{Offline Performance and Maximum Achieved Fitness on Height Chaging Peaks}
	\begin{tabular}{| p{2cm} | p{1cm} | p{1.5cm} | p{1.5cm} | p{1.5cm} | p{1.5cm} | p{1.7cm} | p{1.5cm} |}
		\hline
		Algorithm & Basic & Clearing & Crowding & Fitness Sharing & Genotype Removal & Incest Prevention & Island Model \\ \hline
		Offline Performance & 89.825 & 92.099 & 89.763 & 90.461 & 89.866 & 89.722 & 87.632 \\ \hline
		Maximum Achieved Fitness & 100\newline 93\newline 87 & 95\newline 99\newline 99 & 91\newline 96\newline 98 & 94\newline 98\newline 94 & 100\newline 93\newline 87 & 100\newline 90\newline 90 & 95\newline 93\newline 87 \\ \hline
		Finds Both Peaks & No & Yes & Yes & Yes & No & No & No \\ \hline
		\hline
	\end{tabular}
\end{center}

The results of this problem show a clear benefit to having a diverse population. Clearing, crowding and fitness sharing have subpopulations at both peaks and are thus able to recover when the global maxima changes peak. Clearing performed the best on this problem as out of the algorithms able to find both peaks it was repeatedly the closest to the optima.

The other algorithms initially move to one of the peaks at random, due to them having the same height at the start of the problem. The population then stays at that peak even when it is no longer as good of a solution. This behaviour is not unexpected as previous problems have shown that these algorithms struggle to maintain multiple subpopulations.

\section{Moving Height Changing Peaks}

This problem is a combination of the previous two problems, it was initialised with two peaks at $\{0,0,\dots\}$ and $\{1,1,\dots\}$ both at height 100. At generations $150$ and $300$ both peaks move using a bit flip function with probability of flipping $p=0.1$. The heights alter such that at generation $150$ peak 0 has height 80 and peak 1 has height 100 and at generation $300$ peak 0 has height 100 and peak 1 has height 80. This means that the global maxima is always at 100. 

\begin{figure}[H]
	\centering
	\subfloat[Diversity]{\includegraphics[width=0.5\textwidth,height=5cm]{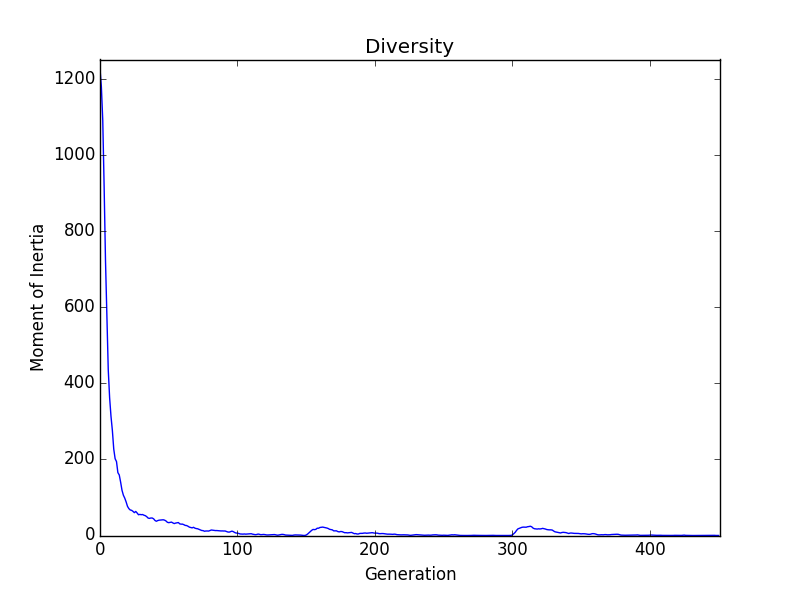}\label{fig:f1}}
	\hfill
	\subfloat[Fitness]{\includegraphics[width=0.5\textwidth,height=5cm]{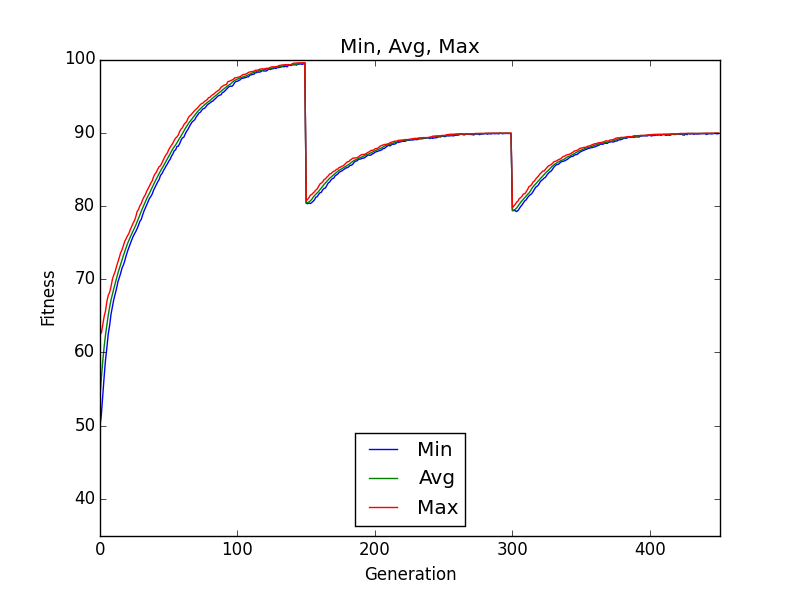}\label{fig:f2}}
	\caption{Performance of Basic on Moving Height Changing Peaks}
\end{figure}

As the problem was a combination of the previous two problems the results (figure 5.42) were expected to be also. The drop in population fitness is more severe than on the previous problems as the peaks movement and height change happen simultaneously. After the drop basic is able to reconverge to the peak it was already near but unable to find the optima after the first movement of the peaks. The difference between this and height changing peaks is that the population initially converged to each peak at a more equal rate, as the population reaches the same average fitness after both peak moves.

\begin{figure}[H]
	\centering
	\subfloat[Diversity]{\includegraphics[width=0.5\textwidth,height=5cm]{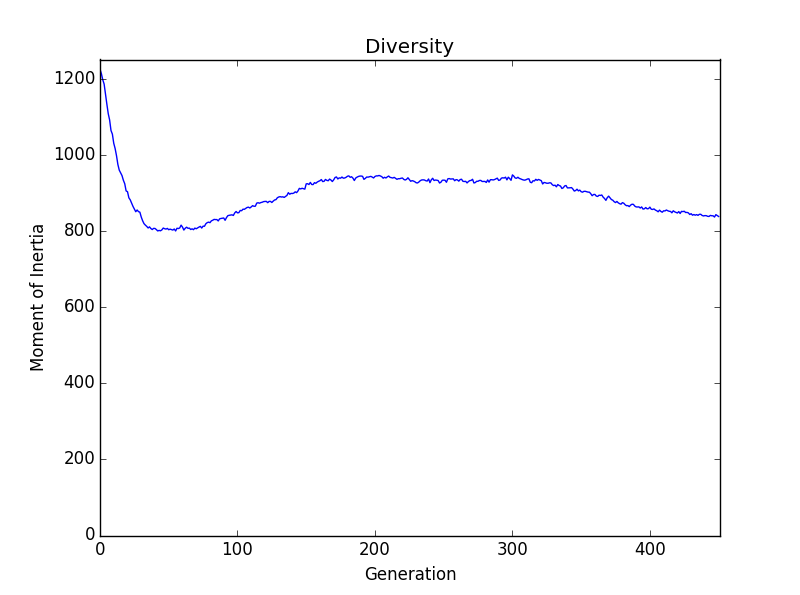}\label{fig:f1}}
	\hfill
	\subfloat[Fitness]{\includegraphics[width=0.5\textwidth,height=5cm]{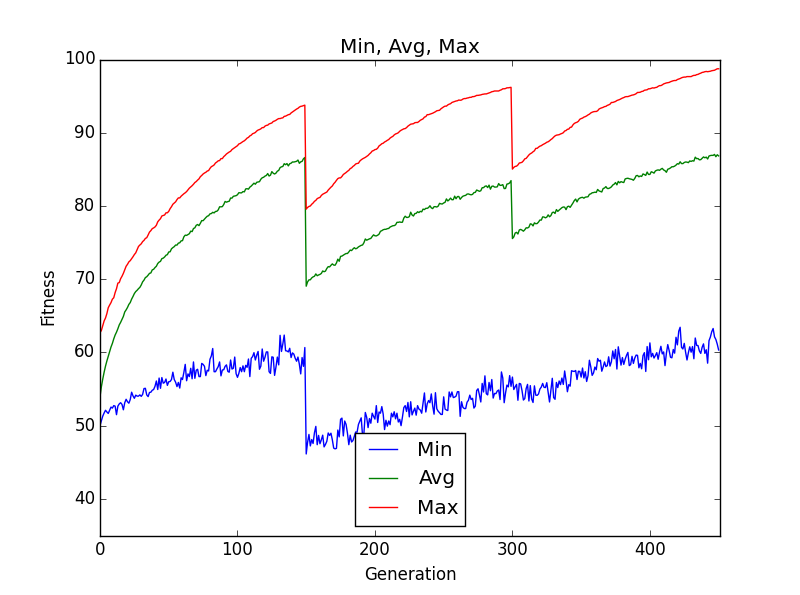}\label{fig:f2}}
	\caption{Performance of Clearing on Moving Height Changing Peaks}
\end{figure}

\begin{figure}[H]
	\centering
	\subfloat[Diversity]{\includegraphics[width=0.5\textwidth,height=5cm]{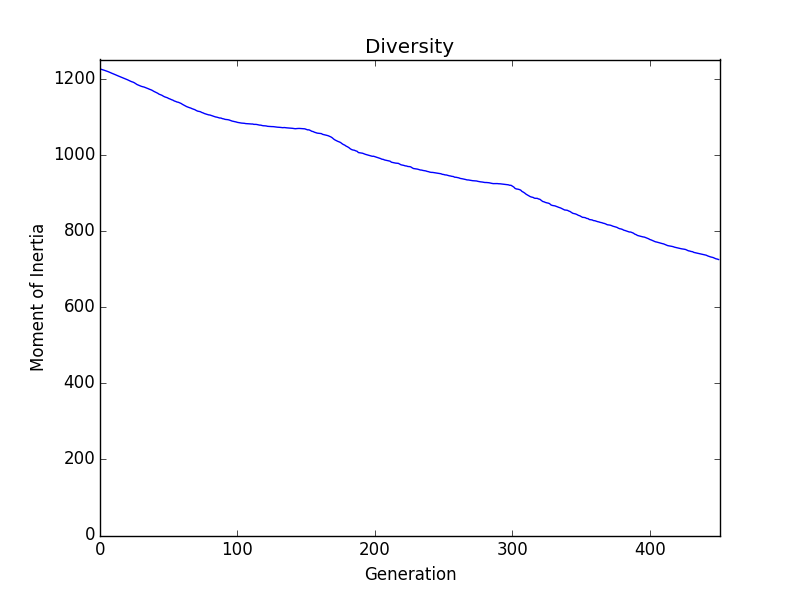}\label{fig:f1}}
	\hfill
	\subfloat[Fitness]{\includegraphics[width=0.5\textwidth,height=5cm]{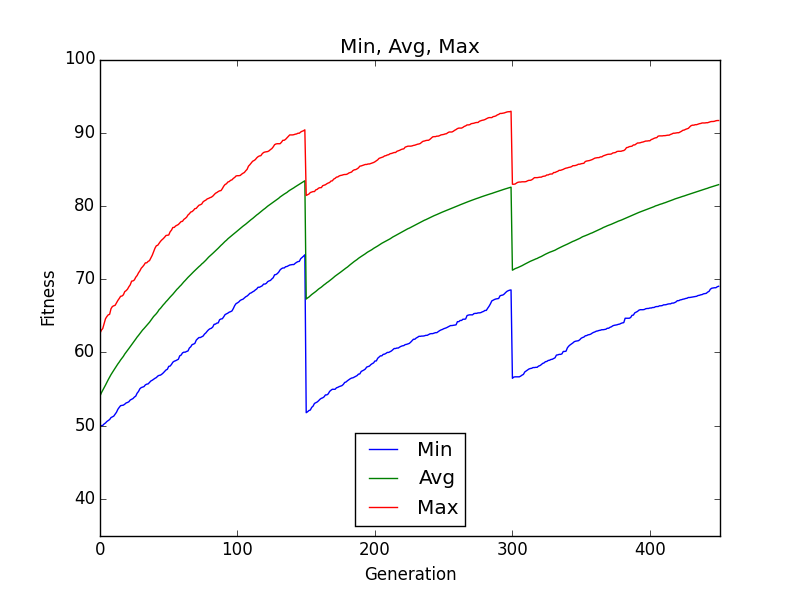}\label{fig:f2}}
	\caption{Performance of Crowding on Moving Height Changing Peaks}
\end{figure}

\begin{figure}[H]
	\centering
	\subfloat[Diversity]{\includegraphics[width=0.5\textwidth,height=5cm]{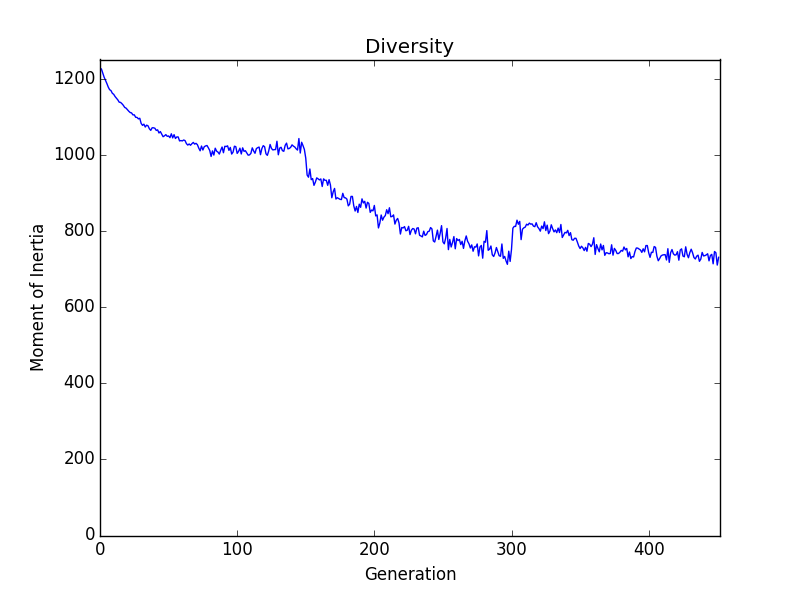}\label{fig:f1}}
	\hfill
	\subfloat[Fitness]{\includegraphics[width=0.5\textwidth,height=5cm]{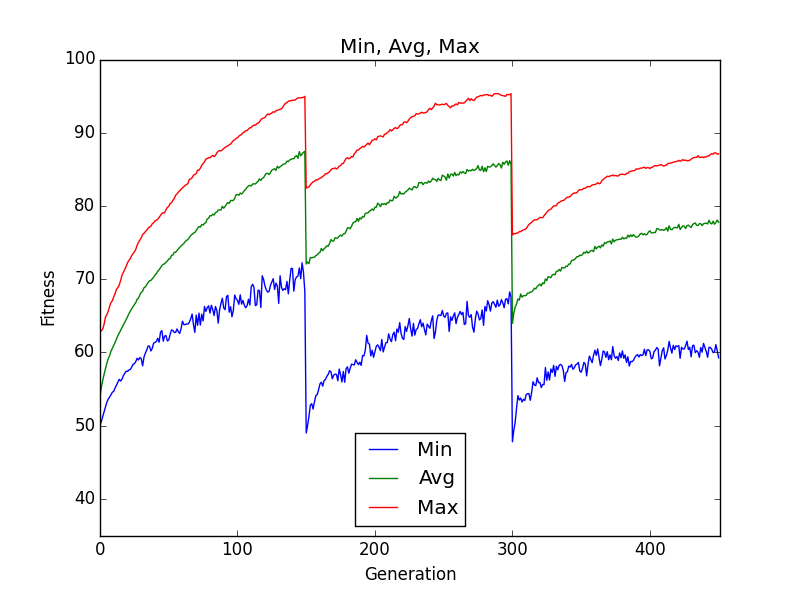}\label{fig:f2}}
	\caption{Performance of Fitness Sharing on Moving Height Changing Peaks}
\end{figure}

\begin{figure}[H]
	\centering
	\subfloat[Diversity]{\includegraphics[width=0.5\textwidth,height=5cm]{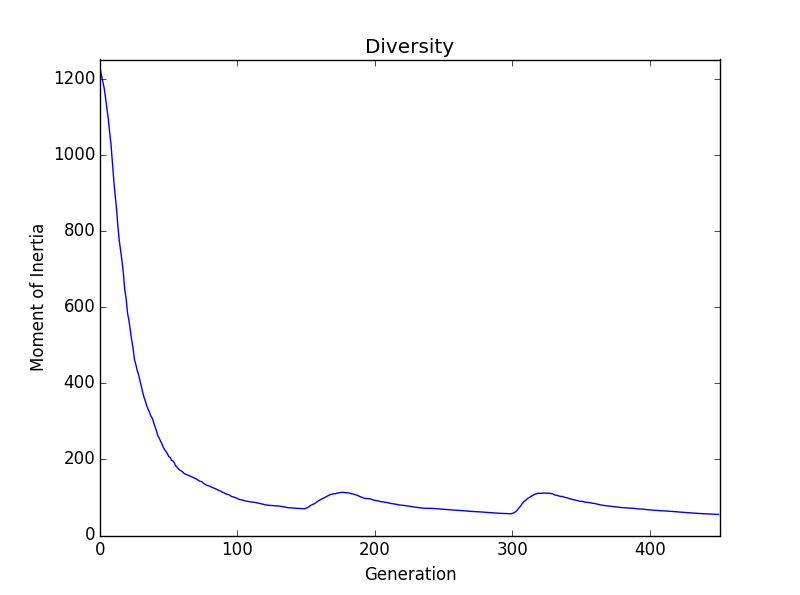}\label{fig:f1}}
	\hfill
	\subfloat[Fitness]{\includegraphics[width=0.5\textwidth,height=5cm]{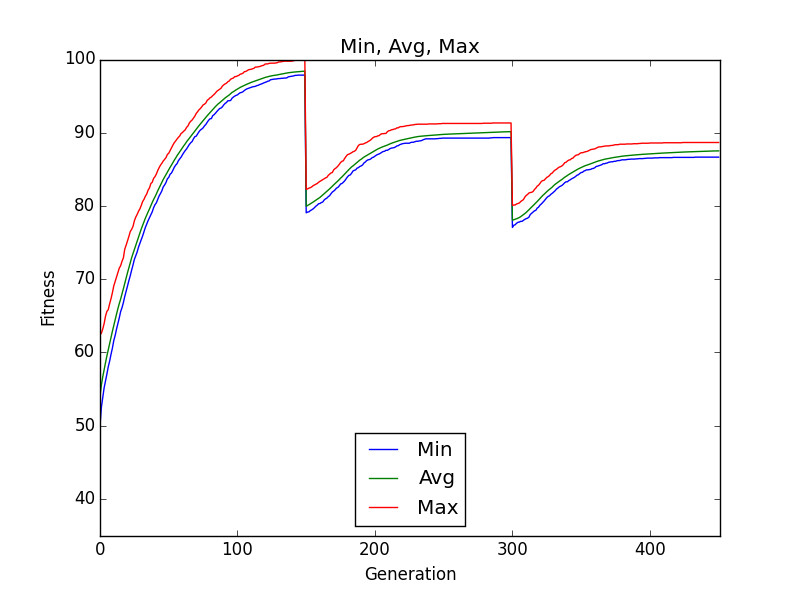}\label{fig:f2}}
	\caption{Performance of Removal of Genotype on Moving Height Changing Peaks}
\end{figure}

\begin{figure}[H]
	\centering
	\subfloat[Diversity]{\includegraphics[width=0.5\textwidth,height=5cm]{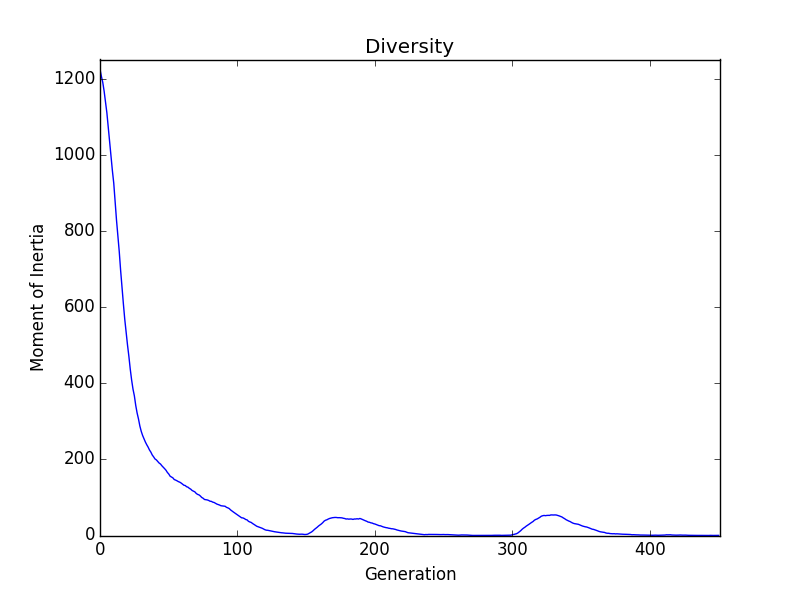}\label{fig:f1}}
	\hfill
	\subfloat[Fitness]{\includegraphics[width=0.5\textwidth,height=5cm]{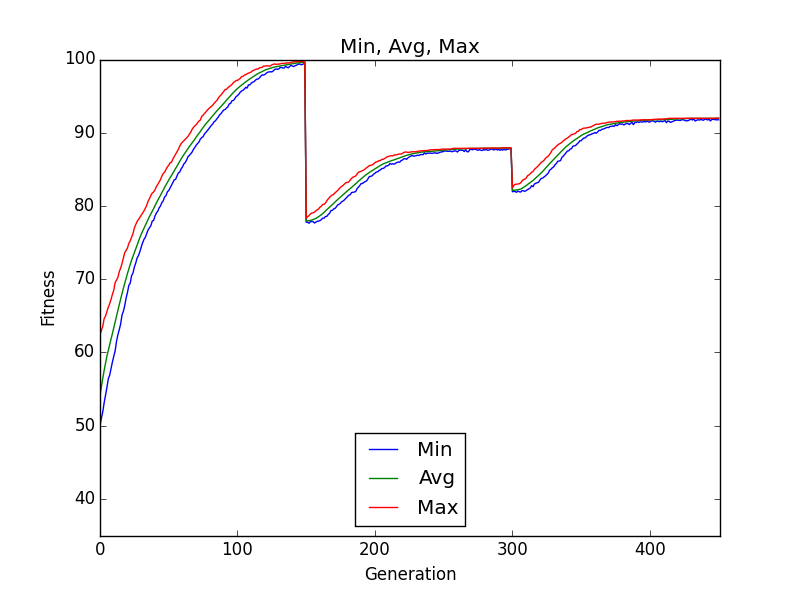}\label{fig:f2}}
	\caption{Performance of Incest Prevention on Moving Height Changing Peaks}
\end{figure}

\begin{figure}[H]
	\centering
	\subfloat[Diversity]{\includegraphics[width=0.5\textwidth,height=5cm]{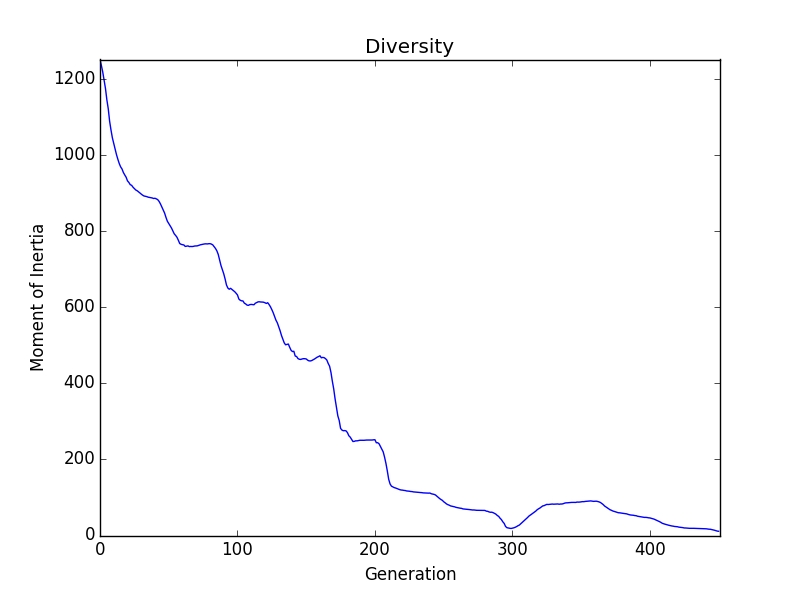}\label{fig:f1}}
	\hfill
	\subfloat[Fitness]{\includegraphics[width=0.5\textwidth,height=5cm]{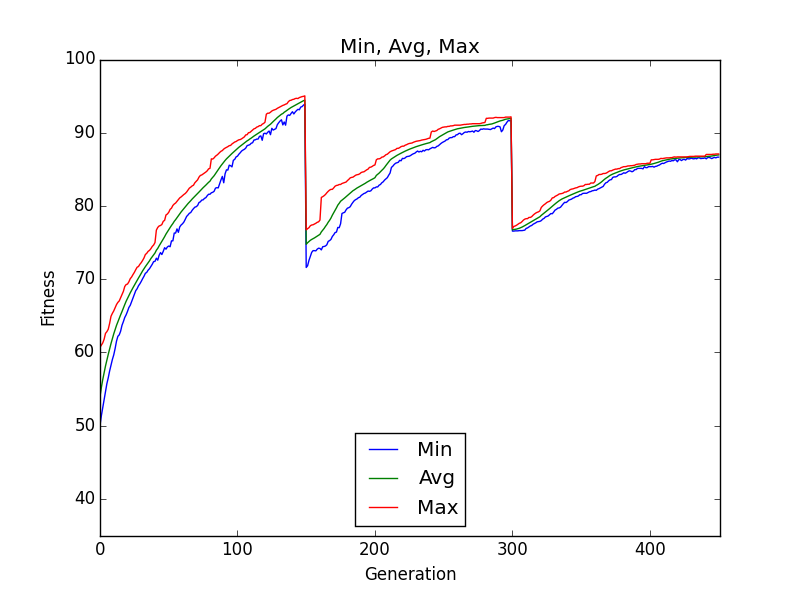}\label{fig:f2}}
	\caption{Performance of Islands Models on Moving Height Changing Peaks}
\end{figure}

\begin{figure}[H]
	\centering
	\subfloat[Diversity]{\includegraphics[width=0.7\textwidth,height=5cm]{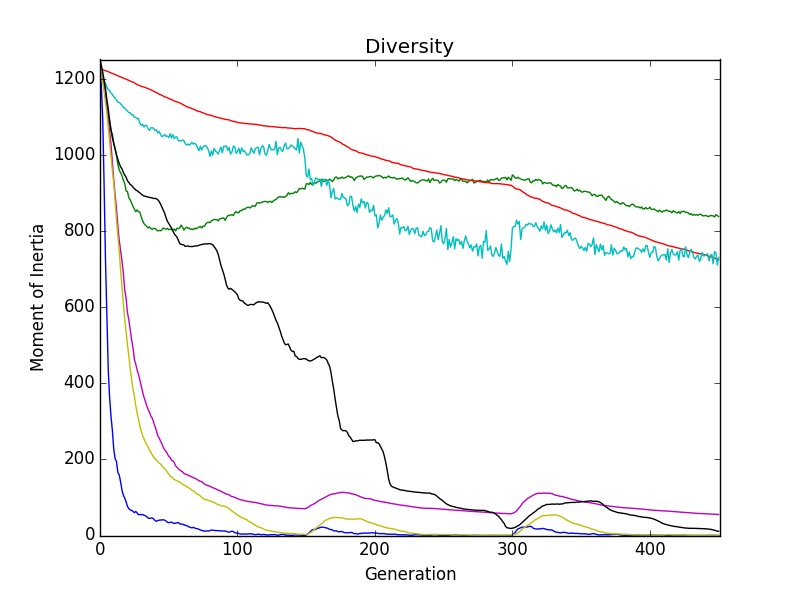}\label{fig:f1}}
	%	\hfill
	\subfloat[Legend]{\includegraphics[width=0.3\textwidth,height=5cm]{figures/legend.png}\label{fig:f2}}
	\\
	\hspace*{-4.8cm}
	\subfloat[Fitness]{\includegraphics[width=0.7\textwidth,height=5cm]{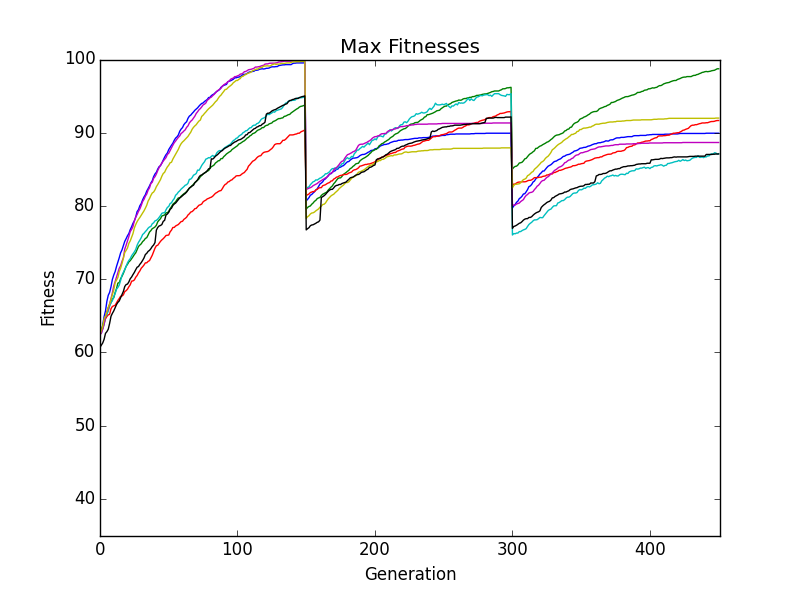}\label{fig:f2}}
	\caption{Performance of All Algorithms on Moving Height Changing Peaks}
\end{figure}

\begin{center}
	\captionof{table}{Offline Performance and Maximum Achieved Fitness on Moving Height Changing Peaks}
	\begin{tabular}{| p{2cm} | p{1cm} | p{1.5cm} | p{1.5cm} | p{1.5cm} | p{1.5cm} | p{1.7cm} | p{1.5cm} |}
		\hline
		Algorithm & Basic & Clearing & Crowding & Fitness Sharing & Genotype Removal & Incest Prevention & Island Model \\ \hline
		Offline Performance & 88.644 & 88.706 & 84.827 & 86.46 & 88.622 & 88.263 & 85.506 \\ \hline
		Maximum Achieved Fitness & 100\newline 90\newline 90 & 94\newline 96\newline 99 & 90\newline 93\newline 92 & 95\newline 95\newline 87 & 100\newline 91\newline 89 & 100\newline 88\newline 92 & 95\newline 92\newline 87 \\ \hline
		Finds Both Peaks & No & Yes & Yes & Yes & No & No & No \\ \hline
		\hline
	\end{tabular}
\end{center}

All of the algorithms struggle to perform well on this algorithm. Clearing, crowding and fitness sharing maintain subpopulations near both peaks but all of them are unable to ever find the global optima, although clearing does come close after the last time the peaks move. The other algorithms show ability to reconverge back to a peak when it moves but they are all trapped on a single peak.

Again these results are not surprising, the problems was a combination of the two moving peaks and height changing peaks problems and the results reflect this. As only clearing crowding and fitness sharing seem able to find multiple peaks they were the only algorithms that could have solved this problem, however these algorithms converge slower which meant that they struggled to keep up with the peaks as they moved.

\section{Interpreting Results}

From looking at the results it appears that the algorithms can be split into two categories. Category one contains Clearing, Crowding and Fitness Sharing. Category two contains Genotype Removal, Incest Prevention and Island models. 

Category one algorithms are better at promoting diversity within a population and as such perform better on problems where having subpopulations on each peak is beneficial. 

Category two algorithms fail to maintain as diverse a population and thus perform poorly on problems where the peak on which the global maxima is located changes. However, these algorithms are shown to converge faster than the algorithms in category one and as such are good at tracking a moving peak. These two categories coincide with whether an algorithm is niching or non niching as was discussed in the literature review, with category one being niching algorithms and category two being non niching. This suggests that niching algorithms are better at promoting diversity than other algorithms.

All of the algorithms performed poorly on both the two moving peaks problem and the moving height changing peaks problem. The failure on the two moving peaks problem came from none of the algorithms being able to locate the second peak as it was at a lower height than the first peak. When peaks are initialised to the same height and then change later category one algorithms are capable of maintaining subpopulations of the lower peaks. This would suggest that the diversity would need to be enforced earlier to keep individuals on both peaks or the individuals would need a way to jump from one peak to another as was briefly discussed.

I suspect that the reason all the algorithms failed on the moving height changing peaks problem is that to successfully solve it an algorithm would need characteristics from both categories of algorithms. To successfully solve this problem an algorithm would need to maintain subpopulations near both peaks but would also need to have fast convergence to track the peaks as they move. I suspect that a new algorithm with traits from both categories would outperform all of the current algorithms on the moving height changing peaks problem problem.

\section{A new Algorithm}

In order to test if a algorithm with traits from both categories would perform better, I have created one. The new algorithm takes traits from the current best performing algorithm on the moving height changing peaks problem (clearing) and traits from the algorithm that performs best on the single moving peak problem (genotype removal).

The new algorithm will have the clearing algorithm method of splitting the population into niches and only letting the best individuals in each niche keep their fitness. This will be combined with the selection method from the genotype removal algorithm which will mean that the population will be made up of unique individuals. The new algorithm should therefore be able to maintain subpopulations at each peak and each subpopulation should consist of unique individuals. It is hoped that the new algorithm will inherit the best performance traits of the algorithms that inspired it, having the diversity of the clearing algorithm with the convergence speed from genotype removal.

In order to test the new algorithm I will run it on the four dynamic test problems faced by the other algorithms. The results are as follows (all results are averaged over 30 runs and have the same random seed as the previous algorithms)

\begin{figure}[H]
	\centering
	\subfloat[Diversity]{\includegraphics[width=0.5\textwidth,height=5cm]{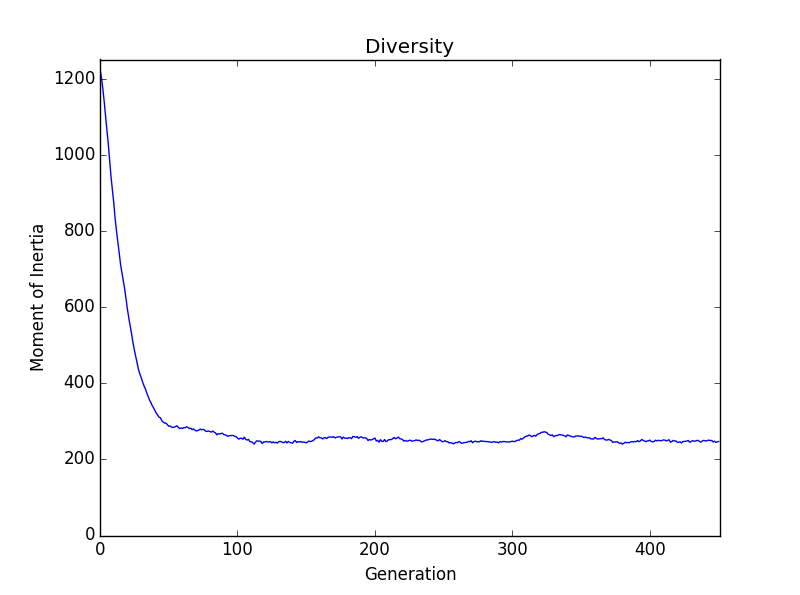}\label{fig:f1}}
	\hfill
	\subfloat[Fitness]{\includegraphics[width=0.5\textwidth,height=5cm]{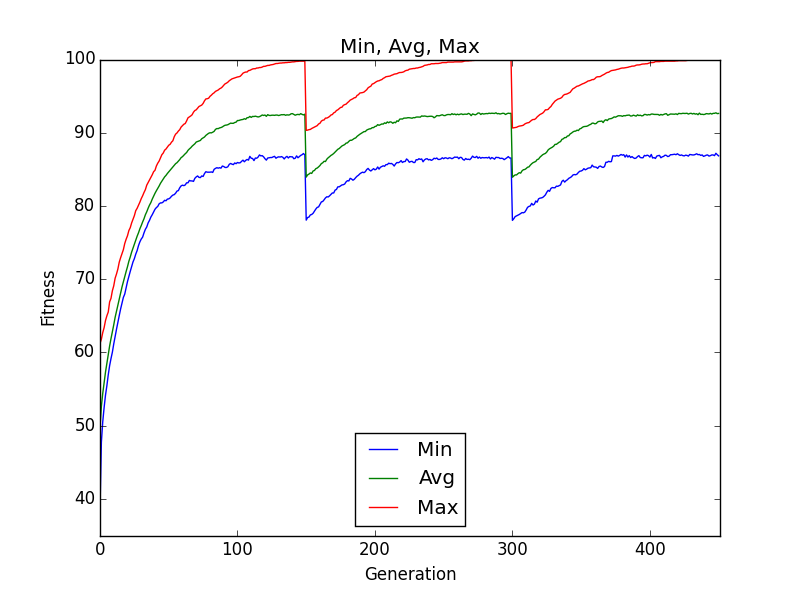}\label{fig:f2}}
	\caption{Performance of New algorithm on One Moving Peak}
\end{figure}

On the single moving peak problem the new algorithm shows better performance than clearing but worse performance than genotype removal. I suspect that this is due to the main factor in success on this problem being convergence time which the clearing component of the algorithm may have slowed slightly. This theory would fit with the algorithms performance being between the two algorithms it is based on. 

\begin{figure}[H]
	\centering
	\subfloat[Diversity]{\includegraphics[width=0.5\textwidth,height=5cm]{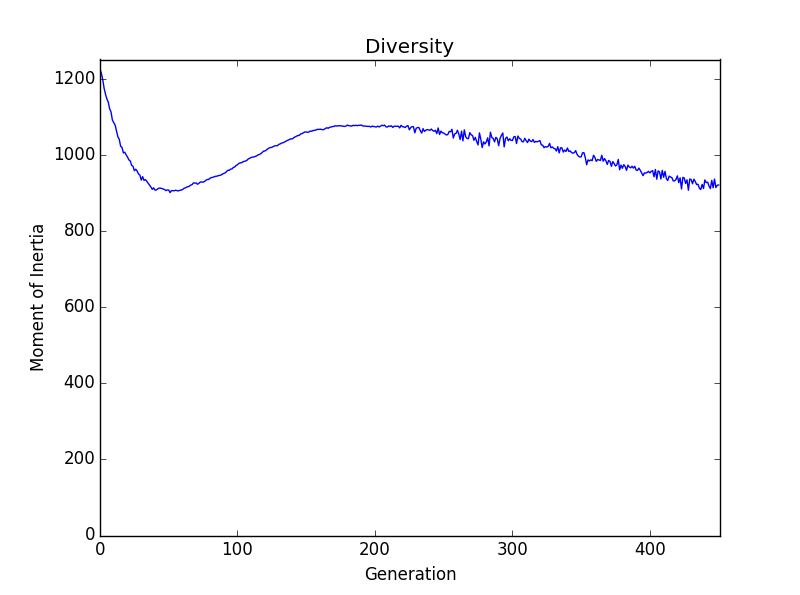}\label{fig:f1}}
	\hfill
	\subfloat[Fitness]{\includegraphics[width=0.5\textwidth,height=5cm]{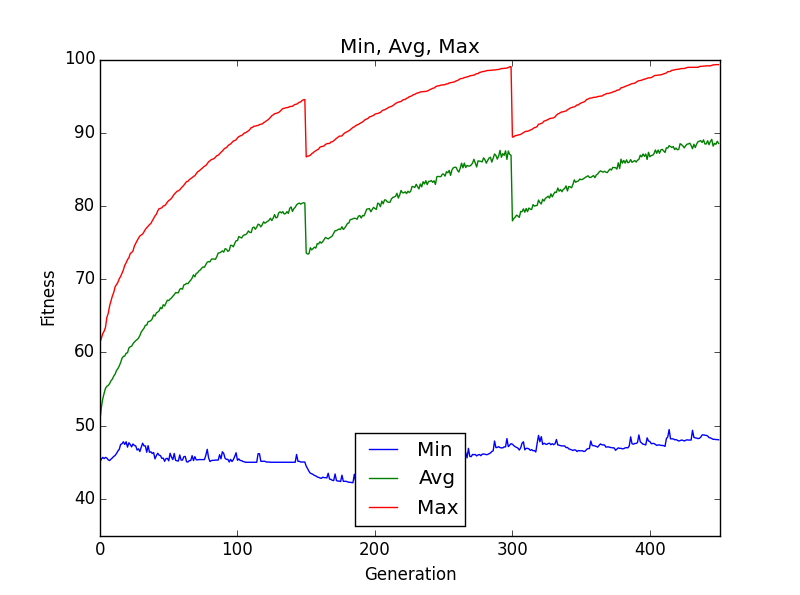}\label{fig:f2}}
	\caption{Performance of New algorithm on Two Moving Peaks}
\end{figure}

The new algorithm performs substantially better than all of the previous algorithms on the two moving peaks problem. It is the only algorithm capable of locating both of the peaks and this is a notable improvement over the other algorithms, where none of them were able to do this.

\begin{figure}[H]
	\centering
	\subfloat[Diversity]{\includegraphics[width=0.5\textwidth,height=5cm]{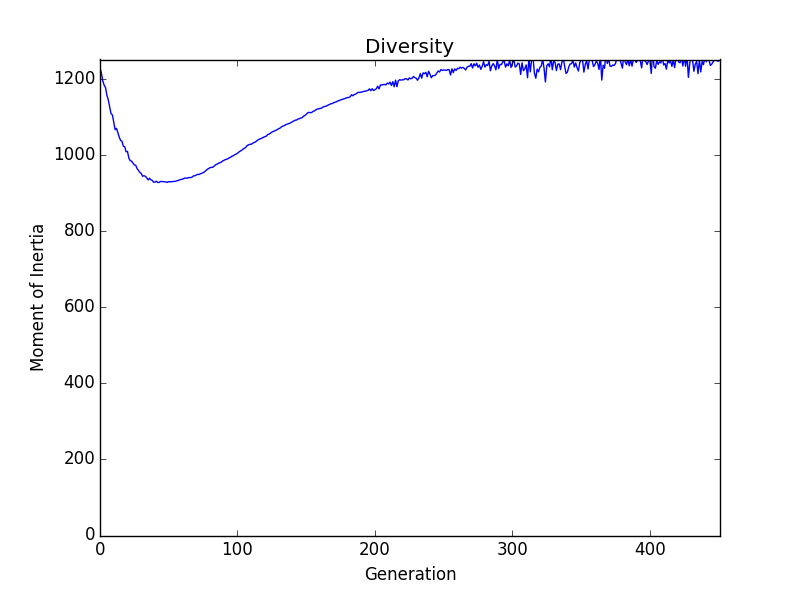}\label{fig:f1}}
	\hfill
	\subfloat[Fitness]{\includegraphics[width=0.5\textwidth,height=5cm]{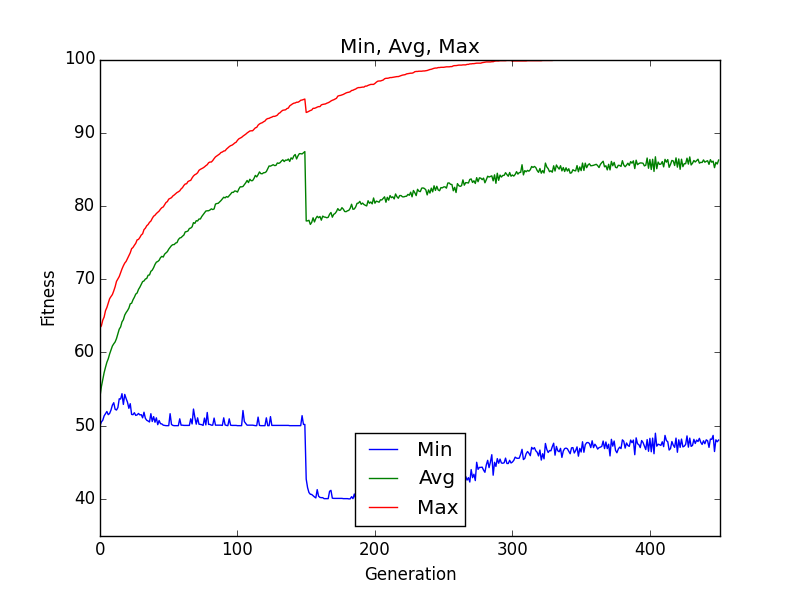}\label{fig:f2}}
	\caption{Performance of New algorithm on Height Changing Peaks}
\end{figure}

The new algorithm shows excellent performance on the height changing peaks problem, particularly when it has converged to both of the peaks. In figure 5.52 (b) it can be seen that the second time the peaks move the maximum fitness in the population doesn't drop at all. This indicates that the algorithm had an individual at the maxima of each peak. This lead to the algorithm having the best offline performance of all the algorithms on this problem.

\begin{figure}[H]
	\centering
	\subfloat[Diversity]{\includegraphics[width=0.5\textwidth,height=5cm]{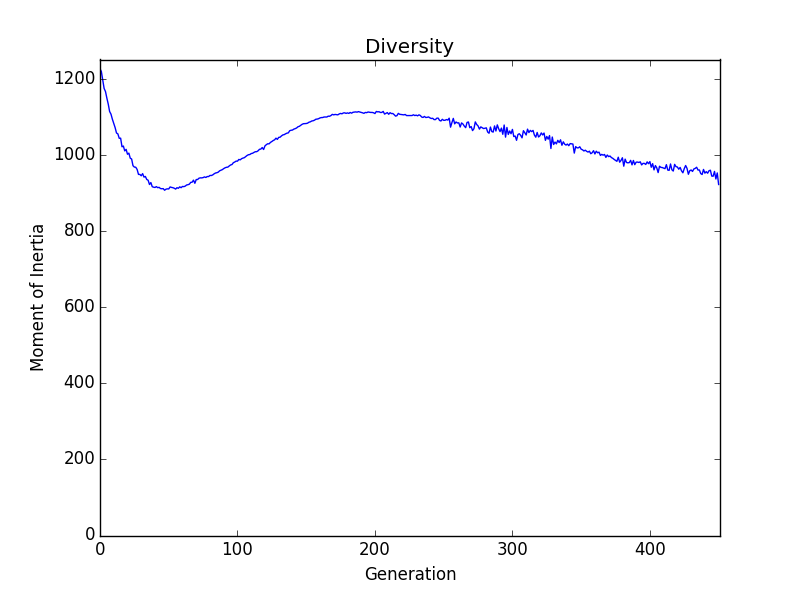}\label{fig:f1}}
	\hfill
	\subfloat[Fitness]{\includegraphics[width=0.5\textwidth,height=5cm]{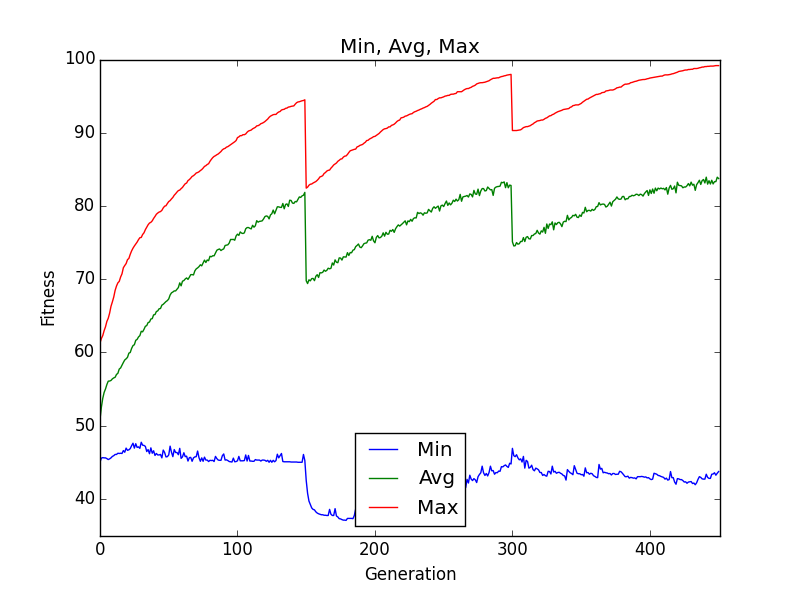}\label{fig:f2}}
	\caption{Performance of New algorithm on Moving Height Changing Peaks}
\end{figure}

On the moving height changing peaks problem the new algorithm again outperforms all the other algorithms. While it is unable to locate the exact optima it comes much closer than any of the other algorithms that found both peaks did. Being consistently near the optima meant that the new algorithm had the highest offline performance of any algorithm on this problem. 

\begin{center}
	\captionof{table}{Offline Performance and Maximum Achieved Fitness of the New Algorithm on Dynamic Problems}
	\begin{tabular}{| p{2cm} | p{2cm} | p{2cm} | p{2cm} | p{2cm} |}
		\hline
		Problem & One Moving Peak & Two Moving Peaks & Height Changing Peaks & Moving Height Changing Peaks \\ \hline
		Offline Performance & 94.910 & 91.030 & 93.716 & 90.249  \\ \hline
		Maximum Achieved Fitness&100\newline 100\newline 100 & 95\newline 99\newline 99  & 95\newline 100\newline 100 & 95\newline 98\newline 99  \\ \hline
		Finds Both Peaks & N/A & Yes & Yes & Yes  \\ \hline
		
		\hline
	\end{tabular}
\end{center}

Overall the new algorithm was very successful, it had the strongest performance on all of the problems that required tracking multiple peaks and still performed well on the single peak problem. These results would be an indication that my idea about taking the strengths from the different categories of algorithms is a valid one. However these results are generated on a small number of problems using a single algorithm and would require more extensive testing to be confirmed.

\chapter{Conclusions}

This dissertation set out to investigate the performance and behaviour of evoutionary algorithms on dynamic optimisation problems, with specific focus on how a range of diversity mechanisms affect said performance.

Throughout the results the algorithms tested in this report have been found to fall into two categories. Category one algorithms are strongest on problems where it is necessary to locate multiple peaks in the fitness landscape but are comparatively weak on problems where fast convergence is necessary. Category two algorithms are strong where category one algorithms are weak. As such they perform well on problems needing quick convergence but are fail on problems requiring high population diversity as they are unable to maintain subpopulations.

All of the mandatory and desired requirements have been met. The optional requirement of increasing how often the peaks move would not have provided useful results. This is due to all of the algorithms failing to adequately solve the current hardest problem, thus increasing the difficulty would not yield any new information. The other optional requirement of calculating the best-of-generation fitness was met, however the values were so similar to the offline performance values that they have been omitted from this report.

While this dissertation did not initially set out to create a new algorithm one was made after the algorithms under investigation failed to solve the harder problems. This new algorithm was very successful and has prompted further research questions.

\section{Further Work}

The work carried out in this project can be taken further in multiple ways. A simple project extension would be to look at the performance of diversity mechanisms not covered in this report and see how they stack up against the ones that were tested. 

Further investigation could be made into the effects that diversity mechanisms have on the performance of evolutionary algorithms by varying some of the factors that were kept static in this investigation. For example looking at the effects of altering population size, crossover and mutation rates or using different crossover and mutation functions. Changing these parameters could potentially have notable consequences on algorithm performance as was seen in the altered bit flip probability on the two moving peaks problem.

The final avenue for further work would be to look at how other combinations of algorithms perform. The one algorithm that was created in this project showed promising performance and investigation into other algorithms created in this way could find some highly successful algorithms.

\bibliographystyle{unsrt} 
\bibliography{dissertationbibliography}

\end{document}